\documentclass[twoside,11pt]{article}

\usepackage{csvsimple}

\usepackage{jmlr2e}

\usepackage{amsmath, bm, bbm}
\usepackage{amssymb}
\usepackage{enumitem}
\usepackage{xcolor}

\usepackage{mathtools}
\usepackage{placeins}
\usepackage{relsize}
\usepackage{xfrac}
\usepackage{booktabs}
\usepackage{lastpage}
\usepackage[figuresright]{rotating}

\usepackage[ruled,vlined]{algorithm2e}
\usepackage{verbatim}
\usepackage{booktabs}

\DeclareMathOperator*{\argmin}{arg\,min}
\DeclareMathOperator*{\argmax}{arg\,max}

\firstpageno{1}

\begin{document}
\jmlrheading{24}{2023}{1-\pageref{LastPage}}{9/21}{10/23}{21-1130}{Dimitris Bertsimas, Ryan Cory-Wright, and Nicholas A. G. Johnson}
\ShortHeadings{Sparse Plus Low Rank Matrix Decomposition: A Discrete Optimization Approach}{Bertsimas, Cory-Wright and Johnson}

\title{Sparse Plus Low Rank Matrix Decomposition: \\ A Discrete Optimization Approach}

\author{\name Dimitris Bertsimas \email dbertsim@mit.edu \\
       \addr Massachusetts Institute of Technology\\
       Cambridge, MA 02139, USA
       \AND
       \name Ryan Cory-Wright \email r.cory-wright@imperial.ac.uk \\
       \addr Imperial College Business School \\
       London, SW7 2AZ, UK\\
       \AND
       Nicholas A. G. Johnson \email nagj@mit.edu\\
       \addr Massachusetts Institute of Technology\\
       Cambridge, MA 02139, USA}

\editor{Sathiya Keerthi}

\maketitle

\begin{abstract}
We study the Sparse Plus Low-Rank decomposition problem (SLR), which is the problem of decomposing a corrupted data matrix 
into a sparse matrix 
{\color{black}of perturbations} plus a low{\color{black}-}rank matrix {\color{black}containing the ground truth}. 
SLR is a fundamental problem in Operations Research and Machine Learning {\color{black}which arises} in {\color{black}various} applications, {\color{black}including} data compression, latent semantic indexing, collaborative filtering{\color{black},} and medical imaging. We introduce a novel formulation for SLR that directly models {\color{black}its} underlying discreteness. For this formulation, we develop an alternating minimization heuristic {\color{black}that computes} high{\color{black}-}quality solutions and a novel semidefinite relaxation that provides meaningful bounds for the solutions returned by our heuristic. {\color{black}We also} develop a custom branch{\color{black}-}and{\color{black}-}bound {\color{black}algorithm} that leverages our heuristic and convex relaxation{\color{black}s to} solve small instances of SLR to certifiable {\color{black}(}near{\color{black})} optimality. {\color{black}Given an input $n$-by-$n$ matrix,} our heuristic scale{\color{black}s to solve instances where} $n=10000$ in {\color{black} minutes}, our relaxation scale{\color{black}s} to {\color{black}instances where} $n=200$ in hours, and our branch-and-bound algorithm scale{\color{black}s} to {\color{black}instances where} $n=25$ in minutes. Our numerical results demonstrate that our approach outperforms existing state-of-the-art approaches {\color{black} in terms of rank, sparsity, and mean-square error while maintaining a comparable runtime.}
\end{abstract}

\begin{keywords}
  Sparsity; Rank; Matrix Decomposition; Convex Relaxation; Branch{\color{black}-}and{\color{black}-b}ound
\end{keywords}

\section{Introduction}

The \textit{Sparse Plus Low Rank} (SLR) decomposition problem, or the problem of approximately decomposing a data matrix $\bm{D} \in \mathbb{R}^{n \times n}$ into a sparse matrix $\bm{Y}$ plus a low-rank matrix $\bm{X}$,  arises throughout many fundamental applications in Operations Research, Machine Learning, and Statistics, including collaborative filtering \citep{recht2010guaranteed}, medical resonance imaging \citep{chen2017low}, and economic modeling \citep{Basu_2019} among others. Formally, given a target rank $k_0$ and a target sparsity $k_1$, we solve:
\begin{equation}
\begin{aligned}
    \min_{\bm{X}, \bm{Y} \in \mathbb{R}^{n \times n}} & \Vert\bm{D} - \bm{X} - \bm{Y}\Vert_F^2 + \lambda \Vert\bm{X}\Vert_F^2 + \mu \Vert\bm{Y}\Vert_F^2 \ \text{s.t.} \ \mathrm{Rank}(\bm{X}) \leq k_0, \Vert\bm{Y}\Vert_0 \leq k_1,
\end{aligned} \label{opt:main_problem}
\end{equation} where $\lambda, \mu>0$ are parameters that control sensitivity to noise and are to be cross-validated {\color{black} by minimizing a validation metric \citep[see, e.g.,][]{validation} to obtain strong out-of-sample performance in theory and practice \citep{bousquet2002stability}}. 



In SLR decomposition problem{\color{black}s}, the sparse matrix $\bm{Y}$ accounts for a small number of {\color{black}potentially} large corruptions in $\bm{D}$, while $\bm{X}$ models the leading principal components of $\bm{D}$ after this corruption is removed. This is well justified, 
because SLR robustifies Principal Component Analysis (PCA), a leading technique for finding low-rank approximations of noiseless datasets 
\citep{pearson1901liii}, 
which performs poorly in high-dimensional settings and in the presence of noise \citep{negahban2011estimation}. 
In an opposite direction, SLR robustly accounts for noise via the sparse matrix $\bm{Y}$, while $\bm{X}$ recovers the uncorrupted principal component directions of $\bm{D}$. {\color{black} Correspondingly, SLR decomposition schemes, which are also called Robust PCA since at least the work of \citet{candes2011robust}, are widely regarded as state-of-the-art approaches for high-dimensional matrix estimation problems \citep{chandrasekaran2011rank, negahban2011estimation}.

Our formulation \eqref{opt:main_problem} is also well-justified from an information-theoretic perspective. Indeed, several authors \citep{arous2020free, gamarnik2021overlap} have demonstrated for special cases of Problem \eqref{opt:main_problem} that when the ground truth is sparse and/or low-rank, exact sparse and/or low-rank formulations recover the ground truth at least as accurately as any polynomial time method, and indeed there is a gap between the amount of data required for an ``exact'' sparse plus low-rank formulation to recover the ground truth, and the amount of data required for a polynomial time approach \citep[an Overlap Gap Property][]{gamarnik2021overlap}.
}
A key characteristic of Problem \eqref{opt:main_problem} is that it directly employs a sparsity constraint on $\bm{Y}$ and a rank constraint on $\bm{X}$. These constraints are non-convex, which make \eqref{opt:main_problem} a difficult problem to solve exactly, both in practice—where the {\color{black}best-known} exact algorithms cannot certify optimality beyond $n=10$ \citep{lee2014optimal}—and in theory, where the problem is NP-hard by reduction from low{\color{black}-}rank matrix approximation \citep{gillis2011low}. 

In this work, we develop an alternating minimization heuristic and convex relaxation which collectively provide very small bound gaps for \eqref{opt:main_problem} and scale to high-dimensional settings. Our heuristic scales to $n=10000$ in {\color{black} minutes} and our convex relaxation scales to $n=200$ in hours. A key feature of the approach is that it leverages the underlying discreteness of the problem to obtain tight yet computationally cheap lower bounds. We further demonstrate that the alternating minimization heuristic and convex relaxation can be embedded within a branch-and-bound tree to solve \eqref{opt:main_problem} to certifiable near-optimality for instances of size up to $n = 25$.

\subsection{Contribution and Structure}

The key contributions of the paper {\color{black}are threefold:}

\begin{itemize}[topsep=0.5ex,itemsep=-0.25ex]
    \item {\color{black}First, from a methodological perspective,} we introduce a novel formulation \eqref{opt:main_problem} for the SLR decomposition problem that directly exploits the underlying discreteness of the problem. Our formulation is inspired {\color{black}by} incorporating robustness against adversarial perturbations in the input data in SLR, which is useful in noisy setting{\color{black}s}.
    \item {\color{black}Second, from an algorithmic perspective,} we develop a heuristic {\color{black}that obtains} high quality feasible solutions to {\color{black}Problem} \eqref{opt:main_problem} {\color{black} in Section \ref{sec:AM} and} derive a convex relaxation of \eqref{opt:main_problem} that provides {\color{black}high-quality} bounds for the solutions returned by our heuristic {\color{black} in Section \ref{sec:sdp_lb}. We also interpret the convex relaxation as a novel reverse Huber penalty which penalizes the sparse and low-rank matrices in a convex manner}. {\color{black}Further, w}e present a branch-and-bound framework that solves \eqref{opt:main_problem} to certifiable near-optimality for small problem instances {\color{black}in Section \ref{sec:bnb}}.
    \item {\color{black}Third, from a computational perspective, we extensively benchmark our proposed approach. Across a suite of numerical experiments, we} demonstrate {\color{black}in Section \ref{sec:experiments}} that our approach {\color{black}outperforms state-of-the-art non-convex methods like {\color{black} AccAltProj,} GoDec and ScaledGD by obtaining sparser and lower rank matrices with a lower mean-squared error than via prior attempts, in a comparable amount of computational time. Moreover, our approach scales to successfully solve problem instances with $10000 \times 10000$ matrices.}
\end{itemize}




\paragraph{Notation:} We let nonbold face characters such as $b$ denote scalars, lowercase bold-faced characters such as $\bm{x}$ denote vectors, uppercase bold-faced characters such as $\bm{X}$ denote matrices, and calligraphic uppercase characters such as $\mathcal{Z}$ denote sets. We let $[n]$ denote the set of running indices $\{1, ..., n\}$ {\color{black}and $\langle \cdot, \cdot\rangle$ denote the Euclidean (Frobenius) inner product between two vectors (matrices) of the same dimension}. We let $\mathbf{e}$ denote a vector of all $1$'s, $\bm{0}$ denote a vector of all $0$'s, and $\mathbb{I}$ denote the identity matrix. {\color{black}Finally, we let $\mathcal{S}^n$ ($\mathcal{S}^n_+$) denote the cone of $n \times n$ symmetric (positive semidefinite) matrices.}

\section{Literature Review and SLR Formulation Properties} \label{sec:litreview_robustness}

In this section, we {\color{black}judiciously characterize Problem \eqref{opt:main_problem} and state-of-the-art approaches for addressing it. First, in Section \ref{sec:litreview}, we cast a deliberate eye over existing attempts at solving Problem \eqref{opt:main_problem} that are currently considered to be state-of-the-art and establish that these approaches are either heuristics that do not provide performance guarantees or branch-and-bound methods that do not scale to even moderate problem sizes. Next, in Section \ref{ssec:ofv}, we establish several key properties of Problem \eqref{opt:main_problem}'s objective function that we invoke throughout the paper. Further, in Section \ref{ssec:robustness}, we justify the regularization terms in our formulation by interpreting our formulation through the lens of robust optimization. Finally, in Section \ref{sec:matcomp}, we 
characterize the conditions under which Problem \eqref{opt:main_problem} admits a reduction to matrix completion, a famous and frequently studied cousin of Problem \eqref{opt:main_problem} which is notoriously computationally challenging \citep{candes2010matrix}.}

\subsection{Literature Review}\label{sec:litreview}
{\color{black}In this section,} we {\color{black}selectively }review several formulations from the literature that have been employed to solve the sparse {\color{black}plus} low-rank decomposition problem {\color{black}and are currently considered to be state-of-the-art}. {\color{black}Most} of these approaches are heuristic in nature and do not provide valid lower bounds to certify {\color{black}the (}sub{\color{black})} optimality of the output solution. 

\subsubsection{Stable 
Principal Component Pursuit}
{\color{black}Optimizing over low-rank matrices is notoriously computationally challenging in both theory and practice \citep{recht2010guaranteed, bertsimas2020mixed}. Accordingly,} a popular approach is to replace the rank and sparsity terms with {\color{black}their nuclear norm and $\ell_1$ norm surrogates}, as advocated by \cite{chandrasekaran2011rank, candes2011robust} {\color{black}among others}. In the presence of noise, this {\color{black}substitution} leads to the following formulation, which was originally proposed by \cite{stable} and is {\color{black}called} Stable Principal Component Pursuit (S-PCP):

\begin{equation}
\begin{aligned}
    \min_{\bm{X}, \bm{Y} \in \mathbb{R}^{n \times n}} \quad & \Vert\bm{X}\Vert_* + \frac{1}{\sqrt{n}}\Vert\bm{Y}\Vert_1 + \frac{1}{2\mu}\Vert\bm{D}-\bm{X}-\bm{Y}\Vert_F^2.
\end{aligned} \label{opt:stable_candes}
\end{equation} 

Problem \eqref{opt:stable_candes} can either be reformulated as a semidefinite problem over a $2n \times 2n$ matrix as advocated by \cite{candes2011robust}, solved in the original space using a nonsymmetric interior point method as proposed by \cite{skajaa2015homogeneous} or solved in a semidefinite free fashion using an augmented Lagrangian approach as advocated by \citet{yuan2009sparse}. Unfortunately, all three approaches require repeatedly performing operations such as a singular value decomposition or a Newton step, which has an $O(n^3)$ or higher time/memory cost. Correspondingly, {\color{black}all such semidefinite optimization approaches require too much memory to be successfully implemented in a standard computational environment when $n=200$, at least with current technology \citep[see][for a review of the state-of-the-art in semidefinite optimization]{majumdar2020recent}. Moreover, these methods are usually only guaranteed to recover a ground truth model under a mutual incoherence condition (or similar) on the ground truth \citep[see][for a review]{tillmann2013computational}, which implies that performance guarantees for such semidefinite methods are challenging to obtain indeed.}

\subsubsection{GoDec}

{\color{black} Many} existing formulations for SLR employ convex relaxations of the rank function and the $\ell_0$ norm function rather than exploiting the inherent discreteness of the problem. {\color{black}An exception to this pattern is the work of \cite{goDec}, who} leverage discreteness to obtain higher quality solutions to SLR. Their formulation is given by:

\begin{equation}
\begin{aligned}
    \min_{\bm{X}, \bm{Y} \in \mathbb{R}^{n \times n}} & \Vert\bm{D} - \bm{X} - \bm{Y}\Vert_F^2\ \text{s.t.} \ \mathrm{Rank}(\bm{X}) \leq k_0, \Vert\bm{Y}\Vert_0 \leq k_1.
\end{aligned} \label{opt:goDec}
\end{equation} 
Note that \eqref{opt:goDec} differs from \eqref{opt:main_problem} by the absence of regularization terms {\color{black}on} $\bm{X}$ and $\bm{Y}$. \cite{goDec} obtain a feasible solution to \eqref{opt:goDec} by performing alternating minimization on $\bm{X}$, $\bm{Y}$. Their algorithm{\color{black}, called GoDec,} is similar in structure to the algorithm we develop {\color{black}in Section \ref{sec:AM}} to obtain {\color{black}high-quality} solutions to {\color{black}Problem} \eqref{opt:main_problem}. {\color{black}In a related direction, \cite{yan2015simultaneous} adopt a similar approach to GoDec in the special case where} the{\color{black}ir} design matrix is taken to be the identity. {\color{black} \cite{kyrillidis2012matrix} adopt a similar formulation as GoDec, however, they instead minimize the reconstruction error between an observation vector and a vector-valued linear map of the sum of the low-rank and sparse matrices. In a somewhat different vein, \cite{zhang2017robust} consider an explicit rank constraint but not a sparsity constraint and proceed by leveraging manifold optimization techniques.}

{\color{black} \subsubsection{Low Rank Matrix Parameterization}}

An extensively studied family of methods parameterizes the low-rank matrix $\bm{X}$ as $\bm{X} = \bm{U} \bm{V}^T$ where $\bm{U}, \bm{V} \in \mathbb{R}^{n \times k_0}$, and performs alternating minimization on $\bm{U}, \bm{V}$. Originally proposed in the context of low-rank semidefinite optimization by \cite{burer2003nonlinear, burer2005local} \citep[see also][]{jain2013low}, it has since evolved into an extensively used and practical approach for SLR problems \citep{netrapalli2014non, chen2015fast, gu2016low, cai2019accelerated}. This approach eliminates the rank constraint and can substantially reduce the number of variables when $n \ll k_0$ at the expense of introducing non-convexity in the objective. Remarkably, in many circumstances, the induced non-convexity is benign and the resulting Burer-Monteiro reformulation can be solved efficiently from both a theoretical and a practical perspective. We refer readers to \cite{chi2019nonconvex} for a detailed overview. 

Two important parametrization-based approaches to SLR are Fast RPCA \citep{yi2016fast} and Scaled Gradient Descent \citep{tong2021accelerating}. In Fast RPCA, after parametrizing the low-rank matrix, \cite{yi2016fast} augment the objective with a regularization term on the norm of $(\bm{U}^T\bm{U} - \bm{V}^T\bm{V}) \in \mathbb{R}^{k_0 \times k_0}$ before performing alternating minimization on $\bm{U}$ and $\bm{V}$. In an alternate direction, \cite{tong2021accelerating} perform{\color{black}s} iterative gradient descent updates on $\bm{U}$ and $\bm{V}$ in Scaled Gradient Descent after designing an effective gradient preconditioner that results in desirable convergence behavior even for ill-conditioned problems. However, {\color{black}existing} performance guarantees for these approaches rely on assumptions on the structure of the ground truth, such as mutual incoherence, that are difficult to verify without independent access to the ground truth or on being initialized within a ``basin of attraction'' which similarly is difficult to verify. {\color{black}We point out, however, that one could either use the dual bounds derived in this paper, or side information such as scoring by humans (e.g., in video background separation applications) to provide performance guarantees when the ground truth is not known.}

\subsubsection{Branch and Bound}
{\color{black}To our knowledge, the only existing work that provides guarantees on the quality of solutions to Problem \eqref{opt:main_problem} is \cite{lee2014optimal}, who propose} a branch-and-bound {\color{black}algorithm} for solving Problem \eqref{opt:main_problem} to near-optimality. Specifically, they assume that the spectral norm of $\bm{X}$ is bounded from above by $\beta$, i.e., $\beta \geq \Vert \bm{X}\Vert_\sigma$, and invoke the following inequality to obtain valid lower bounds for each partially specified sparsity pattern {\color{black}\citep[see also][]{fazel2002matrix}}:
\begin{align}
    \frac{\gamma}{\alpha}\Vert \bm{Y}\Vert_1+\frac{1}{\beta}\Vert \bm{X}\Vert_* \leq \gamma \Vert \bm{Y}\Vert_0+\mathrm{Rank}(\bm{X}),
\end{align}
where $\alpha \geq \Vert \bm{Y}\Vert_\infty$ is a bound on the $\ell_\infty$ norm of $\bm{Y}$, which can either be taken to be equal to some large fixed constant $M$ \citep{glover1975improved} or treated as a regularization parameter \citep{bertsimas2019unified}. Unfortunately, while \citet{lee2014optimal}'s bound is often reasonable, it was not developed by taking the convex envelope of an appropriate substructure of Problem \eqref{opt:main_problem}, and therefore is not strong enough to solve Problem \eqref{opt:main_problem} {\color{black}to optimality} at even small problem sizes {\color{black} \citep[see also][for a related discussion on the weakness of big-M bounds]{bienstock2010eigenvalue}}. Indeed, the authors reported bound gaps but not optimal solutions {\color{black}for SLR problems} when $n=10$. Nonetheless, this lower bound is potentially interesting in its own right, since it demonstrates that the PCP formulation {\color{black}supplies} a valid lower bound on Problem \eqref{opt:main_problem} if one is willing to {\color{black} either }make a big-$M$ assumption on the spectral norm of the low-rank matrix {\color{black} or compute a valid $M$ \citep[c.f.][Section 3.5]{bertsimas2020mixed}}.

{ \color{black}
\subsection{Objective Function Properties} \label{ssec:ofv}}

We now derive several key properties of Problem \eqref{opt:main_problem} that we leverage throughout the paper and present a probabilistic interpretation of \eqref{opt:main_problem} which is motivated by Bayesian inference. Specifically, we establish that \eqref{opt:main_problem}'s objective is strongly convex, Lipschitz continuous, and the Maximum A Posteriori (MAP) estimator of a suitably defined probabilistic model under a Gaussian prior. Recall that a function $f(\bm{Z})$ is said to be strongly convex with parameter $m$ ($m$-strongly convex) if the function $f(\bm{Z}) - \frac{m}{2}\Vert \bm{Z} \Vert_F^2$ is convex. Similarly, a function $f(\bm{Z})$ is said to be Lipschitz continuous with constant $L$ ($L$-Lipschitz) if the function $\frac{L}{2}\Vert \bm{Z} \Vert_F^2 - f(\bm{Z})$ is convex. Formally, we have the following results (proofs deferred to Appendix \ref{sec:formulation_properties_proofs}):

\begin{proposition} \label{prop:strong_convex}
The function $f(\bm{X}, \bm{Y}) = \Vert\bm{D} - \bm{X} - \bm{Y}\Vert_F^2 + \lambda \Vert\bm{X}\Vert_F^2 + \mu \Vert\bm{Y}\Vert_F^2$ is jointly $m$-strongly convex in $(\bm{X}, \bm{Y})$ over $\mathbb{R}^{n \times n} \times \mathbb{R}^{n \times n}$, i.e., $g(\bm{X}, \bm{Y}) = f(\bm{X}, \bm{Y}) - \frac{m}{2}(\Vert\bm{X}\Vert_F^2+\Vert\bm{Y}\Vert_F^2)$ is jointly convex in $(\bm{X}, \bm{Y})$, for $m = 2 \cdot \min (\lambda, \mu)$. 
\end{proposition}

\begin{proposition} \label{prop:lipschitz}
The function $f(\bm{X}, \bm{Y}) = \Vert\bm{D} - \bm{X} - \bm{Y}\Vert_F^2 + \lambda \Vert\bm{X}\Vert_F^2 + \mu \Vert\bm{Y}\Vert_F^2$ is $L$-Lipschitz continuous in $(\bm{X}, \bm{Y})$ over $\mathbb{R}^{n \times n} \times \mathbb{R}^{n \times n}$ for $L = 2 \cdot \max (\lambda, \mu) + 6$.
\end{proposition}
Note that Propositions \ref{prop:strong_convex}--\ref{prop:lipschitz} collectively imply that the condition number $\kappa$ of $f(\bm{X}, \bm{Y})$ is \begin{align}
    \kappa = \frac{L}{m} = \frac{2 \cdot \max (\lambda, \mu) + 6}{2 \cdot \min (\lambda, \mu)}. \label{eqn:cond}
\end{align}

We now provide a probabilistic interpretation of $f(\bm{X}, \bm{Y})$. Suppose the data $\bm{D} \in \mathbb{R}^{n \times n}$ {are} sampled from
\begin{equation}
    \bm{D} = \bm{X} + \bm{Y} + \bm{\epsilon}, \label{eq:data_gen}
\end{equation} where $\bm{X}, \bm{Y} \in \mathbb{R}^{n \times n}$ are unknown parameters to be estimated and $\bm{\epsilon} \in \mathbb{R}^{n \times n}$, $\epsilon_{ij} \sim N(0, \sigma^2)$ is i.i.d Gaussian noise with variance $\sigma^2$. If we adopt independent Gaussian prior beliefs $X_{ij} \sim N(0, \sfrac{\sigma^2}{\lambda})$ and $Y_{ij} \sim N(0, \sfrac{\sigma^2}{\mu})$ over the parameters $\bm{X}, \bm{Y}$, then the Maximum A Posteriori (MAP) estimate of $\bm{X}, \bm{Y}$ after observing $\bm{D}$ is given by $\argmin_{\bm{X}, \bm{Y}} f(\bm{X}, \bm{Y})$.

To see this, note that the posterior probability after observing $\bm{D}$ is given by
\begin{equation}
    \mathbf{P}(\bm{X}, \bm{Y} \vert \bm{D}) = \frac{\bm{P}(\bm{D} \vert \bm{X}, \bm{Y}) \mathbf{P}(\bm{X}) \mathbf{P}(\bm{Y})}{\mathbf{P}(\bm{D})} \propto \bm{P}(\bm{D} \vert \bm{X}, \bm{Y}) \mathbf{P}(\bm{X}) \mathbf{P}(\bm{Y}).
\end{equation} We can now obtain the MAP estimate by maximizing the posterior probability as follows
\begin{equation*}
    \begin{aligned}
        \argmax_{\bm{X}, \bm{Y}} \bm{P}(\bm{D} \vert \bm{X}, \bm{Y}) & \mathbf{P}(\bm{X}) \mathbf{P}(\bm{Y}) = \argmax_{\bm{X}, \bm{Y}} \prod_{1\leq i, j \leq n} \frac{e^{\frac{-(D_{ij}-X_{ij}-Y_{ij})^2}{2\sigma^2}}}{\sigma \sqrt{2\pi}} \cdot \frac{\sqrt{\lambda}e^{\frac{-\lambda X_{ij}^2}{2\sigma^2}}}{\sigma \sqrt{2\pi}} \cdot \frac{\sqrt{\mu}e^{\frac{-\mu Y_{ij}^2}{2\sigma^2}}}{\sigma \sqrt{2\pi}} \\
        &= \argmin_{\bm{X}, \bm{Y}} \Vert\bm{D} - \bm{X} - \bm{Y}\Vert_F^2 + \lambda \Vert\bm{X}\Vert_F^2 + \mu \Vert\bm{Y}\Vert_F^2 =\argmin_{\bm{X}, \bm{Y}} f(\bm{X}, \bm{Y})
    \end{aligned}
\end{equation*} where the second equality follows by taking a log transformation and 
multiplying by $-2\sigma^2$. 

{\color{black}
\subsection{Equivalence Between Regularization and Robustness}\label{ssec:robustness}
}
Real-world datasets are replete with inaccurate and missing data values, which prevents machine-learning models that do not account for these inconsistencies from generalizing well to unseen data. Accordingly, robustness is a highly desirable attribute for machine learning models, in both theory and practice \citep{xu2009robustness, bertsimas2020robust}. In this section, we demonstrate that our regularized problem \eqref{opt:main_problem} is equivalent to a robust optimization (RO) problem. This result motivates the inclusion of the Frobenius regularization terms within \eqref{opt:main_problem} and verifies that (assuming the hyperparameters in  \eqref{opt:main_problem} are correctly cross-validated), regularization improves \eqref{opt:main_problem}'s out-of-sample performance.

We remark that our results should not be too surprising to readers familiar with the RO literature. Indeed, \cite{bertsimas2018characterization} have already derived a similar result for regularized linear regression problems. However, our main result is strictly more general. Indeed, \cite{bertsimas2018characterization} prove that augmenting an $\ell_2$ loss function with an $\ell_2$ regularization penalty is equivalent to solving a RO problem, and conjecture (but do not prove) that their result can be extended to ordinary least squares regression and ridge regularization (with $\ell_2^2$ rather than $\ell_2$ penalties). On the other hand, we prove a matrix analog of their result and generalize their result to the matrix analog of $\ell_2^2$ regularization. Accordingly, this section may be of independent interest to the RO community.

We now connect our work with the work of \cite{bertsimas2018characterization} by deriving a conceptually simple analog of their characterization of the equivalence of regularization and robustness for sparse plus low-rank problems. This result sheds insight into the nature of regularization as a robustifying force in Problem \eqref{opt:main_problem}. Subsequently, we derive an (admittedly more opaque) characterization of Problem \eqref{opt:main_problem} itself as a RO problem. 

Formally, we have the following results (proofs deferred to Appendix \ref{sec:formulation_properties_proofs}):
\begin{proposition} \label{prop:1}

Let $\mathcal{U}_\lambda(\bm{X}) = \{\bm{\Delta} \in \mathbb{R}^{n \times n} : \Vert\bm{\Delta}\Vert_F \leq \lambda \Vert \bm{X} \Vert_F\}$ for $\bm{X} \in \mathbb{R}^{n \times n}, \lambda > 0$. Consider the robust optimization problem:

\begin{equation}
\begin{aligned}
    \min_{\bm{X}, \bm{Y} \in \mathbb{R}^{n \times n}} \max_{\substack{\bm{\Delta}_1 \in \mathcal{U}_\lambda(\bm{X}) \\ \bm{\Delta}_2 \in \mathcal{U}_\mu(\bm{Y})}} \quad & \Vert\bm{D} + \bm{\Delta_1} + \bm{\Delta_2} - \bm{X} - \bm{Y}\Vert_F \
    \text{\rm s.t.} \ \bm{X} \in \mathcal{V}, \bm{Y} \in \mathcal{W},
\end{aligned} \label{thm:robust_prob}
\end{equation} where $\mathcal{V}$ and $\mathcal{W}$ are arbitrary subsets of $\mathbb{R}^{n \times n}$. Then, \eqref{thm:robust_prob} is equivalent to \eqref{thm:reg_prob}.

\begin{equation}
\begin{aligned}
    \min_{\bm{X}, \bm{Y} \in \mathbb{R}^{n \times n}} \quad & \Vert\bm{D} - \bm{X} - \bm{Y}\Vert_F + \lambda \Vert \bm{X} \Vert_F + \mu \Vert \bm{Y} \Vert_F \
    \text{\rm s.t.} \ \bm{X} \in \mathcal{V}, \bm{Y} \in \mathcal{W}.
\end{aligned} \label{thm:reg_prob}
\end{equation}
\end{proposition} 

\begin{proposition}\label{prop:regrobust_full}
    Problem \eqref{opt:main_problem} is equivalent to the following robust optimization problem:
    \begin{equation}
    \begin{aligned}
    \min_{\bm{X}, \bm{Y} \in \mathbb{R}^{n \times n}} \max_{\substack{\bm{\Delta}_1, \bm{\Delta}_2}} \quad & \Vert\bm{D} - \bm{X} - \bm{Y}\Vert_F^2+\langle \bm{X}, \bm{\Delta_1}\rangle+\langle \bm{Y}, \bm{\Delta}_2\rangle-\frac{1}{4\lambda }\Vert \bm{\Delta}_1\Vert_F^2-\frac{1}{4\mu}\Vert \bm{\Delta}_2\Vert_F^2 \\
    \text{\rm s.t.} \quad & \bm{X} \in \mathcal{V}, \bm{Y} \in \mathcal{W}.
    \end{aligned}
    \end{equation}
\end{proposition}

Taking $\mathcal{V}$ to be the set of matrices with rank at most $k_0$ and $\mathcal{W}$ to be the set of matrices with $\ell_0$ norm at most $k_1$, Proposition \ref{prop:1} implies that performing SLR decomposition with Frobenius regularization is equivalent to solving a RO problem that allows for adversarial errors in the input data matrix $\bm{D}$. Moreover, Proposition \ref{prop:regrobust_full} implies that solving Problem \eqref{opt:main_problem} is equivalent to solving a RO problem with a soft robust penalty term in the objective, rather than a hard constraint on the size of the uncertainty set, as such robust equivalent problems usually consist of. This result is perhaps unsurprising in retrospect, since dual problems to quadratically constrained quadratic problems involve quadratic terms in the objective \citep[see also][Section 6.3]{roos2020universal}.

\subsection{Connection to Matrix Completion}\label{sec:matcomp}

Low-rank matrix completion is a canonical problem in the Statistics and Machine Learning communities that has been employed in control theory \citep{boyd1994linear}, computer vision \citep{candes2010matrix}, and signal processing \citep{ji2010robust} among other applications. Given a partially observed matrix $\bm{D} \in \mathbb{R}^{n \times n}$ where $\Omega \subset \{(i, j): 1 \leq i, j \leq n\}$ denotes the set of indices of the revealed entries, the low-rank matrix completion problem is to compute a low-rank matrix $\bm{X}$ that approximates $\bm{D}$. Low-rank matrix completion solves
\begin{equation}
\begin{aligned}
    \min_{\bm{X} \in \mathbb{R}^{n \times n}} \quad & \sum_{(i, j) \in \Omega} (D_{ij}-X_{ij})^2\ \text{s.t.} \ \mathrm{Rank}(\bm{X}) \leq k_0,
\end{aligned} \label{opt:mat_comp}
\end{equation} where $k_0$ is a predefined target rank.

Although we require $\lambda, \mu > 0$ in our formulation of SLR given by \eqref{opt:main_problem}, we {\color{black}now} show that if we take $\mu = 0$ and also fix a sparsity pattern for the sparse matrix $\bm{Y}$, then \eqref{opt:main_problem} reduces to regularized matrix completion. Let $\bm{Z} \in \{0, 1\}^{n \times n}$ be a matrix such that if $Z_{ij} = 0$, we must have $Y_{ij} = 0$. We refer to $\bm{Z}$ as a valid sparsity pattern for \eqref{opt:main_problem} if $\sum_{ij} Z_{ij} \leq k_1$. {\color{black}Formally, we have (proof deferred to Appendix \ref{sec:formulation_properties_proofs})}:

\begin{proposition} \label{prop:2}
Given a valid sparsity pattern $\bm{Z}$, if we take $\mu = 0$ then \eqref{opt:main_problem} reduces to regularized matrix completion with $\Omega = \{(i, j): Z_{ij} = 0\}$.
\end{proposition}

\section{An Alternating Minimization Heuristic}
\label{sec:AM}

In this section, we propose an alternating minimization algorithm that obtains high-quality feasible solutions to \eqref{opt:main_problem} {\color{black}in Section \ref{sec:amalg}}, by iteratively fixing the sparse or low-rank matrix and optimizing the remaining matrix. This is a reasonable strategy{\color{black},} because {\color{black}alternating minimization (AM) strategies are known to obtain high-quality solutions to low-rank problems \citep{jain2013low} and}, as we demonstrate {\color{black}in Section \ref{ssec:subproblems2}}, when one matrix is fixed the other matrix can be {\color{black}optimized} in closed form. Consequently, Problem \eqref{opt:main_problem} is amenable to {\color{black}AM} techniques. {\color{black} Further, in Section \ref{sec:amalg}, we bound the number of iterations required for AM to converge. Finally, in Section \ref{ssec:optimalityfixed}, we establish that for a fixed sparsity pattern and a sufficiently large amount of regularization, AM yields a globally optimal solution to \eqref{opt:main_problem}. This result provides the basis for the branch-and-bound algorithm we develop in Section \ref{sec:bnb}.} 

{\color{black}
\subsection{Two Natural Subproblems} \label{ssec:subproblems2}}
In this subsection, we derive two subproblems of \eqref{opt:main_problem} by fixing either the sparse matrix $\bm{Y}$ (to obtain a low-rank subproblem) or the low-rank matrix $\bm{Y}$ (to obtain a sparse subproblem). Further, we establish that both subproblems admit closed-form solutions.

\paragraph{Low-Rank Subproblem: }
{First, s}uppose that we fix a sparse matrix $\bm{Y}^*$ in Problem \eqref{opt:main_problem}. Then, \eqref{opt:main_problem} becomes:
\begin{equation}
\begin{aligned}
    \min_{\bm{X} \in \mathbb{R}^{n \times n}} & & \Vert\bm{\bar{D}} - \bm{X}\Vert_F^2 + \lambda \Vert\bm{X}\Vert_F^2 \quad \text{s.t.} \quad \mathrm{Rank}(\bm{X}) \leq k_0,
\end{aligned} \label{opt:fixed_Y}
\end{equation} where $\bm{\bar{D}} = \bm{D} - \bm{Y}^*$ and we omit the regularization term on $\bm{Y}$ since it does not depend on $\bm{X}$. We refer to Problem \eqref{opt:fixed_Y} as the low-rank subproblem. We now demonstrate that this problem admits a closed-form solution, via the following result:

\begin{proposition} \label{prop:rank_subproblem}
Let $\bm{X}^*$ be a matrix such that \[\bm{X}^* = \frac{1}{1+\lambda}\bm{\bar{D}}_{k_0},\] where $\bm{\bar{D}}_{k_0}$ is a top-$k_0$ SVD approximation of $\bm{\bar{D}}$, i.e., $\bm{\bar{D}}_{k_0}= \bm{U}_{k_0} \bm{\Sigma}_{k_0} \bm{V}_{k_0}^T$ where $\bm{\bar{D}} = \bm{U} \bm{\Sigma} \bm{V}^T$ is a singular value decomposition of $\bm{\bar{D}}$. Then, $\bm{X}^\star$ is an optimal solution to Problem \eqref{opt:fixed_Y}.
\end{proposition}

\begin{proof}
It is well known that the solution of the problem 
\begin{equation*}
\begin{aligned}
    \min_{\bm{X} \in \mathbb{R}^{n \times n}} & & \Vert\bm{A} - \bm{X}\Vert_F^2 \quad \text{s.t.} \quad \mathrm{Rank}(\bm{X}) \leq k_0
\end{aligned} \label{opt:svd}
\end{equation*} is given by $\bm{X}^* = \bm{A}_{k_0}$, a projection of $\bm{A}$ onto its first $k_0$ principal components \citep{PCA}. Moreover, since 
\begin{equation*}
\begin{aligned}
    \Vert\bm{\bar{D}} - \bm{X}\Vert_F^2 + \lambda \Vert\bm{X}\Vert_F^2 -\frac{\lambda}{1+\lambda}\Vert\bm{\bar{D}}\Vert_F^2 &= 
     (1+\lambda) \bigg{\Vert}\frac{1}{1+\lambda}\bm{\bar{D}}-\bm{X}\bigg{\Vert}_F^2,
\end{aligned}
\end{equation*} 
it follows that Problem \eqref{opt:fixed_Y} is equivalent to (has the same optimal solution set as) solving 
\begin{equation}
\begin{aligned}
    \min_{\bm{X} \in \mathbb{R}^{n \times n}} & & \bigg{\Vert}\frac{1}{1+\lambda}\bm{\bar{D}} - \bm{X}\bigg{\Vert}_F^2 \quad
    \text{s.t.} \quad \mathrm{Rank}(\bm{X}) \leq k_0.
\end{aligned} 
\label{opt:fixed_Y_equiv}
\end{equation} 
\end{proof} 
In Appendix \ref{sec:appendix_rank_proof}, we provide an alternate proof of Proposition \ref{prop:rank_subproblem} via {strong} duality which reveals that \eqref{opt:fixed_Y} exhibits hidden convexity {in the sense of \cite{ben2014hidden}}.
\begin{remark}
Observe that $\bm{X}^\star$ can be computed exactly in $O(n^2 k)$ time, since we need not compute a full SVD of $\bm{\bar{D}}$. Alternatively, it can be computed approximately using randomized SVD in $O(n^2 \log k)$ time \citep{halko2011finding}.
\end{remark}

\paragraph{Sparse Subproblem: }
{Now, s}uppose we fix a low-rank matrix $\bm{X}^*$ in Problem \eqref{opt:main_problem}. Then, \eqref{opt:main_problem} problem becomes:
\begin{equation}
\begin{aligned}
    \min_{\bm{Y} \in \mathbb{R}^{n \times n}} & & \Vert\bm{\Tilde{D}} - \bm{Y}\Vert_F^2 + \mu \Vert\bm{Y}\Vert_F^2 \quad \text{s.t.} \quad \Vert\bm{Y}\Vert_0 \leq k_1,
\end{aligned} \label{opt:fixed_X}
\end{equation} where $\bm{\Tilde{D}} = \bm{D} - \bm{X}^*$ and we have omitted the regularization term on the low-rank matrix because it does not depend on $\bm{Y}$. We refer to Problem \eqref{opt:fixed_X} as the sparse matrix subproblem. We now demonstrate that this problem also admits a closed-form solution:

\begin{proposition}\label{prop:sparsesubproblem}
Let $\bm{Y}^*$ be a matrix such that \[\bm{Y}^* = \bm{S}^* \circ \bigg{(}\frac{\bm{\Tilde{D}}}{1 + \mu}\bigg{)},\] where $\bm{S}^*$ is a $n \times n$ binary matrix with $k_1$ entries $S_{ij}^\star = 1$ such that $S_{i,j}^\star \geq S_{k,l}^\star$  if $\vert \Tilde{D}_{i,j}\vert \geq \vert \Tilde{D}_{k,l}\vert$ and $\circ$ denotes the Hadamard product operation $((\bm{A} \circ \bm{B})_{ij} = A_{ij} \times B_{ij})$. Then, $\bm{Y}^\star$ solves Problem \eqref{opt:fixed_X}. 
\end{proposition}

\begin{proof}
It is straightforward to show that the solution of:
\begin{equation*}
\begin{aligned}
    \min_{\bm{Y} \in \mathbb{R}^{n \times n}}\quad & \Vert\bm{B} - \bm{Y}\Vert_F^2 \quad \text{s.t.} \quad \Vert\bm{Y}\Vert_0 \leq k_1
\end{aligned} \label{opt:hard_threshold}
\end{equation*} is given by $\bm{Y}^* = \bm{T}^* \circ \bm{B}$ where $\bm{T}^*$ is a $n \times n$ binary matrix with $k_1$ entries $T_{ij}^\star = 1$ such that $T_{i,j}^\star \geq T_{k,l}^\star$  if $\vert B_{i,j}\vert \geq \vert B_{k,l}\vert$. Moreover, since 
\begin{equation*}
\begin{aligned}
    \Vert\bm{\Tilde{D}} - \bm{Y}\Vert_F^2 + \mu \Vert\bm{Y}\Vert_F^2 -\frac{\mu}{1+\mu}\Vert\bm{\Tilde{D}}\Vert_F^2 &= 
     (1+\mu) \bigg{\Vert}\frac{1}{1+\mu}\bm{\Tilde{D}}-\bm{Y}\bigg{\Vert}_F^2,
\end{aligned}
\end{equation*} 
it follows that Problem \eqref{opt:fixed_X} is equivalent to (i.e., has the same optimal solution set as):
\begin{equation}
\begin{aligned}
    \min_{\bm{Y} \in \mathbb{R}^{n \times n}} \quad & \bigg{\Vert}\frac{1}{1+\mu}\bm{\Tilde{D}} - \bm{Y}\bigg{\Vert}_F^2\quad \text{s.t.} \quad \Vert\bm{Y}\Vert_0 \leq k_1.
\end{aligned} \label{opt:fixed_X_equiv}
\end{equation} 
\end{proof} 
In Appendix \ref{sec:appendix_sparse_proof}, we {provide} an alternative proof of Proposition \ref{prop:sparsesubproblem} via strong second-order cone duality which {may be of independent interest as it }reveals that Problem \eqref{opt:fixed_X_equiv} is equivalent to a convex optimization problem.

\begin{remark}
Observe that $\bm{Y}^\star$ can be computed in $O(n^2)$ time, by forming $\bm{\Tilde{D}}$ and partitioning around its $k$th largest absolute element via quicksort. Correspondingly, this step is computationally cheaper than computing an optimal low-rank matrix. Moreover, since $\bm{\Tilde{D}} \in \mathbb{R}^{n \times n}$, this operation is linear in the number of entries of $\bm{\Tilde{D}}$.
\end{remark}



\subsection{{\color{black}An} Alternating Minimization Algorithm}\label{sec:amalg}
{\color{black}By iteratively solving the sparse matrix subproblem and the low-rank matrix subproblem until we either converge to a stationary point or exceed a prespecified number of iterations, we arrive at a feasible solution to \eqref{opt:main_problem}. We formalize this iterative procedure in Algorithm \ref{alg:AM}, and} let $$f(\bm{X}, \bm{Y}) = \Vert\bm{D} - \bm{X} - \bm{Y}\Vert_F^2 + \lambda \Vert\bm{X}\Vert_F^2 + \mu \Vert\bm{Y}\Vert_F^2,$$ {\color{black}be our overall objective function and} $\mathcal{V} = \{\bm{X} \in \mathbb{R}^{n \times n}: \mathrm{Rank}(\bm{X}) \leq k_0\}$, $\mathcal{W}=\{\bm{Y} \in \mathbb{R}^{n \times n}: \sum_{ij} \mathbbm{1}\{Y_{ij} \neq 0\} \leq k_1\}$ {\color{black}denote our respective feasible regions}.

\begin{algorithm}[h] \label{alg:AM}
\SetKwRepeat{Do}{do}{while}
\SetAlgoLined
\KwData{$\bm{D} \in \mathbb{R}^{n \times n}, \lambda, \mu > 0, k_0, k_1 \in \mathbb{Z}^+$, tolerance parameter $\epsilon > 0$.}
\KwResult{$(\bm{\bar{X}}, \bm{\bar{Y}})$ feasible and stationary for Problem \eqref{opt:partial_pattern}}
$\bm{X}_0 \xleftarrow[]{} \bm{0}$; $\bm{Y}_0 \xleftarrow[]{} \bm{0}$\;
$f_0 \xleftarrow[]{} f(\bm{X}_0, \bm{Y}_0)$\;
$t \xleftarrow[]{} 0$\;
\Do{$f_t > 0$ and $\frac{f_{t-1}-f_t}{f_t} \geq \epsilon$}{
  $t \xleftarrow[]{} t + 1$\;
  $\bm{Y}_t \xleftarrow[]{} \argmin_{\bm{Y} \in \mathcal{W}} f(\bm{X}_{t-1}, \bm{Y})$\;
  $\bm{X}_t \xleftarrow[]{} \argmin_{\bm{X} \in \mathcal{V}} f(\bm{X}, \bm{Y}_t)$\;
  $f_t \xleftarrow[]{} f(\bm{X}_t, \bm{Y}_t)$\;}
 \Return{$\bm{\bar{X}} = \bm{X}_t$, $\bm{\bar{Y}} = \bm{Y}_t$}
 \caption{Alternating Minimization Heuristic}
\end{algorithm}

{\color{black}
We note that the initialization {\color{black}strategy} $\bm{X}_0 \xleftarrow[]{} \bm{0}$ and $\bm{Y}_0 \xleftarrow[]{} \bm{0}$ is {\color{black}arbitrary and any initialization strategy could equivalently} be employed. For instance, one could employ a greedy rounding of the solution to the semidefinite relaxation we derive in Section \ref{sec:sdp_lb} as an initialization \citep[see also][Section 4.3]{bertsimas2020mixed}. Moreover, Algorithm \ref{alg:AM} can be executed multiple times for different initializations of $\bm{X}_0$ and $\bm{Y}_0$ to obtain an even higher quality feasible solution to \eqref{opt:partial_pattern}. This could be performed in parallel to avoid significantly increasing computational time. 
}

{\color{black}
It is well-documented in the optimization and machine learning literature that alternating minimization schemes such as Algorithm \ref{alg:AM} produce a sequence of non-increasing iterates that converge to a local minimum; for Algorithm \ref{alg:AM}, this can be shown as a straightforward corollary of \citep[Theorem 1]{goDec}. Building upon this, we now demonstrate that, for a given relative improvement tolerance $\epsilon$,  }Algorithm \ref{alg:AM} terminates in a finite number of iterations. {\color{black}Indeed, }Algorithm \ref{alg:AM} terminates at iteration $t$ if either $f_t = 0$ or $f_t > \big{(}\frac{1}{1+\epsilon}\big{)}f_{t-1}$. For any iteration $t$, the update rules for $\bm{X}_{t+1}$ and $\bm{Y}_{t+1}$ imply that $f_{t+1} = f(\bm{X}_{t+1}, \bm{Y}_{t+1}) \leq f(\bm{X}_{t}, \bm{Y}_{t+1}) \leq f(\bm{X}_{t}, \bm{Y}_{t}) = f_t$. This implies that the sequence $\{f_t\}$ is strictly non-increasing. 


\begin{proposition} \label{prop:AM_termination}
Algorithm \ref{alg:AM} terminates after at most $\frac{\log \frac{\mu+\lambda+\mu\lambda}{\mu\lambda}}{\log 1+\epsilon}$ iterations.
\end{proposition}

\begin{proof}
Assume that $\bm{D} \neq 0$. The case when $\bm{D} = 0$ is trivial as in this setting, Algorithm \ref{alg:AM} terminates immediately because $f_0 = 0$. Suppose Algorithm \ref{alg:AM} has yet to terminate after iteration $t$. This implies that \[0 < f_t \leq \bigg{(}\frac{1}{1+\epsilon}\bigg{)}f_{t-1} \leq \bigg{(}\frac{1}{1+\epsilon}\bigg{)}^t f_0.\] Recall that $f_0 = f(\bm{0}, \bm{0}) = \Vert \bm{D} \Vert _F^2$. Moreover, for all $t$ we must have \[f_t \geq \min_{\bm{X} \in \mathcal{V}, \bm{Y} \in \mathcal{W}}f(\bm{X}, \bm{Y}) \geq \min_{\bm{X}, \bm{Y} \in \mathbb{R}^{n \times n}}f(\bm{X}, \bm{Y}).\] Simple unconstrained minimization gives $\min_{\bm{X}, \bm{Y} \in \mathbb{R}^{n \times n}}f(\bm{X}, \bm{Y}) = \frac{\mu\lambda}{\mu+\lambda+\mu\lambda}\Vert \bm{D} \Vert _F^2$. Combining the above inequalities, we obtain \[\frac{\mu\lambda}{\mu+\lambda+\mu\lambda}\Vert \bm{D} \Vert _F^2 \leq f_t \leq \bigg{(}\frac{1}{1+\epsilon}\bigg{)}^t \Vert \bm{D} \Vert _F^2.\] The result follows by noting that the above inequality is violated if $t > \frac{\log \frac{\mu+\lambda+\mu\lambda}{\mu\lambda}}{\log 1+\epsilon}$.
\end{proof}

{\color{black}In} {\color{black}Section \ref{sec:sdp_lb}, }we {\color{black}complement this result by introducing} a lower bound that can be used to certify the quality of the solution returned by Algorithm \ref{alg:AM}. {\color{black}Moreover, in Section \ref{sec:experiments},} we demonstrate {\color{black}numerically} that Algorithm \ref{alg:AM} produces high{\color{black}-}quality solutions to \eqref{opt:partial_pattern}. 
{\color{black}
\subsection{Optimality of Algorithm \ref{alg:AM} for a Fixed Sparsity Pattern} \label{ssec:optimalityfixed}} 
In this section, we establish the optimality of Algorithm \ref{alg:AM} for a fixed sparsity pattern under certain easy-to-verify conditions that often hold in practice. Accordingly, here and throughout this section, we assume we are given a collection of indices $\mathcal{I}_0 \subset \{(i, j): 1\leq i, j \leq n\}$, $\Vert \mathcal{I}_0 \Vert = n^2 - k_1$ that correspond to entries of the sparse matrix $\bm{Y}$ that {must} take value $0$, and that $\bm{S}^\star$ is a binary matrix that encodes this sparsity pattern. The collection $\mathcal{I}_0$ specifies a complete feasible sparsity pattern for the matrix $\bm{Y}$. 

Given the sparsity pattern specified by $\mathcal{I}_0$, Problem \eqref{opt:main_problem} reduces to
\begin{equation}
\begin{aligned}
    \min_{\bm{X}, \bm{Y} \in \mathbb{R}^{n \times n}} \quad & \Vert\bm{D} - \bm{X} - \bm{Y}\Vert_F^2 + \lambda \cdot \Vert\bm{X}\Vert_F^2 + \mu \cdot \Vert\bm{Y}\Vert_F^2 \\ \text{s.t.} \quad & \mathrm{Rank}(\bm{X}) \leq k_0, \ Y_{ij} = 0 \ \forall (i, j) \in \mathcal{I}_0.
\end{aligned} \label{opt:full_pattern}
\end{equation} Algorithm \ref{alg:AM} can be easily adapted to produce a feasible solution to Problem \eqref{opt:full_pattern}. Indeed, 
by Proposition \ref{prop:sparsesubproblem}, an optimal binary matrix $\bm{Y}^\star$ in \eqref{opt:full_pattern} is given by
\[\bm{Y}^* = \bm{S}^* \circ \bigg{(}\frac{\bm{D}-\bm{X}}{1 + \mu}\bigg{)}.\] 
Moreover, applying Algorithm \ref{alg:AM} with a fixed sparsity pattern and fixed low-rank matrix recovers this sparse matrix automatically. Thus, applying Algorithm \ref{alg:AM} to Problem \eqref{opt:full_pattern} is equivalent to solving the following non-convex optimization problem:
\begin{equation}
\begin{aligned}
    \min_{\bm{X} \in \mathbb{R}^{n \times n}} \quad & \bigg{\Vert}\bm{D} - \bm{X} - \bm{S}^* \circ \bigg{(}\frac{\bm{D}-\bm{X}}{1 + \mu}\bigg{)}\bigg{\Vert}_F^2 + \lambda \cdot \Vert\bm{X}\Vert_F^2 + \mu \cdot \bigg{\Vert}\bm{S}^* \circ \bigg{(}\frac{\bm{D}-\bm{X}}{1 + \mu}\bigg{)}\bigg{\Vert}_F^2 \\[1em]
    \text{s.t.} \quad & \mathrm{Rank}(\bm{X}) \leq k_0.
\end{aligned} \label{opt:full_pattern_X}
\end{equation}

Let us now define some additional notation: let $g(\bm{X})$ denote the objective value function of \eqref{opt:full_pattern_X}, $\Omega = \{\bm{X} \in \mathbb{R}^{n \times n}: \text{Rank}(\bm{X}) \leq k_0\}$ denote the set of $n$-by-$n$ matrices with rank at most $k_0$, $\mathcal{P}_\mathcal{X}(\cdot)$ denote the projection operator onto a set $\mathcal{X} \subseteq \mathbb{R}^{n \times n}$, i.e., $\mathcal{P}_\mathcal{X}(\bm{Y}) = \argmin_{\bm{X} \in \mathcal{X}} \Vert \bm{Y}-\bm{X} \Vert_F^2$, and let $\gamma_k(\bm{X}) = \sfrac{\sigma_{k+1}(\bm{X})}{\sigma_{k}(\bm{X})} \leq 1$ denote the ratio between the $(k+1)$th and the $k$th singular values of $\bm{X}$. 

We have the following result (proof deferred to Appendix \ref{sec:formulation_properties_proofs}): 

\begin{proposition} \label{prop:PGD}

Given a full sparsity pattern $\mathcal{I}_0 \subset \{(i, j): 1\leq i, j \leq n\}$, $\Vert \mathcal{I}_0 \Vert = n^2 - k_1$, if we constrain the binary matrix $\bm{S}^*$ in the solution of the sparse matrix subproblem \eqref{opt:fixed_X} to satisfy $S^*_{ij} = 0 \iff (i, j) \in \mathcal{I}_0$, then Algorithm \ref{alg:AM} is equivalent to performing Projected Gradient Descent on \eqref{opt:full_pattern_X} given by $\bm{X}_{t+1} = \mathcal{P}_\Omega(\bm{X}_t - \eta \nabla g(\bm{X}_t))$ with step size $\eta = \frac{1}{2(1+\lambda)}$. By equivalent, we mean that the two algorithms produce the same sequence of feasible low-rank iterates $\bm{X}_t$ and that we have $f(\bm{X}_t, \bm{Y}_t)=g(\bm{X}_t)$ for all iterations $t$ where $\bm{Y}_t$ denotes the sparse matrix iterates produced by Algorithm \ref{alg:AM}.
\end{proposition}

We are now ready to establish the main result. We have:

\begin{theorem} \label{thm:AM_opt_fixed}

Given a full sparsity pattern $\mathcal{I}_0 \subset \{(i, j): 1\leq i, j \leq n\}$, $\Vert \mathcal{I}_0 \Vert = n^2 - k_1$, let $\bm{S}^\star$ be the binary matrix satisfying $S^\star_{ij} = 0 \iff (i, j) \in \mathcal{I}_0$. Let $\bm{X}^\star$ denote the optimal low-rank matrix for \eqref{opt:full_pattern_X} and define $\Tilde{\bm{D}} = \bigg{(}\frac{1}{1+\lambda} \bigg{[}\bm{D}-\bm{S}^* \circ \bigg{(}\frac{\bm{D}-\bm{X}^\star}{1 + \mu}\bigg{)}\bigg{]}\bigg{)}$. 

Assume $\mathrm{Rank}(\bm{X}^\star) = k_0$ and suppose that the following two conditions hold:
\begin{enumerate}[topsep=0.5ex,itemsep=-0.25ex]
    \item $\lambda + \frac{2\mu}{1+\mu} - 1 > 0$;
    \item $\gamma_{k_0}(\Tilde{\bm{D}}) < \frac{1}{1+\lambda}
    \Big{(}\lambda + \frac{2\mu}{1+\mu} - 1. \Big{)}$.
\end{enumerate} Alternatively, assume $\mathrm{Rank}(\bm{X}^\star) < k_0$ and suppose only the first condition listed above holds. In both of these two settings, Algorithm \ref{alg:AM} converges linearly to the unique optimal solution of Problem \eqref{opt:full_pattern} (where we constrain the binary matrix $\bm{S}^*$ in the solution of the sparse subproblem \eqref{opt:fixed_X} to satisfy $S^*_{ij} = 0 \iff (i, j) \in \mathcal{I}_0$). Specifically, letting $\{(\bm{X}_t, \bm{Y}_t)\}_{t=1}^\infty$ denote the sequence of iterates generated by Algorithm \ref{alg:AM} and $(\bm{X}^*, \bm{Y}^*)$ denote the optimal solution of \eqref{opt:full_pattern}, we have
{\color{black}\begin{equation*}
    \frac{f(\bm{X}_{t+1}, \bm{Y}_{t+1}) - f(\bm{X}^*, \bm{Y}^*)}{f(\bm{X}_t, \bm{Y}_t) - f(\bm{X}^*, \bm{Y}^*)} \leq \frac{1}{(2\lambda+1)(1+\mu)+\mu} \quad \forall \ t.
\end{equation*}}
\end{theorem}
Note that the first condition on the regularization parameters $\lambda$ and $\mu$ in Theorem \ref{thm:AM_opt_fixed} is equivalent to requiring that the objective function of \eqref{opt:full_pattern_X} has a small condition number. The second condition is a more technical one that requires that the gradient of the objective function at the optimal solution of \eqref{opt:full_pattern_X} is never too large. 

\begin{remark}
Theorem \ref{thm:AM_opt_fixed} implies that there is a phase transition in Problem \eqref{opt:main_problem}'s difficulty as the amount of regularization increases. Indeed, when $\mu=0$ and the sparsity pattern is fixed, Problem \eqref{opt:main_problem} is equivalent to matrix completion (Proposition \ref{prop:2}), which is a problem that may admit multiple local minima \citep{bertsimas2020mixed}, and this may cause Algorithm \ref{alg:AM} to converge to a non-global local optimum. On the other hand, our main result implies that, with a sufficiently large regularization term, Problem \eqref{opt:main_problem} can be solved to certifiable optimality by enumerating the sparsity patterns and running alternating minimization on each fixed sparsity pattern. Thus, regularization partially controls the complexity of \eqref{opt:main_problem}.
\end{remark}

\begin{proof}
We establish the result by invoking Theorem 3.3 from \cite{ha2020equivalence}. We prove the result for the more involved case where $\mathrm{Rank}(\bm{X}^\star) = k_0$. The proof for the case where $\mathrm{Rank}(\bm{X}^\star) < k_0$ follows similar reasoning by combining Proposition \ref{prop:PGD} with \citep[Theorem 3.3]{ha2020equivalence}. We observe that the objective function $g(\bm{X})$ of \eqref{opt:full_pattern_X} is $m$-strongly convex and $L$-Lipschitz continuous with $m=2\lambda + \frac{2\mu}{1+\mu}$ and $L = 2\lambda+2$. To see this, note that we have
\[g(\bm{X})-\frac{m}{2}\Vert \bm{X} \Vert_F^2 = \Big{(} \lambda - \frac{m}{2}\Big{)} \Vert \bm{X} \Vert _F^2 + \sum_{(i, j) \in \mathcal{I}_0}(D_{ij}-X_{ij})^2 + \sum_{(i, j) \notin \mathcal{I}_0} (D_{ij}-X_{ij})^2 \cdot \frac{\mu}{1+\mu},\] which is convex when $m = 2\lambda + \frac{2\mu}{1+\mu}$. Similarly, we have
\[\frac{L}{2}\Vert \bm{X} \Vert_F^2 - g(\bm{X})= \Big{(}\frac{L}{2} - \lambda\Big{)} \Vert \bm{X} \Vert _F^2 - \sum_{(i, j) \in \mathcal{I}_0}(D_{ij}-X_{ij})^2 - \sum_{(i, j) \notin \mathcal{I}_0} (D_{ij}-X_{ij})^2 \cdot \frac{\mu}{1+\mu},\] which is convex when $L = 2\lambda+2$. Suppose that $\bm{X}^\star$ is a global minimizer of \eqref{opt:full_pattern_X}. We claim that gradient of $g(\bm{X})$ at $\bm{X}^\star$ satisfies: \[
\Vert \nabla g(\bm{X}^\star) \Vert_\sigma = 2(1+\lambda) \gamma_{k_0}(\Tilde{\bm{D}}) \sigma_{k_0}(\bm{X}^\star),
\] where $\Vert \bm{X} \Vert_\sigma = \sigma_1(\bm{X})$ denotes the spectral norm of $\bm{X}$. To see this, note that since $\bm{X}^\star$ is an optimal solution, it must be a fixed point of \eqref{eq:AM_update}. Thus, we have
\begin{equation*}
    \begin{aligned}
        \Vert \nabla g(\bm{X}^\star) \Vert_\sigma &= 2 (1+\lambda) \bigg{\Vert} \bm{X}^\star- \frac{1}{1+\lambda} \bigg{(} \bm{D} - \bm{S}^* \circ \bigg{(}\frac{\bm{D}-\bm{X}^\star}{1 + \mu}\bigg{)} \bigg{)} \bigg{\Vert}_\sigma \\
        &= 2 (1+\lambda) \Vert \bm{X}^\star - \Tilde{\bm{D}} \Vert_\sigma \\
        &= 2 (1+\lambda) \sigma_{k_0+1}(\Tilde{\bm{D}}) \\
        &= 2 (1+\lambda) \gamma_{k_0}(\Tilde{\bm{D}}) \sigma_{k_0}(\Tilde{\bm{D}}) \\
        &= 2(1+\lambda) \gamma_{k_0}(\Tilde{\bm{D}}) \sigma_{k_0}(\bm{X}^\star),
    \end{aligned}
\end{equation*} where the third and fifth equalities follow from $\bm{X}^\star$ being a fixed point of \eqref{eq:AM_update} and the fourth equality follows from the definition of $\gamma_{k_0}(\Tilde{\bm{D}})$. It is easy to verify that when the first condition of Theorem \ref{thm:AM_opt_fixed} holds, the condition number $\kappa = \frac{L}{m}$ of $g(\bm{X})$ satisfies $\kappa < 2$. Moreover, when the second condition of Theorem \ref{thm:AM_opt_fixed} holds, it can similarly be verified that the gradient of $g(\bm{X})$ at $\bm{X}^\star$ satisfies $\Vert \nabla g(\bm{X}^\star) \Vert_\sigma < (2m-L) \sigma_{k_0}(\bm{X}^\star)$. Invoking the result of Theorem 3.3 from \cite{ha2020equivalence}, $\bm{X}^\star$ is the unique fixed point of Projected Gradient Descent with step size $\eta = \frac{1}{2(1+\lambda)}$. Invoking Proposition \ref{prop:PGD}, this immediately implies that Algorithm \ref{alg:AM} converges to $\bm{X}^\star$.

Finally, it is known that Projected Gradient Descent converges linearly with rate $\frac{\kappa - 1}{\kappa + 1}$ for strongly convex functions \citep{recht2012projected}. Combining this with Proposition \ref{prop:PGD}, we have
\[\frac{g(\bm{X}_{t+1}) - g(\bm{X}^*)}{g(\bm{X}_t) - g(\bm{X}^*)} = \frac{f(\bm{X}_{t+1}, \bm{Y}_{t+1}) - f(\bm{X}^*, \bm{Y}^*)}{f(\bm{X}_t, \bm{Y}_t) - f(\bm{X}^*, \bm{Y}^*)} \leq \frac{\kappa - 1}{\kappa + 1} = \frac{1}{(2\lambda+1)(1+\mu)+\mu},\] which holds for all $t$. This completes the proof. \end{proof}






\section{A Convex Relaxation} \label{sec:sdp_lb}

In this section, we reformulate \eqref{opt:main_problem} as a mixed-integer, mixed-projection optimization problem. We then employ the {\color{black}(matrix)} perspective relaxation \citep{gunluk2012perspective,bertsimas2020mixed, bertsimas2021perspective} to construct a convex relaxation {\color{black}of} \eqref{opt:main_problem}. {\color{black}We illustrate the power of our convex relaxation in Section \ref{ssec:hiddenconvexity}, by demonstrating that it reflects the hidden convexity of the low-rank subproblem we derived in the previous section and allows this subproblem to be solved via convex optimization. Further, we compare our convex relaxation to the previously derived relaxation of \cite{lee2014optimal} in Section \ref{ssec:comparison} and demonstrate that when both relaxations make the same assumptions, our relaxation is at least as powerful, and sometimes strictly more powerful. Finally, in Section \ref{ssec:penaltyintepretation}, we interpret (a slightly modified version of, where the sparsity and rank are penalized rather than constrained) our convex relaxation as a convex penalty.}

To model the sparsity pattern of the sparse matrix $\bm{Y}$, we introduce binary variables $\bm{Z} \in \{0, 1\}^{n \times n}$ and require that {\color{black}$Y_{ij} =0$ if $Z_{i,j}=0$ by imposing the nonlinear constraint $Y_{i,j}=Y_{i,j}Z_{i,j}$, and also require that} $\sum_{ij}Z_{ij} \leq k_1$. To model the column space of $\bm{X}$, we introduce an orthogonal projection matrix $\bm{P} \in \mathcal{P}$ and require that $\text{tr}(\bm{P}) \leq k_0$ and $\bm{X} = \bm{P}\bm{X}$. Let $\mathcal{Z}_{k_1} = \{\bm{Z} \in \{0, 1\}^{n \times n}: \sum_{ij}Z_{ij} \leq k_1\}$ and $\mathcal{P}_{k_0} = \{\bm{P} \in \mathcal{S}^n: \bm{P}^2=\bm{P}, \text{tr}(\bm{P}) \leq k_0\}$. This gives the following reformulation of \eqref{opt:main_problem}: 

\begin{equation}
\begin{aligned}
    \min_{\bm{Z} \in \mathcal{Z}_{k_1}, \bm{P} \in \mathcal{P}_{k_0}} \min_{\bm{X}, \bm{Y} \in \mathbb{R}^{n \times n}} \quad & \Vert\bm{D} - \bm{X} - \bm{Y}\Vert_F^2 + \lambda \cdot \Vert\bm{X}\Vert_F^2 + \mu \cdot \Vert\bm{Y}\Vert_F^2 \\
    \text{s.t.} \quad & \bm{X} = \bm{P}\bm{X}, \bm{Y} = \bm{Z} \circ \bm{Y}.
\end{aligned} \label{opt:main_problem_reform}
\end{equation}
{\color{black}We now have the following result (proof deferred to Appendix \ref{sec:formulation_properties_proofs}):}

\begin{proposition} \label{prop:main_reform}
Problem \eqref{opt:main_problem_reform} is a valid reformulation of Problem \eqref{opt:main_problem}.
\end{proposition}

The constraints $\bm{X} = \bm{P}\bm{X}$ and $\bm{Y} = \bm{Z} \circ \bm{Y}$ in \eqref{opt:main_problem_reform} are complicating because they are non-convex in the decision variables $(\bm{Z}, \bm{P}, \bm{X}, \bm{Y})$. {\color{black}Accordingly, to model these constraints in a convex manner, we invoke }the {\color{black}(matrix)} perspective reformulation \citep{gunluk2012perspective,bertsimas2020mixed, bertsimas2021perspective}. Specifically, to model the sparse matrix $\bm{Y}$, we introduce variables $\bm{\alpha} \in \mathbb{R}^{n \times n}$ where $\alpha_{ij}$ models $Y_{ij}^2$, and the constraint $\alpha_{ij}Z_{ij} \geq Y_{ij}^2$, which is second-order cone representable. 
To model the low-rank matrix $\bm{X}$, we introduce a variable $\bm{\Theta} \in \mathbb{R}^{n \times n}$ that models $\bm{X}^T\bm{X}$, and the constraint $\begin{psmallmatrix}\bm{\Theta} & \bm{X}\\ \bm{X}^T & \bm{P}\end{psmallmatrix} \succeq 0$.

This {\color{black}yields} the following reformulation of \eqref{opt:main_problem_reform}:

\begin{equation}
\begin{aligned}
    \min_{\bm{Z} \in \mathcal{Z}, \bm{P} \in \mathcal{P}} \min_{\bm{X}, \bm{Y} \in \mathbb{R}^{n \times n}} \quad & \Vert\bm{D} - \bm{X} - \bm{Y}\Vert_F^2 + \lambda \cdot {\color{black}\mathrm{tr}(\bm{\Theta}) + \mu \cdot \langle \bm{E}, \bm{\alpha}\rangle} \\
    \text{s.t.} \quad & \bm{Y} \circ \bm{Y} \leq \bm{\alpha} \circ \bm{Z},\ \begin{pmatrix}\bm{\Theta} & \bm{X}\\ \bm{X}^T & \bm{P}\end{pmatrix} \succeq 0,
\end{aligned} \label{opt:main_problem_reform_v2}
\end{equation} where $\bm{E}$ denotes {\color{black}a matrix of all ones of appropriate dimension}.

Problem \eqref{opt:main_problem_reform_v2} is a reformulation of Problem \eqref{opt:main_problem} where the problem's non-convexity is entirely captured by the non-convex sets $\mathcal{Z}_{k_1}$ and $\mathcal{P}_{k_0}$. We now obtain a convex relaxation of \eqref{opt:main_problem} by solving \eqref{opt:main_problem_reform_v2} with $\bm{Z} \in \text{conv}(\mathcal{Z}_{k_1})$ and $\bm{P} \in \text{conv}(\mathcal{P}_{k_0})$ where $\text{conv}(\mathcal{X})$ denotes the convex hull of the set $\mathcal{X}$. It is straightforward to see that $\text{conv}(\mathcal{Z}_{k_1}) = \{\bm{Z} \in [0, 1]^{n \times n}: \sum_{ij}Z_{ij} \leq k_1\}$. Moreover, we have $\text{conv}(\mathcal{P}_{k_0}) = \{\bm{P} \in \mathcal{S}_+^n: \mathbb{I} - \bm{P} \succeq 0, \text{tr}(\bm{P}) \leq k_0\} $\citep{overton1992sum}. This gives the following convex optimization problem:
{\color{black}
\begin{equation}
\begin{aligned}
    \min_{\bm{X}, \bm{Y}, \bm{Z}, \bm{P}, \bm{\Theta}, \bm{\alpha} \in \mathbb{R}^{n \times n}} \quad & \Vert\bm{D} - \bm{X} - \bm{Y}\Vert_F^2 + {\lambda\cdot \color{black}\mathrm{tr}(\bm{\Theta}) + \mu \cdot \langle \bm{E}, \bm{\alpha}\rangle} \\
    \text{s.t.} \quad & \bm{Y} \circ \bm{Y} \leq \bm{\alpha} \circ \bm{Z},\ \langle \bm{E},\bm{Z}\rangle \leq k_1, \ \bm{0} \leq \bm{Z} \leq \bm{E}, \\
    & \bm{P} \succeq 0, \ \mathbb{I} - \bm{P} \succeq 0, \ \text{tr}(\bm{P}) \leq k_0, \ \begin{pmatrix}\bm{\Theta} & \bm{X}\\ \bm{X}^T & \bm{P}\end{pmatrix} \succeq 0.
\end{aligned} \label{opt:convex_relax}
\end{equation}
}

{\color{black}We now have the following result (proof deferred to Appendix \ref{sec:formulation_properties_proofs}):}

\begin{theorem} \label{thm:convex_relax}
Problem \eqref{opt:convex_relax} is a valid convex relaxation of \eqref{opt:main_problem}.
\end{theorem}

{\color{black} Note that Problem \eqref{opt:convex_relax} only produces a nontrivial lower bound to \eqref{opt:main_problem} when the regularization parameters satisfy $\lambda, \mu > 0$. If either $\lambda = 0$ or $\mu = 0$, it can easily be shown that the optimal value of \eqref{opt:convex_relax} is $0$.} In Section \ref{sec:experiments}, we employ this convex relaxation to produce bounds for feasible solutions returned by Algorithm \ref{alg:AM}. Moreover, we show that \eqref{opt:convex_relax} can be embedded within a branch-and-bound framework.

{\color{black}

\subsection{Hidden Convexity in the Low Rank Subproblem}\label{ssec:hiddenconvexity} }
In this section, we demonstrate that the low-rank subproblem derived in the previous section exhibits hidden convexity in the sense of \cite{ben2014hidden}. This result allows us to establish the strength of our overall convex relaxation in the next section. Formally, we have the following result (proof deferred to Appendix \ref{sec:proofofconv}):

\begin{theorem}
Consider the semidefinite optimization problem:
\begin{equation}
\begin{aligned}
    \min_{\bm{P}, \bm{\Theta} \in \mathcal{S}^n_+, \bm{X} \in \mathcal{S}^n} \quad & \Vert\bm{\Bar{D}}\Vert_F^2 + (1+\lambda)\cdot \mathrm{tr}(\bm{\Theta}) - 2\cdot \langle \bm{X},\bm{\Bar{D}}\rangle\\
    \text{\rm s.t.} \quad & \mathrm{tr}(\bm{P}) \leq k_0, \ \mathbb{I} - \bm{P} \succeq 0, \ \begin{pmatrix}\bm{\Theta} & \bm{X}\\ \bm{X}^T & \bm{P}\end{pmatrix} \succeq 0.
\end{aligned} \label{opt:pca_sdp}
\end{equation} Solving Problem \eqref{opt:fixed_Y} is equivalent to solving Problem \eqref{opt:pca_sdp} in that both problems have the same optimal objective value and given an optimal solution to either problem, an optimal solution to the other problem can be constructed efficiently. \label{thm:pca}
\end{theorem}

{\color{black}
\subsection{Comparison With the Relaxation of Lee and Zou}\label{ssec:comparison}}

To illustrate the power of our convex relaxation, we now present a formal comparison between \eqref{opt:convex_relax} and the relaxation proposed by \cite{lee2014optimal} and demonstrate that our relaxation is at least as powerful and sometimes strictly more powerful. Accordingly, here and throughout this subsection, we assume that the spectral norm of the low-rank matrix $\bm{X}$ and the infinity norm of the sparse matrix $\bm{Y}$ are bounded as otherwise the relaxation proposed by \cite{lee2014optimal} yields a lower bound of zero. Explicitly, we assume that $\Vert \bm{X} \Vert_\sigma = \max_i \sigma_i(\bm{X}) \leq \beta$ and $\Vert \bm{Y} \Vert_\infty = \max_{ij} \vert Y_{ij} \vert \leq \gamma$ where $\sigma_i(\bm{X})$ denotes the $i^{th}$ singular value of $\bm{X}$ for $\beta, \gamma \in \mathbb{R}_+$.

\cite{lee2014optimal} obtain their relaxation by noting that under the spectral and infinity norm boundedness assumptions, convex lower bounds of the non-convex rank and $\ell_0$ norm functions can be obtained as $\text{Rank}(\bm{X}) \geq \frac{1}{\beta} \Vert \bm{X} \Vert_\star$ and $\Vert \bm{Y} \Vert_0 \geq \frac{1}{\gamma} \Vert \bm{Y} \Vert_1$ respectively. Noting that the $\ell_1$ norm can be trivially linearized and that the nuclear norm of a matrix $\bm{X}$ admits a well-known semidefinite characterization given by
\begin{equation*}
\begin{aligned}
    \min_{\bm{W}_1, \bm{W}_2 \in \mathcal{S}^n} \quad & \frac{1}{2} \text{tr}(\bm{W}_1+\bm{W}_2) \ \text{s.t.} \ \begin{pmatrix}\bm{W}_1 & \bm{X}\\ \bm{X}^T & \bm{W}_2\end{pmatrix} \succeq 0,
\end{aligned}
\end{equation*}
we can express \cite{lee2014optimal}'s relaxation of \eqref{opt:main_problem} as follows:
\begin{equation}
\begin{aligned}
    \min_{\bm{X}, \bm{Y}, \bm{V}, \bm{W}_1, \bm{W}_2 \in \mathbb{R}^{n \times n}} \quad & \Vert\bm{D} - \bm{X} - \bm{Y}\Vert_F^2 + \lambda \cdot \Vert \bm{X} \Vert_F^2 + \mu \cdot \Vert \bm{Y} \Vert_F^2 \\
    \text{s.t.} \quad & -\bm{V} \leq \bm{Y} \leq \bm{V},\ \frac{1}{\gamma} \langle \bm{E}, \bm{V}\rangle \leq k_1, \\
    & \frac{1}{2\beta} \text{tr}(\bm{W}_1)+\frac{1}{2\beta}\text{tr}(\bm{W}_2) \leq k_0, \ \begin{pmatrix}\bm{W}_1 & \bm{X}\\ \bm{X}^T & \bm{W}_2\end{pmatrix} \succeq 0.
\end{aligned} \label{opt:convex_relax_lee}
\end{equation} 

To allow for a fair comparison between our relaxation and that given by \eqref{opt:convex_relax_lee}, we note that under the assumptions $\Vert \bm{X} \Vert_\sigma \leq \beta$ and $\Vert \bm{Y} \Vert_\infty \leq \gamma$, we can strengthen \eqref{opt:convex_relax} as follows:

\begin{equation}
\begin{aligned}
    \min_{\bm{X}, \bm{Y}, \bm{Z}, \bm{P}_c, \bm{P}_r, \bm{\Theta}, \bm{\alpha} \in \mathbb{R}^{n \times n}} \quad & \Vert\bm{D} - \bm{X} - \bm{Y}\Vert_F^2 + \lambda \cdot \text{tr}(\bm{\Theta}) + \mu \cdot \langle \bm{E}, \bm{\alpha}\rangle \\
    \text{s.t.} \quad &\bm{Y} \circ \bm{Y} \leq \bm{\alpha} \circ \bm{Z}, \ \langle \bm{E}, \bm{Z}\rangle \leq k_1, \ \bm{0} \leq \bm{Z} \leq \bm{E}, \ -\gamma \bm{Z} \leq\bm{Y} \leq \gamma \bm{Z},\\
    & \bm{P}_c \succeq 0, \ \mathbb{I} - \bm{P}_c \succeq 0, \ \text{tr}(\bm{P}_c) \leq k_0, \\
    & \bm{P}_r \succeq 0, \ \mathbb{I} - \bm{P}_r \succeq 0, \ \text{tr}(\bm{P}_r) \leq k_0,\\
    & \begin{pmatrix}\bm{\Theta} & \bm{X}\\ \bm{X}^T & \bm{P}_c\end{pmatrix} \succeq 0, \ \begin{pmatrix}\beta\bm{P}_r & \bm{X}\\ \bm{X}^T & \beta\bm{P}_c\end{pmatrix} \succeq 0.
\end{aligned} \label{opt:convex_relax_strengthened}
\end{equation} 

The constraint $-\gamma Z_{ij} \leq Y_{ij} \leq \gamma Z_{ij}$ in \eqref{opt:convex_relax_strengthened} emerges immediately from the bound on the infinity norm of the sparse matrix. The last four constraints in \eqref{opt:convex_relax_strengthened} follow from the bound on the spectral norm of the low-rank matrix. The variable $\bm{P}_c$ plays the role of $\bm{P}$ in \eqref{opt:convex_relax} and models the $k_0$ dimensional column space of $\bm{X}$ as before while the variable $\bm{P}_r$ models the $k_0$ dimensional row space of $\bm{X}$. To see that these four constraints are valid, consider any matrix $\bm{\Bar{X}}$ satisfying $\Vert \bm{\Bar{X}} \Vert_\star \leq \beta$ and $\text{Rank}(\bm{\Bar{X}}) \leq k_0$, and let $\bm{\Bar{X}} = \bm{U}\bm{\Sigma}\bm{V}^T$ be its singular value decomposition. Define $\bm{\Bar{P}}_c = \bm{U}\bm{U}^T$ and $\bm{\Bar{P}}_r = \bm{V}\bm{V}^T$. We have $\beta^2\bm{\Bar{P}}_r \succeq \bm{\Bar{P}}_r\bm{\Bar{X}}^T \bm{\Bar{X}} = \bm{\Bar{X}}^T \bm{\Bar{P}}_c \bm{\Bar{X}} = \bm{\Bar{X}}^T \bm{\Bar{P}}_c^\dag \bm{\Bar{X}}$ so we have $\begin{pmatrix}\beta\bm{\Bar{P}}_r & \bm{\Bar{X}}\\ \bm{\Bar{X}}^T & \beta\bm{\Bar{P}}_c\end{pmatrix} \succeq 0$. Feasibility of $\bm{\Bar{P}}_c$ and $\bm{\Bar{P}}_r$ for the remaining constraints follows the same reasoning employed in Theorem \ref{thm:convex_relax}. Note that if we restrict $\bm{X}$ to be symmetric, we can take $\bm{P}_r=\bm{P}_c$ in \eqref{opt:convex_relax_strengthened} as the row space and the column space of $\bm{X}$ will be the same.

\begin{proposition} \label{prop:our_relax_lee}
For any input data $\bm{D}, k_0, k_1$ and hyperparameters $\lambda, \mu$, the optimal value of \eqref{opt:convex_relax_strengthened} is no less than the optimal value of \eqref{opt:convex_relax_lee}.
\end{proposition}

\begin{proof}
To establish the proposition, we show that for any feasible solution to \eqref{opt:convex_relax_strengthened} we can construct a feasible solution to \eqref{opt:convex_relax_lee} that achieves the same or lower objective value.

Fix any input data $\bm{D} \in \mathbb{R}^{n \times n}, k_0, k_1 \in \mathbb{N}_+$ and any hyperparameters $\lambda, \mu > 0$. Consider an arbitrary feasible solution $\mathcal{S}_1 = (\bm{\Bar{X}}, \bm{\Bar{Y}}, \bm{\Bar{Z}}, \bm{\Bar{P}}_c, \bm{\Bar{P}}_r, \bm{\Bar{\Theta}}, \bm{\Bar{\alpha}})$ to \eqref{opt:convex_relax_strengthened}. Let $\bm{\Bar{V}} = \gamma \bm{\Bar{Z}}, \bm{\Bar{W}}_1 = \beta \bm{\Bar{P}}_c$ and $\bm{\Bar{W}}_2 = \beta \bm{\Bar{P}}_r$. We will show that the solution $\mathcal{S}_2 = (\bm{\Bar{X}}, \bm{\Bar{Y}}, \bm{\Bar{V}}, \bm{\Bar{W}}_1, \bm{\Bar{W}}_2)$ is feasible to \eqref{opt:convex_relax_lee} and achieves an objective value that is no larger than the objective value achieves by $\mathcal{S}_2$ in \eqref{opt:convex_relax_strengthened}. From feasibility of $\mathcal{S}_1$ in \eqref{opt:convex_relax_strengthened}, we have $-\gamma \Bar{Z}_{ij} \leq \Bar{Y}_{ij} \leq \gamma \Bar{Z}_{ij} \implies -\Bar{V}_{ij} \leq \Bar{Y}_{ij} \leq \Bar{V}_{ij}$ and $\langle \bm{E}, \bm{\Bar{Z}}\rangle \leq k_1 \implies \frac{1}{\gamma} \langle \bm{E}, \Bar{V}_{ij}\rangle \leq k_1$. Moreover, we have
\[
\frac{1}{2\beta}\text{tr}(\bm{\Bar{W}}_1+\bm{\Bar{W}}_2) = \frac{1}{2\beta}\text{tr}(\beta \bm{\Bar{P}}_c +\beta \bm{\Bar{P}}_r) = \frac{1}{2}\text{tr}(\bm{\Bar{P}}_c) + \frac{1}{2}\text{tr}(\bm{\Bar{P}}_c) \leq \frac{k_0}{2} + \frac{k_0}{2} = k_0
\] We conclude that $\mathcal{S}_2$ is feasible to \eqref{opt:convex_relax_lee} by noting that the last constraint in \eqref{opt:convex_relax_lee} reduces to the fourth from last constraint in \eqref{opt:convex_relax_strengthened} after substituting the definitions of $\bm{\Bar{W}}_1$ and $\bm{\Bar{W}}_2$. We observe that $\mathcal{S}_2$ achieves an objective value in \eqref{opt:convex_relax_lee} no greater than that achieved by $\mathcal{S}_1$ in \eqref{opt:convex_relax_strengthened} by noting that feasibility of $\mathcal{S}_1$ implies that $\text{tr}(\bm{\Bar{\Theta}}) \geq \Vert \bm{\Bar{X}} \Vert_F^2$ and $\langle \bm{E},\bm{\Bar{\alpha}}\rangle \geq \Vert \bm{\Bar{Y}} \Vert_F^2$. Since this construction holds for every feasible solution to \eqref{opt:convex_relax_strengthened}, it must hold for any optimal solution, which implies that the optimal value of \eqref{opt:convex_relax_lee} is no greater than the optimal value of \eqref{opt:convex_relax_strengthened}. This completes the proof.
\end{proof} 

Proposition \ref{prop:our_relax_lee} establishes that our relaxation is at least as strong as \eqref{opt:convex_relax_lee}, but does not in and of itself demonstrate its utility since it does not preclude the possibility of the optimal value of \eqref{opt:convex_relax_strengthened} always coinciding with the optimal value of \eqref{opt:convex_relax_lee}. To address this, Proposition \ref{prop:we_are_better} which establishes the existence of problem instances for which the optimal value of \eqref{opt:convex_relax_strengthened} is strictly greater than the optimal value of \eqref{opt:convex_relax_lee}. Taken together, Propositions \ref{prop:our_relax_lee} and \ref{prop:we_are_better} show that \eqref{opt:convex_relax_strengthened} is a (strictly) stronger convex relaxation to \eqref{opt:main_problem} than \eqref{opt:convex_relax_lee}.

\begin{proposition} \label{prop:we_are_better}
There exists input data $\bm{D}, k_0, k_1$ and hyperparameters $\lambda, \mu$ such that the optimal value of \eqref{opt:convex_relax_strengthened} is strictly greater than the optimal value of \eqref{opt:convex_relax_lee}.
\end{proposition}

\begin{proof}
We establish the result constructively. Let $n=2, \bm{D} = \mathbb{I}_2, k_0 = 1, k_1 = 0, \lambda = 1$ and $\mu=1$. With these values, \eqref{opt:main_problem} reduces to 
\begin{equation}
\begin{aligned}
    \min_{\bm{X} \in \mathbb{R}^{2 \times 2}} \quad & \Vert \mathbb{I}_2 - \bm{X}\Vert_F^2 + \Vert\bm{X}\Vert_F^2 \ \text{s.t.} \ \mathrm{Rank}(\bm{X}) \leq 1.
\end{aligned} \label{opt:lee_toy}
\end{equation} It follows immediately from Proposition \ref{prop:rank_subproblem} that the optimal solution to \eqref{opt:lee_toy} is $\bm{X}^\star = \begin{pmatrix}0.5 & 0 \\ 0 & 0\end{pmatrix}$ and the optimal objective value is $\frac{3}{2}$. Let $\beta = 2$ and $\gamma = 1$. Note that $\gamma$ can be chosen arbitrarily since the optimal sparse matrix is $\bm{Y}^\star = \bm{0}$. Consider solving \eqref{opt:convex_relax_strengthened} and \eqref{opt:convex_relax_lee} for this problem data. From Theorem \ref{thm:pca}, it follows that the optimal value of \eqref{opt:convex_relax_strengthened} coincides with the optimal value of \eqref{opt:lee_toy}. Next, note that if we ignore the rank constraint, it can easily be verified that the unconstrained minimum of \eqref{opt:lee_toy} is given by $\bm{\Tilde{X}}=\frac{1}{2}\mathbb{I}$ and achieves an objective value of $1$. Finally, observe that taking $\bm{\Tilde{Y}} = \bm{\Tilde{V}} = \bm{0}, \bm{\Tilde{W}}_1 = \bm{\Tilde{W}}_2 = \mathbb{I}$, the solution $(\bm{\Tilde{X}}, \bm{\Tilde{Y}}, \bm{\Tilde{V}}, \bm{\Tilde{W}}_1, \bm{\Tilde{W}}_2)$ is feasible to \eqref{opt:convex_relax_lee} and achieves an objective value of $1$. This completes the proof. \end{proof}

{\color{black}\subsection{Penalty Interpretation of Relaxation}\label{ssec:penaltyintepretation}}
We now consider instances where the sparsity and rank of the matrices are penalized in the objective rather than constrained and interpret the resulting relaxation as a penalty function in the tradition of \cite{fazel2002matrix, recht2010guaranteed, pilanci2015sparse, bertsimas2020mixed} among others. Formally, we have the following result\footnote{Note that the statement of our result is slightly different to the statement in \cite{pilanci2015sparse}, because, as noted by \citet{dong2015regularization}, the original result contains some minor typos.}, which can be deduced by combining \citep[Corollary 3]{pilanci2015sparse} with \citep[Lemma 6]{bertsimas2020mixed}:
\begin{proposition}\label{prop:revhuber}
The following two optimization problems are equivalent:
\begin{equation}
\begin{aligned}
    \min_{\bm{X}, \bm{Y}, \bm{Z}, \bm{P}, \bm{\Theta}, \bm{\alpha} \in \mathbb{R}^{n \times n}} \quad & \Vert\bm{D} - \bm{X} - \bm{Y}\Vert_F^2 + {\lambda\color{black}\cdot \text{\rm tr}(\bm{\Theta}) + \mu\cdot \langle \bm{E}, \bm{\alpha}\rangle}+\rho_1 \cdot \mathrm{tr}(\bm{P})+\rho_2 \cdot \langle \bm{E}, \bm{Z}\rangle \\
    \text{\rm s.t.} \quad & \bm{Y} \circ \bm{Y} \leq \bm{\alpha} \circ \bm{Z},\ \bm{0} \leq \bm{Z} \leq \bm{E}, \ \bm{P} \succeq 0, \ \mathbb{I} - \bm{P} \succeq 0, \ \begin{pmatrix}\bm{\Theta} & \bm{X}\\ \bm{X}^T & \bm{P}\end{pmatrix} \succeq 0.
\end{aligned} \label{opt:penalty_1}
\end{equation}

\begin{equation}
\begin{aligned}
    \min_{\bm{X}, \bm{Y}} \quad \Vert\bm{D} - \bm{X} - \bm{Y}\Vert_F^2  & + \sum_{i \in [n]} \min\left(\sqrt{\rho_1 \lambda}\sigma_i(\bm{X}), \rho_1+\lambda \sigma_i(\bm{X})^2\right)\\
    &+\sum_{i,j \in [n]}\min\left(\sqrt{\mu \rho_2}Y_{i,j}, \rho_2+\mu Y_{i,j}^2\right).
\end{aligned} \label{opt:penalty_2}
\end{equation}
    
\end{proposition}

The above result demonstrates that our regularized relaxation generalizes the reverse Huber penalty \citep[c.f.][]{pilanci2015sparse} to sparse plus low-rank optimization problems. This is quite different from unregularized low-rank problems. Indeed, it follows directly from \citep[Lemma 7]{bertsimas2020mixed} that under a standard big-$M$ assumption on the $\ell_\infty$ norm of the sparse matrix and the spectral norm of the low-rank matrix, an unregularized relaxation of the form
\begin{equation}
\begin{aligned}
    \min_{\bm{X}, \bm{Y}, \bm{Z}, \bm{P} \in \mathbb{R}^{n \times n}} \quad & \Vert\bm{D} - \bm{X} - \bm{Y}\Vert_F^2 +\rho_1 \mathrm{tr}(\bm{P})+\rho_2 \langle \bm{E}, \bm{Z}\rangle \\
    \text{s.t.} \quad & \vert Y_{ij}\vert \leq m Z_{ij} \ \forall i,j \in [n],\ \bm{0} \leq \bm{Z} \leq \bm{E}, \\
    & \bm{P} \succeq 0, \ \mathbb{I} - \bm{P} \succeq 0, \ \begin{pmatrix}M P & \bm{X}\\ \bm{X}^T & M \bm{P}\end{pmatrix} \succeq 0
\end{aligned} \label{opt:convex_relax3}
\end{equation}
is equivalent to the Lasso and nuclear norm regularized problem
\begin{equation}
\begin{aligned}
    \min_{\bm{X}, \bm{Y}, \bm{Z}, \bm{P} \in \mathbb{R}^{n \times n}} \quad & \Vert\bm{D} - \bm{X} - \bm{Y}\Vert_F^2 +\frac{\rho_1}{M} \Vert \bm{X}\Vert_* +\frac{\rho_2}{m}\Vert \bm{Y}\Vert_1.
\end{aligned} \label{opt:convex_relax4}
\end{equation}

Moreover, as demonstrated by \cite{pilanci2015sparse, bertsimas2020sparse} among others, reverse Huber penalties outperform Lasso penalties for sparse regression problems both theoretically—by requiring fewer data to recover the ground truth under a restricted isometry model \cite{pilanci2015sparse}, and empirically—by providing a significantly lower false discovery rate and comparable accuracy rate after observing the same amount of data \cite{bertsimas2020sparse}. This is because Lasso-type penalties are robust estimators but not sparse estimators \citep{bertsimas2018characterization}, while reverse Huber penalties are sparse estimators that recover the ground truth after observing slightly more data than via an exact approach \citep[c.f.][]{askari2022approximation}. Since SLR decomposition is a generalization of sparse regression, this partially explains the superior numerical performance of our alternating minimization method compared to GoDec, as reflected in Section \ref{sec:experiments}.

\section{Branch and Bound} \label{sec:bnb}

{\color{black}
In this section, we propose a branch-and-bound algorithm in the sense of \citep{Land2010, little1966branch} that computes certifiably (near) optimal solutions to Problem \eqref{opt:main_problem} in a practical amount of time. Specifically, we state explicitly our subproblem strategy in Section \ref{ssec:subprroblems}, before stating our overall algorithmic approach in  Section \ref{ssec:overallbnb}. We also provide a sufficient condition for branch-and-bound to obtain a globally optimal solution in Section \ref{ssec:overallbnb}. We remark that branch-and-bound strategies have previously been leveraged for matrix optimization problems \citep{bertsimas2017certifiably, lee2014optimal}.}

Let $h(\bm{Z}, \bm{P})$ {\color{black}denote} the optimal value of the inner minimization problem in \eqref{opt:main_problem_reform}, i.e.:

\begin{equation*}
\begin{aligned}
    h(\bm{Z}, \bm{P}) := \min_{\bm{X}, \bm{Y} \in \mathbb{R}^{n \times n}} \quad & \Vert\bm{D} - \bm{X} - \bm{Y}\Vert_F^2 + \lambda \cdot \Vert\bm{X}\Vert_F^2 + \mu \cdot \Vert\bm{Y}\Vert_F^2 \\
    \text{s.t.} \quad & \bm{X} = \bm{P}\bm{X}, \bm{Y} = \bm{Z} \circ \bm{Y}.
\end{aligned}
\end{equation*} Proposition \ref{prop:main_reform} established that solving \eqref{opt:main_problem} is equivalent to solving $\min_{\bm{Z} \in \mathcal{Z}_{k_1}, \bm{P} \in \mathcal{P}_{k_0}} h(\bm{Z}, \bm{P})$. In Section \ref{sec:sdp_lb}, we illustrated how to obtain a lower bound for the optimal value of \eqref{opt:main_problem} by solving $\min_{\bm{Z} \in \text{conv}(\mathcal{Z}_{k_1}), \bm{P} \in \text{conv}(\mathcal{P}_{k_0})} h(\bm{Z}, \bm{P})$ which we formulated as a semidefinite program in \eqref{opt:convex_relax}. Suppose we wanted to compute a stronger lower bound for \eqref{opt:main_problem}. Two natural {\color{black}Lagrangean relaxations} to consider are: 
\begin{equation}
    \min_{\bm{Z} \in \text{conv}(\mathcal{Z}_{k_1}), \bm{P} \in \mathcal{P}_{k_0}} h(\bm{Z}, \bm{P}), \label{opt:lagrange_Z}
\end{equation}
\begin{equation}
    \min_{\bm{Z} \in \mathcal{Z}_{k_1}, \bm{P} \in \text{conv}(\mathcal{P}_{k_0})} h(\bm{Z}, \bm{P}). \label{opt:lagrange_P}
\end{equation} 

It is not immediately clear which of these two problems produces a stronger lower bound for \eqref{opt:main_problem}. However, as there does not yet exist an efficient method to branch over the set of $n \times n$ orthogonal projection matrices with trace at most $k_0$ \citep{bertsimas2020mixed}, we focus on developing a branch-and-bound algorithm that can solve the second problem, \eqref{opt:lagrange_P}. {\color{black} Moreover, Theorem \ref{thm:AM_opt_fixed} provides sufficient conditions under which we can exactly compute $\min_{\bm{P} \in \mathcal{P}_{k_0}} h(\bm{Z}_0, \bm{P})$ for any fixed $\bm{Z}_0 \in \mathcal{Z}_{k_1}$. Thus, provided these conditions hold, we can solve $\min_{\bm{Z} \in \mathcal{Z}_{k_1}, \bm{P} \in \mathcal{P}_{k_0}} h(\bm{Z}, \bm{P})$ to optimality by branching over the set $\mathcal{Z}_{k_1}$.}

\subsection{Subproblems}\label{ssec:subprroblems}
We construct an enumeration tree that branches on the entries of the binary matrix $\bm{Z}$, which models the sparsity pattern of the sparse matrix $\bm{Y}$. Each node in the tree is defined by a (partial or complete) sparsity pattern, described by collections $\mathcal{I}_0, \mathcal{I}_1 \subset \{(i, j): 1 \leq i, j \leq n\}$ where we have $|\mathcal{I}_0| \leq n^2-k_1$, $|\mathcal{I}_1| \leq k_1$ and $\mathcal{I}_0 \cap \mathcal{I}_1 = \emptyset$, and has an accompanying subproblem. We note that \cite{berk2019certifiably} use a similar notion of partially-determined support when developing a custom branch-and-bound algorithm for the Sparse Principal Component Analysis problem. For indices $(i, j) \in \mathcal{I}_0$, we constrain $Z_{ij} = 0$ and for indices $(i, j) \in \mathcal{I}_1$, we constrain $Z_{ij} = 1$. We say that $\mathcal{I}_0$ and $\mathcal{I}_1$ define a complete sparsity pattern if either $|\mathcal{I}_0| = n^2-k_1$ or $|\mathcal{I}_1| = k_1$, otherwise we say that $\mathcal{I}_0$ and $\mathcal{I}_1$ define a partial sparsity pattern. A terminal node is a node in the tree that can be described by a complete sparsity pattern.

At any given node in the enumeration defined by collections $\mathcal{I}_0$ and $\mathcal{I}_1$, we consider the subproblem given by:

\begin{equation}
\begin{aligned}
    \min_{\bm{X}, \bm{Y} \in \mathbb{R}^{n \times n}} \quad & \Vert\bm{D} - \bm{X} - \bm{Y}\Vert_F^2 + \lambda \cdot \Vert\bm{X}\Vert_F^2 + \mu \cdot \Vert\bm{Y}\Vert_F^2 \\
    \text{s.t.} \quad & \mathrm{Rank}(\bm{X}) \leq k_0, \sum_{(i, j) \, \not\in \, \mathcal{I}_0 \cup \mathcal{I}_1}\mathbbm{1}\{Y_{ij} \neq 0\} \leq k_1 - |\mathcal{I}_1|, \ Y_{ij} = 0 \ \forall (i, j) \in \mathcal{I}_0.
\end{aligned} \label{opt:partial_pattern}
\end{equation} This subproblem can equivalently be expressed as

\begin{equation}
\begin{aligned}
    \min_{\bm{Z} \in \mathcal{Z}_{k_1}, \bm{P} \in \mathcal{P}_{k_0}} \quad & h(\bm{Z}, \bm{P}) \
    \text{s.t.} \ Z_{ij} = 0 \ \forall (i, j) \in \mathcal{I}_0, \ Z_{ij} = 1 \ \forall (i, j) \in \mathcal{I}_1.
\end{aligned} \label{opt:partial_pattern_v2}
\end{equation} Note that if $\mathcal{I}_0 = \mathcal{I}_1 = \emptyset$, \eqref{opt:partial_pattern} and \eqref{opt:partial_pattern_v2} are equivalent to \eqref{opt:main_problem}.

\subsubsection{Subproblem Upper Bound}

We adapt Algorithm \ref{alg:AM} to compute feasible solutions to \eqref{opt:partial_pattern}. Suppose that we fix a sparse matrix $\bm{Y}^*$ in Problem \eqref{opt:partial_pattern}. Then, the problem exactly reduces to \eqref{opt:fixed_Y}, which we know how to solve by Proposition \ref{prop:rank_subproblem}. Suppose we fix a low-rank matrix $\bm{X}^*$ in Problem \eqref{opt:partial_pattern}. Then, the problem becomes:
\begin{equation}
\begin{aligned}
    \min_{\bm{Y} \in \mathbb{R}^{n \times n}} \quad & \Vert\bm{\Tilde{D}} - \bm{Y}\Vert_F^2 + \mu \cdot \Vert\bm{Y}\Vert_F^2 \\
    \text{s.t.} \quad & \sum_{(i, j) \, \not\in \, \mathcal{I}_0 \cup \mathcal{I}_1}\mathbbm{1}\{Y_{ij} \neq 0\} \leq k_1 - |\mathcal{I}_1|, \ Y_{ij} = 0 \ \forall (i, j) \in \mathcal{I}_0.
\end{aligned} \label{opt:fixed_X_partial}
\end{equation} where $\bm{\Tilde{D}} = \bm{D} - \bm{X}^*$ and we have omitted the regularization term on the low-rank matrix because it does not depend on $\bm{Y}$. Similarly to \eqref{opt:fixed_X}, \eqref{opt:fixed_X_partial} admits a closed-form solution:

\begin{proposition}\label{prop:sparsesubproblem_partial}
Let $\bm{Y}^*$ be a matrix such that \[\bm{Y}^* = \bm{S}^* \circ \bigg{(}\frac{\bm{\Tilde{D}}}{1 + \mu}\bigg{)},\] where $\bm{S}^*$ is a $n \times n$ binary matrix with $k_1$ entries $S_{ij}^\star = 1$ such that $S_{ij}^\star = 0 \,\, \forall \, (i, j) \in \mathcal{I}_0, S_{ij}^\star = 1 \,\, \forall \, (i, j) \in \mathcal{I}_1$ and $S_{i,j}^\star \geq S_{k,l}^\star$ if $\vert \Tilde{D}_{i,j}\vert \geq \vert \Tilde{D}_{k,l}\vert \,\, \forall \, (i,j), (k, l) \notin \mathcal{I}_0 \cup \mathcal{I}_1$. Then, $\bm{Y}^\star$ solves Problem \eqref{opt:fixed_X_partial}. 
\end{proposition} Thus, by replacing the update $\bm{Y}_t \xleftarrow[]{} \argmin_{\bm{Y} \in \mathcal{W}} f(\bm{X}_{t-1}, \bm{Y})$ in Algorithm \ref{alg:AM} by the update $\bm{Y}_t \xleftarrow[]{} \argmin_{\bm{Y} \in \Bar{\mathcal{W}}} f(\bm{X}_{t-1}, \bm{Y})$ where $\Bar{\mathcal{W}}=\{\bm{Y} \in \mathbb{R}^{n \times n}: \sum_{ij} \mathbbm{1}\{Y_{ij} \neq 0\} \leq k_1 - |\mathcal{I}_1|, Y_{ij} = 0 \,\, \forall \, (i, j) \in \mathcal{I}_0\}$ using the result of Proposition \ref{prop:sparsesubproblem_partial}, Algorithm \ref{alg:AM} can be readily adapted to obtain high quality feasible solutions to \eqref{opt:partial_pattern}.

\subsubsection{Subproblem Lower Bound}

To obtain a lower bound for the objective value of a subproblem given by \eqref{opt:partial_pattern_v2}, we solve the relaxation given by

\begin{equation}
\begin{aligned}
    \min_{\bm{Z} \in \text{Conv}(\mathcal{Z}_{k_1}),\ \bm{P} \in \text{Conv}(\mathcal{P}_{k_0})} \quad & h(\bm{Z}, \bm{P}) \ \text{s.t.} \ Z_{ij} = 0 \ \forall (i, j) \in \mathcal{I}_0, \ Z_{ij} = 1 \ \forall (i, j) \in \mathcal{I}_1.
\end{aligned} \label{opt:partial_pattern_relax}
\end{equation} From Section \ref{sec:sdp_lb}, it follows that \eqref{opt:partial_pattern_relax} can be expressed as the following semidefinite problem:

\begin{equation}
\begin{aligned}
    \min_{\bm{X}, \bm{Y}, \bm{Z}, \bm{P}, \bm{\Theta}, \bm{\alpha} \in \mathbb{R}^{n \times n}} \quad & \Vert\bm{D} - \bm{X} - \bm{Y}\Vert_F^2 + \lambda \cdot \text{tr}(\bm{\Theta}) + \mu \cdot \text{tr}\langle \bm{E},\bm{\alpha}\rangle \\
    \text{s.t.} \quad & \bm{Y} \circ \bm{Y} \leq \bm{\alpha} \circ \bm{Z},\ \langle \bm{E}, \bm{Z}\rangle \leq k_1, \ \bm{0} \leq \bm{Z} \leq \bm{E}, \\
    & \bm{P} \succeq 0, \ \mathbb{I} - \bm{P} \succeq 0, \ \text{tr}(\bm{P}) \leq k_0, \ \begin{pmatrix}\bm{\Theta} & \bm{X}\\ \bm{X}^T & \bm{P}\end{pmatrix} \succeq 0, \\
    & Z_{ij} = 0 \ \forall (i, j) \in \mathcal{I}_0, \ Z_{ij} = 1 \ \forall (i, j) \in \mathcal{I}_1.
\end{aligned} \label{opt:bnb_problem_relax}
\end{equation}

\subsection{Branch and Bound Algorithm}\label{ssec:overallbnb}

Having specified the subproblem we consider at each node in the tree and how we compute upper bounds (feasible solutions) and lower bounds by leveraging Algorithm \ref{alg:AM} and the convex relaxation given by \eqref{opt:bnb_problem_relax}, it remains to specify the branching rule and the node selection rule. Algorithm \ref{alg:BNB} describes our approach. Branching and node selection rules for branch-and-bound form a rich literature \citep{MORRISON201679}. In our current implementation of Algorithm \ref{alg:BNB}, we employ the most fractional branching rule. Specifically, for an arbitrary non-terminal node $p$, let $\bm{Z}^*$ be the optimal matrix $\bm{Z}$ of the node's convex relaxation given by \eqref{opt:bnb_problem_relax}. We branch on entry $(i^*, j^*) = \argmin_{(i, j) \notin \mathcal{I}_0 \cup \mathcal{I}_1} |Z_{ij} - 0.5|$. When selecting which node to investigate in the tree, we choose a node having a lower bound equal to the current global lower bound. {\color{black} Let $\{(\bm{\Bar{X}}_i, \bm{\Bar{Y}}_i)\}_i$ denote the collection of feasible solutions produced by Algorithm 1 across all nodes that are visited during the execution of Algorithm 2 and let $g(\mathcal{I}_0, \mathcal{I}_1)$ denote the optimal value of Problem \eqref{opt:bnb_problem_relax}. The final upper bound returned by Algorithm 2 is given by $\min_i f(\bm{\Bar{X}}_i, \bm{\Bar{Y}}_i)$, the smallest objective value achieved by the feasible solution returned by Algorithm 1 for any subproblem explored during the execution of Algorithm 2. The final lower bound returned by Algorithm 2 is given by $\min_{(\mathcal{I}_0, \mathcal{I}_1) \in \mathcal{N}} g(\mathcal{I}_0, \mathcal{I}_1)$ where $\mathcal{N}$ denotes the set of nodes that have not been discarded upon the termination of Algorithm 2.}

\begin{theorem}

Algorithm \ref{alg:BNB} terminates in a finite number of iterations and either returns an $\epsilon$ globally optimal solution to \eqref{opt:main_problem} or returns the solution of \eqref{opt:lagrange_P}.

\end{theorem}

\begin{proof}To see that Algorithm \ref{alg:BNB} terminates in a finite number of iterations, it suffices to note that Algorithm \ref{alg:BNB} can never visit a node more than once and that there is a finite number of partial and complete sparsity patterns (each corresponding to a possible tree node) because the set $\mathcal{Z}_{k_1}$ is discrete.

Upon termination, we must have either $\frac{ub-lb}{ub} \leq \epsilon$ or $|\mathcal{N}| = 0$ {\color{black} (or both)}. Suppose that $\frac{ub-lb}{ub} \leq \epsilon$. Then, by definition, the output solution $(\bm{\bar{X}}, \bm{\bar{Y}})$ is $\epsilon$ globally optimal to problem \eqref{opt:main_problem} since $lb$ consists of a global lower bound and $(\bm{\bar{X}}, \bm{\bar{Y}})$ is feasible to \eqref{opt:main_problem}. Suppose instead that $|\mathcal{N}| = 0$. Algorithm \ref{alg:BNB} partitions the space of feasible solutions to \eqref{opt:lagrange_P} and only discards elements of the partition that are guaranteed not to contain the globally optimal solution. If $|\mathcal{N}| = 0$ upon termination, then Algorithm \ref{alg:BNB} has explored (or pruned) the entire space of feasible solutions so the output value $lb$ is the optimal objective of \eqref{opt:lagrange_P}. \end{proof}
{\color{black}
\begin{theorem}
Suppose $\lambda + \frac{2\mu}{1+\mu} - 1 > 0$ and for every full sparsity pattern $\mathcal{I}_0 \subset \{(i, j): 1\leq i, j \leq n\}$, $\Vert \mathcal{I}_0 \Vert = n^2 - k_1$, we have
\[\gamma_{k_0}(\Tilde{\bm{D}}) < \frac{1}{1+\lambda} \Big{(}\lambda + \frac{2\mu}{1+\mu} - 1 \Big{)},\] where $\Tilde{\bm{D}}$ is defined in Theorem \ref{thm:AM_opt_fixed}. Then Algorithm \ref{alg:BNB} returns an $\epsilon$-optimal solution to \eqref{opt:main_problem}.

\end{theorem}

\begin{proof}
Upon termination of Algorithm \ref{alg:BNB}, we must have either $\frac{ub-lb}{ub} \leq \epsilon$ or $|\mathcal{N}| = 0$ {\color{black} (or both)}. Suppose that $\frac{ub-lb}{ub} \leq \epsilon$. Then, by definition, the output solution $(\bm{\bar{X}}, \bm{\bar{Y}})$ is $\epsilon$ globally optimal to problem \eqref{opt:main_problem}. Suppose instead that $|\mathcal{N}| = 0$. {\color{black} Then it must be the case that $ub = lb$. To see this, note that Algorithm \ref{alg:BNB} partitions the space of feasible solutions to \eqref{opt:main_problem} and only discards elements of the partition that are guaranteed not to contain the optimal solution. Moreover, at nodes that correspond to complete sparsity patterns, Theorem \ref{thm:AM_opt_fixed} guarantees that Algorithm \ref{alg:BNB} computes the exact solution of \eqref{opt:full_pattern}. Thus, if $|\mathcal{N}| = 0$ upon termination, Algorithm \ref{alg:BNB} has explored (or pruned) the entire space of feasible solutions so the output value $lb$ is equal to $ub$ and is the optimal objective of \eqref{opt:main_problem}.} \end{proof}
}
\begin{algorithm} \label{alg:BNB}
\SetAlgoLined
\KwData{$\bm{D} \in \mathbb{R}^{n \times n}, \lambda, \mu \in \mathbb{R}^+, k_0, k_1 \in \mathbb{Z}^+$ . Tolerance parameter $\epsilon \geq 0$.}
\KwResult{$(\bm{\bar{X}}, \bm{\bar{Y}})$ that solves \eqref{opt:main_problem} within the optimality tolerance $\epsilon$.}
$p_0 \xleftarrow[]{} (\mathcal{I}_0, \mathcal{I}_1) = (\emptyset, \emptyset)$\;
$\mathcal{N} \xleftarrow[]{} \{p_0\}$\;
$(\bm{\bar{X}}, \bm{\bar{Y}}) \xleftarrow[]{}$ solution returned by Algorithm \ref{alg:AM}\;
$ub \xleftarrow[]{} f(\bm{\bar{X}}, \bm{\bar{Y}})$ \;
$lb \xleftarrow[]{}$ optimal value of \eqref{opt:convex_relax}\;
 \While{$\frac{ub - lb}{ub} > \epsilon$ and $|\mathcal{N}| > 0$}{
  select $(\mathcal{I}_0, \mathcal{I}_1) \in \mathcal{N}$\;
  select some element $(i, j) \not \in \mathcal{I}_0 \cup \mathcal{I}_1$\;
  \For{$k = 0, 1$}{
   $l \xleftarrow[]{} (k + 1) \text{ mod } 2$\;
   newnode $\xleftarrow[]{} \Big{(}\big{(}\mathcal{I}_k \cup (i, j)\big{)}, \mathcal{I}_l \Big{)}$\;
    \emph{upper} $\xleftarrow[]{}$ upperBound(newnode) with feasible point $(\bm{X}^*, \bm{Y}^*)$\;
    \emph{lower} $\xleftarrow[]{}$ lowerBound(newnode)\;
   \If{\emph{upper} $< ub$}{
    $ub \xleftarrow[]{}$ \emph{upper}\;
    $(\bm{\bar{X}}, \bm{\bar{Y}}) \xleftarrow[]{} (\bm{X}^*, \bm{Y}^*)$ \;
    remove any node in $\mathcal{N}$ with \emph{lower} $\geq ub$\;
   }
   \If{\emph{lower} $< ub$}{
    add newnode to $\mathcal{N}$
   }
  }
  remove $(\mathcal{I}_0, \mathcal{I}_1)$ from $\mathcal{N}$\;
  update $lb$ to be the lowest value of \emph{lower} over $\mathcal{N}$\;
}
 \Return{$(\bm{\bar{X}}, \bm{\bar{Y}})$, $lb$}
 \caption{Near-Optimal SLR Decomposition}
\end{algorithm}

\section{Computational Results}\label{sec:experiments}

In this section, we evaluate the performance of our alternating minimization heuristic (Algorithm \ref{alg:AM}) and our branch-and-bound method (Algorithm \ref{alg:BNB}) implemented in Julia 1.5.2 using the \verb|JuMP.jl| package version 0.21.7 and solved using \verb|Mosek| version 9.2 for the semidefinite subproblems \eqref{opt:convex_relax}. We compare our methods against GoDec given by \eqref{opt:goDec}, Stable Principal Component Pursuit (S-PCP) given by \eqref{opt:stable_candes}{\color{black}, Fast RPCA (fRPCA) \citep{yi2016fast} {\color{black}, Accelerated Alternating Projections (AccAltProj) \citep{cai2019accelerated}} and Scaled Gradient Descent (ScaledGD) \citep{tong2021accelerating}}. All experiments were performed using synthetic data{\color{black}, and run on MIT’s Supercloud Cluster \citep{reuther2018interactive}, which hosts Intel Xeon Platinum 8260 processors. The maximum RAM used across all trials was 192GB. To bridge the gap between theory and practice, we have made our code freely available on \url{GitHub} at \url{github.com/NicholasJohnson2020/SparseLowRankSoftware}.} {\color{black} For experiments involving AccAltProj, we employ the MATLAB implementation of the method written by \cite{cai2019accelerated} which is available publicly at \url{https://github.com/caesarcai/AccAltProj_for_RPCA/tree/master}.}

We aim to answer the following questions:

\begin{enumerate}[topsep=0.5ex,itemsep=-0.25ex]
    \item {\color{black}How does the performance of Algorithm \ref{alg:AM} compare to state-of-the-art convex and non-convex methods such as GoDec, S-PCP, {\color{black}AccAltProj, } fRPCA and ScaledGD?}
    \item How does the performance of the accelerated implementation of Algorithm \ref{alg:AM} (described in Section \ref{sec:acc_alg_1}) compare to its exact implementation?
    \item How is the performance of Algorithm \ref{alg:AM} affected by the dimension of the data matrix $\bm{D}$, the signal-to-noise level, the rank of the underlying low-rank matrix, and the sparsity of the underlying sparse matrix?
    \item How does the performance of Algorithm \ref{alg:BNB} compare to Algorithm \ref{alg:AM}?
\end{enumerate}

\subsection{Synthetic Data Generation}

All experiments were performed using synthetic data. To generate a synthetic data matrix $\bm{D}$, we first fix a problem dimension $n$, a desired rank for the low-rank matrix $k_0$, a desired sparsity for the sparse matrix $k_1$ and a value $\sigma > 0$ that controls the signal to noise ratio. Next, we generate a random rank $k_0$ matrix and $k_1$ sparse matrix. To generate the low-rank matrix $\bm{L} \in \mathbb{R}^{n \times n}$, we set $\bm{L} = \bm{V}\bm{V}^T$ where $\bm{V} \in \mathbb{R}^{n \times k_0}$ and $V_{ij} \sim N(0, \frac{\sigma^2}{n})$. To generate the sparse matrix $\bm{S} \in \mathbb{R}^{n \times n}$, we randomly select a symmetric set of indices $\mathcal{S} \subset \{(i, j): 1 \leq i, j \leq n\}$ with cardinality $|\mathcal{S}| = k_1$ and let $S_{ij} \sim U(-5, 5)$ if $(i, j) \in \mathcal{S}$ and $S_{ij} = 0$ otherwise. Finally, we set $\bm{D} = \bm{L} + \bm{S} + \bm{N}$ where $N_{ij} = N_{ji} \sim N(0, 1)$. Note that this data generation process is similar to that employed by \cite{candes2011robust}.

{ \color{black}
\subsection{Hyperparameter Tuning}}

We tune the hyperparameters of Algorithm \ref{alg:AM}, fRPCA, and ScaledGD using $30$-fold cross-validation, as proposed by \cite{validation}. For each fold, we randomly sample $l$ columns and rows from the input data matrix $\bm{D}$ and permute the columns and rows of $\bm{D}$ to obtain $\Tilde{\bm{D}} = \begin{pmatrix}\bm{D}_{val} & \bm{D}_{UR}\\ \bm{D}_{LL} & \bm{D}_{train}\end{pmatrix}$ where $\bm{D}_{val} \in \mathbb{R}^{l \times l}$ is the submatrix corresponding to the randomly sampled rows and columns of $\bm{D}$, $\bm{D}_{train} \in \mathbb{R}^{(n-l) \times (n-l)}$, and $\bm{D}_{UR}, \bm{D}_{LL}^T \in \mathbb{R}^{l \times (n-l)}$. We set $l = \lfloor n \cdot (1-\sqrt{0.7}) \rfloor$ so that the training set $\bm{D}_{train}$ contains at least $70\%$ of the input data. For a given choice of hyperparameters, we perform a SLR decomposition on $\bm{D}_{train}$. Letting $\hat{\bm{X}}$ denote the estimated low-rank matrix, we compute the validation score for a single fold as $\frac{\Vert \bm{D}_{val} - \bm{D}_{UR} \hat{\bm{X}}^\dagger \bm{D}_{LL} \Vert_{\color{black}F}^2}{\Vert \bm{D}_{val} \Vert_{\color{black}F}^2}$. The final validation score for a given set of hyperparameters is the average over $30$ folds.

For experiments reported in Section \ref{sec:exp_valid} and Section \ref{sec:acc_alg_1}, we tune the hyperparameters $(\lambda, \mu)$ for Algorithm \ref{alg:AM} from the collection $\Big{(}\frac{10^{-2}}{\sqrt{n}}, \frac{10^{-1}}{\sqrt{n}}, \frac{10^{0}}{\sqrt{n}}, \frac{10^{1}}{\sqrt{n}}\Big{)} \times \Big{(}\frac{10^{-2}}{\sqrt{n}}, \frac{10^{-1}}{\sqrt{n}}, \frac{10^{0}}{\sqrt{n}}, \frac{10^{1}}{\sqrt{n}}\Big{)}$ and we set the hyperparameter $\gamma = \alpha \frac{k_1}{n^2}$ for fRPCA and ScaledGD where $\alpha$ is tuned (independently for each method) from the collection $(0.01, 0.05, 0.1, 0.5, 1, 2, 4, 6, 8, 10)$. For subsequent experiments in Section \ref{sec:exp_no_valid} and beyond, the hyperparameters of Algorithm \ref{alg:AM}, fRPCA, and ScaledGD are fixed respectively to the best-performing hyperparameters selected via cross-validation in Section \ref{sec:exp_valid} and Section \ref{sec:acc_alg_1}. For experiments employing Algorithm \ref{alg:BNB}, we set $\lambda = \mu = \frac{1}{\sqrt{n}}$. We terminate Algorithm 1, GoDec, fRPCA, and ScaledGD when $\frac{f_{t-1}-f_t}{f_t} < 0.001$ where $f_t$ denotes the objective value achieved by the estimate of the low-rank matrix $\bm{X}$ and the sparse matrix $\bm{Y}$ at iteration $t$.
{\color{black}
\subsection{A Comparison Between the Performance of Algorithm \ref{alg:AM}, GoDec, S-PCP{\color{black}, {\color{black} AccAltProj,} fRPCA and ScaledGD}} \label{sec:exp_valid}
}
We present a comparison of Algorithm \ref{alg:AM}, GoDec, S-PCP{\color{black}, 
  {\color{black} AccAltProj, }fRPCA, and ScaledGD} as we vary the dimension $n$ of the input data matrix $\bm{D}$, the rank $k_0$ of the underlying low-rank matrix $\bm{L}$ and the sparsity level $k_1$ of the underlying sparse matrix $\bm{S}$. We report results for the exact implementations of Algorithm \ref{alg:AM} {\color{black} (``Alg 1 Exact")} and GoDec {\color{black} where the singular value decomposition is computed exactly at each step}. We fix $\sigma = 10$ across all trials. For each value of $(n, k_0, k_1)$, we perform $10$ trials.

In Table \ref{tbl:spcp_godec_alg1}, we report the low-rank matrix reconstruction error (L Error) of each method and the rank and sparsity of the solution returned by S-PCP. Let $\bm{\hat{L}}$ denote the low-rank matrix returned by one of the {\color{black} five} methods. We define the low-rank matrix reconstruction error to be $\frac{\Vert\bm{\hat{L}} - \bm{L}\Vert_F^2}{\Vert\bm{L}\Vert_F^2}$. Let $\bm{\hat{L}}$ and $\bm{\hat{S}}$ denote the low-rank and sparse matrices returned by S-PCP. We define the rank of a solution returned by S-PCP to be $\sum_{i=1}^n \mathbbm{1}\{\sigma_i(\bm{\hat{L}}) > 10^{-2}\}$, the number of singular values of $\bm{\hat{L}}$ that are greater than $10^{-2}$. Similarly, we define the sparsity of a solution returned by S-PCP to be $\sum_{ij} \mathbbm{1}\{\hat{S}_{ij} > 10^{-2}\}$, the number of entries of $\bm{\hat{S}}$ that are greater than $10^{-2}$.

For every parameter configuration explored, Algorithm \ref{alg:AM} outperforms {\color{black} all benchmark methods} by producing a solution that has a {\color{black}comparable although slightly lower low-rank matrix reconstruction error and a lower sparse matrix reconstruction error}. Moreover, the solutions returned by S-PCP always have an average rank that is far greater than the target rank $k_0$ and a sparsity level that is far greater than the target sparsity level $k_1$. Further, the numerical threshold used to compute the rank and sparsity of S-PCP solutions, $10^{-2}$, is quite generous. Indeed, using a more common, more restrictive threshold for numerical tolerance would further amplify this discrepancy. 

In Table \ref{tbl:alg1_exact_accelrated_gap}, we report the low-rank matrix reconstruction error of each method, the bound gap between the solution returned by Algorithm \ref{alg:AM} and the solution of \eqref{opt:convex_relax}, and the time required to solve \eqref{opt:convex_relax}. Letting $\hat{f}$ denote the objective value achieved by the solution returned by Algorithm \ref{alg:AM} and letting $f^*$ denote the optimal value of \eqref{opt:convex_relax}, we define the bound gap as $\frac{\hat{f} - f^*}{\hat{f}}$. Thus, not only does Algorithm \ref{alg:AM} outperform S-PCP, GoDec, {\color{black} fRPCA and ScaledGD}, but, by using the relaxation given by \eqref{opt:convex_relax}, we obtain a certificate of Algorithm \ref{alg:AM}'s instance-wise quality.

{\color{black}
\subsection{An Accelerated Implementation of Algorithm \ref{alg:AM} and its Performance} \label{sec:acc_alg_1}}

As noted in Section \ref{sec:AM}, the main bottleneck in our implementation of Algorithm \ref{alg:AM} is the singular value decomposition step that must be performed at each iteration. One commonly proposed technique in the literature to circumvent this difficulty is to employ a randomized SVD \citep[c.f.][]{halko2011finding}, which computes a low-rank matrix less accurately but in significantly less time than via an exact SVD. Accordingly, in this section, we investigate the use of a randomized SVD in Algorithm \ref{alg:AM} (``Alg \ref{alg:AM} Acc'') against an exact SVD step (``Alg \ref{alg:AM} Exact''). In the accelerated implementation of Algorithm \ref{alg:AM}, we compute a randomized SVD at every iteration except the final one, where we employ an exact SVD.

We now present a comparison of the exact and accelerated implementations of Algorithm \ref{alg:AM} as we vary the dimension $n$ of the input data matrix $\bm{D}$, the rank $k_0$ of the underlying low-rank matrix $\bm{L}$ and the sparsity level $k_1$ of the underlying sparse matrix $\bm{S}$. We fix $\sigma = 10$ across all trials. For each value of $(n, k_0, k_1)$, we performed $10$ trials.

In Table \ref{tbl:alg1_exact_accelerated_time}, we report the low-rank matrix reconstruction error and the execution time of the exact and accelerated implementations of Algorithm \ref{alg:AM}. The execution time reported is the {\color{black} average total runtime of each method which includes the time required to perform cross-validation for the hyperparameters $\lambda$ and $\mu$.} The exact implementation of Algorithm \ref{alg:AM} produces a lower reconstruction error than the accelerated implementation across all trials. This behavior is expected given that at each iteration, the exact implementation of Algorithm \ref{alg:AM} solves the low-rank subproblem \eqref{opt:fixed_Y} to optimality, whereas the accelerated implementation only computes a high-quality solution to this subproblem {\color{black}(except at the last step)}. Further, across all trials, the accelerated implementation of Algorithm \ref{alg:AM} has a faster average execution time than the exact implementation, which is consistent with the $O(n^2 \log k)$ complexity of the low-rank update in the accelerated implementation compared to the $O(n^2 k)$ complexity in the exact implementation.

{\color{black}\subsection{Scalability of Algorithm \ref{alg:AM}}} \label{sec:exp_no_valid}

We present a comparison of Algorithm \ref{alg:AM} with GoDec{\color{black}, AccAltProj} and ScaledGD as we vary the dimension of the input data matrix $\bm{D} \in \mathbb{R}^{n \times n }$. We report results for the exact implementations of Algorithm \ref{alg:AM} and GoDec. For the first experiment, we fixed $k_0 = 5$, $k_1 = 500$, $\sigma = 10$ across all trials, considered values of $n \in \{200, 250, 300, ..., 1000\}$, and performed $50$ trials for each $n$. For the second experiment, we fixed $k_0 = 2$, $k_1 = 500$, $\sigma = 10$, considered values of $n \in \{2000, 4000, ..., 10000\}$, and performed $5$ trials for each $n$. We fixed the hyperparameters $(\lambda, \mu) = \Big{(}\frac{0.1}{\sqrt{n}}, \frac{10}{\sqrt{n}}\Big{)}$ (resp. $\gamma=\frac{k_1}{2n^2}$) for Algorithm \ref{alg:AM} (resp. ScaledGD) for these and all subsequent experiments.

\begin{figure*}[h]\centering
  \includegraphics[width=0.9\textwidth]{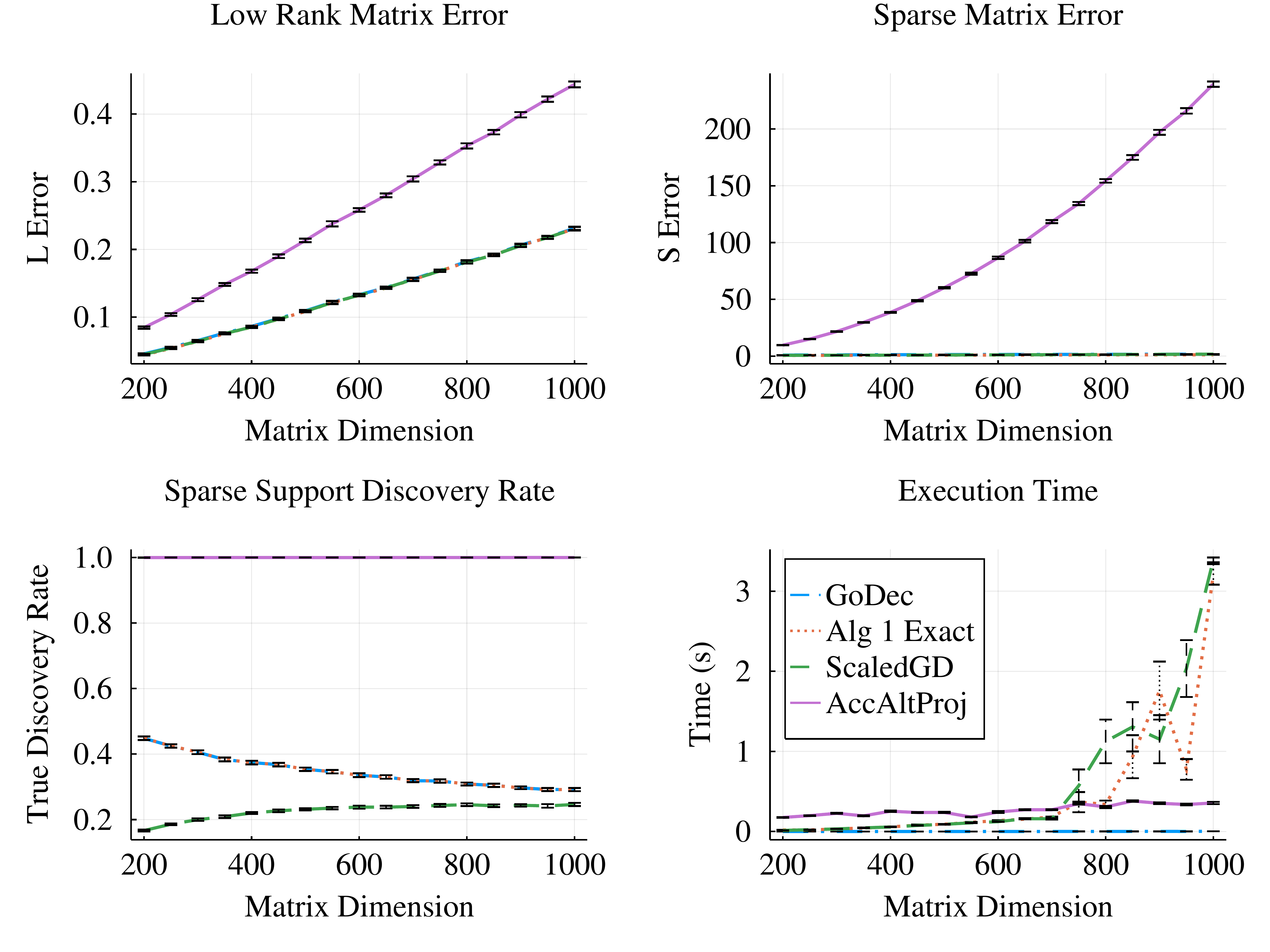}
  \caption{ \color{black}Low-rank matrix reconstruction error (top left), sparse matrix reconstruction error (top right), sparse support discovery rate (bottom left) and execution time (bottom right) versus $n$ with $k_0 = 5$, $k_1 = 500$ and $\sigma = 10$. Averaged over $50$ trials for each parameter configuration.}
  \label{fig:scalabiliy}
\end{figure*}

\begin{figure*}[h]\centering
  \includegraphics[width=0.9\textwidth]{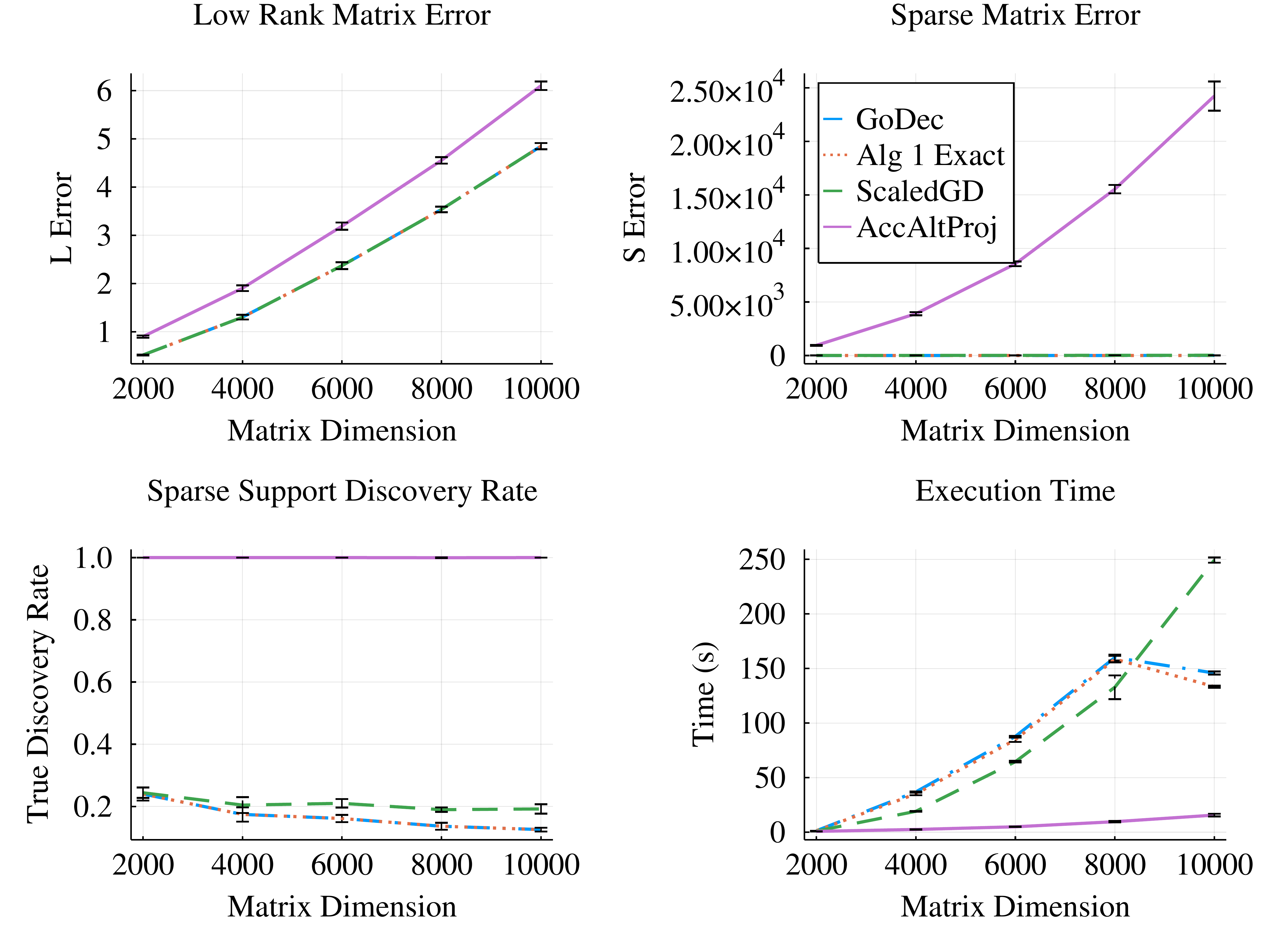}
  \caption{{ \color{black} Low-rank matrix reconstruction error (top left), sparse matrix reconstruction error (top right), sparse support discovery rate (bottom left) and execution time (bottom right) versus $n$ with $k_0 = 2$, $k_1 = 500$ and $\sigma = 10$. Averaged over $5$ trials for each parameter configuration.}}
  \label{fig:scalabiliy_bigN}
\end{figure*}

We report the low-rank matrix reconstruction error, the sparse matrix reconstruction error, the sparse support discovery rate, and the execution time for each method in Figures \ref{fig:scalabiliy}--\ref{fig:scalabiliy_bigN}. We additionally report the low-rank matrix reconstruction error, the sparse matrix reconstruction error and the execution time for {\color{black}Algorithm 1, GoDec and ScaledGD} in Table \ref{tbl:scaling_data} of Appendix \ref{sec:appendix_tables}. Let $\bm{\hat{S}}$ denote the sparse matrix returned by either Algorithm \ref{alg:AM} or GoDec. We define the sparse matrix reconstruction error analogously to the low-rank matrix reconstruction error as $\frac{\Vert\bm{\hat{S}} - \bm{S}\Vert_F^2}{\Vert\bm{S}\Vert_F^2}$. {Let $\mathcal{I}(\bm{S}) = \{(i, j): S_{ij} \neq 0\}$ denote the support of the sparse matrix $\bm{S}$, i.e., the set of indices for which the matrix $\bm{S}$ takes non zero values. Then,} we define the sparse support discovery rate to be $\frac{1}{k_1}\sum_{(i, j) \in \mathcal{I}(\bm{S})}\mathbbm{1}(\hat{S}_{ij} \neq 0)$. The execution time reported is the average runtime for a single trial of a given method. {\color{black}We note that if AccAltProj were implemented in Julia, it would very likely exhibit more favorable runtimes than its publicly available MATLAB implementation \citep{bezanson2017julia}.} The performance metric of greatest interest is the low-rank matrix reconstruction error {\color{black} followed by the sparse matrix reconstruction error}.

Our main findings from this set of experiments are:
\begin{enumerate}
    \item Algorithm \ref{alg:AM} outperforms GoDec{\color{black}, AccAltProj} and ScaledGD across most trials by obtaining lower sparse and low-rank reconstruction errors, while having a comparable execution time. 
    
    
    \item The low-rank matrix reconstruction error scales linearly with matrix dimension for Algorithm \ref{alg:AM}, {\color{black} AccAltProj,} ScaledGD, and GoDec. It can be shown that for our data generation process, $\lim_{n \to \infty} \mathbb{E}[\Vert \bm{L} \Vert_F^2] = C(k_0, \sigma)$ where $C(k_0, \sigma)$ is a constant that depends only on the rank of $\bm{L}$ and the signal-to-noise level. This implies that for all methods, $\mathbb{E}[\Vert\bm{\hat{L}} - \bm{L}\Vert_F^2]$ is $\Theta (n)$.
    
    \item The sparse matrix reconstruction error appears to scale linearly with matrix dimension for Algorithm \ref{alg:AM}, ScaledGD, and GoDec, {\color{black} while scaling superlinearly with the matrix dimension for AccAltProj. Note that AccAltProj does not allow the cardinality of the sparse matrix to be explicitly constrained. Accordingly, AccAltProj tends to return a sparse matrix that is considerably denser than the desired level. This produces a high sparse support discovery rate (true positive rate) at the expense of a high false discovery rate.}. The sparse support discovery rate declines as the matrix dimension increases for {\color{black} GoDec and Algorithm 1}in the regime investigated in Figure \ref{fig:scalabiliy}. ScaledGD underperforms {\color{black} GoDec and Algorithm 1} with respect to sparse support discovery rate in low-dimensional settings (Figure \ref{fig:scalabiliy}) but outperforms in high-dimensional settings (Figure \ref{fig:scalabiliy_bigN}). This is to be expected as with increasing matrix dimension while $k_1$ is held fixed, it becomes increasingly difficult to identify the underlying sparsity pattern.
    
\end{enumerate}

\subsection{Sensitivity to Noise}

We present a comparison of Algorithm \ref{alg:AM} with GoDec{\color{black}, AccAltProj} {\color{black}and ScaledGD} as we vary the signal to noise level $\sigma$ of the input data matrix $\bm{D}$. Large values of $\sigma$ correspond to a greater signal in the low-rank matrix $\bm{L}$ compared to the perturbation matrix $\bm{N}$. We report results for the exact implementations of Algorithm \ref{alg:AM} and GoDec that exactly compute the singular value decomposition step. We fixed $n=100$, $k_0 = 5$, $k_1 = 500$ across all trials and considered values of $\sigma \in \{1, 2, 3, ..., 30\}$. For each value of $\sigma$, we performed $50$ trials.

\begin{figure*}[h]\centering
  \includegraphics[width=0.9\textwidth]{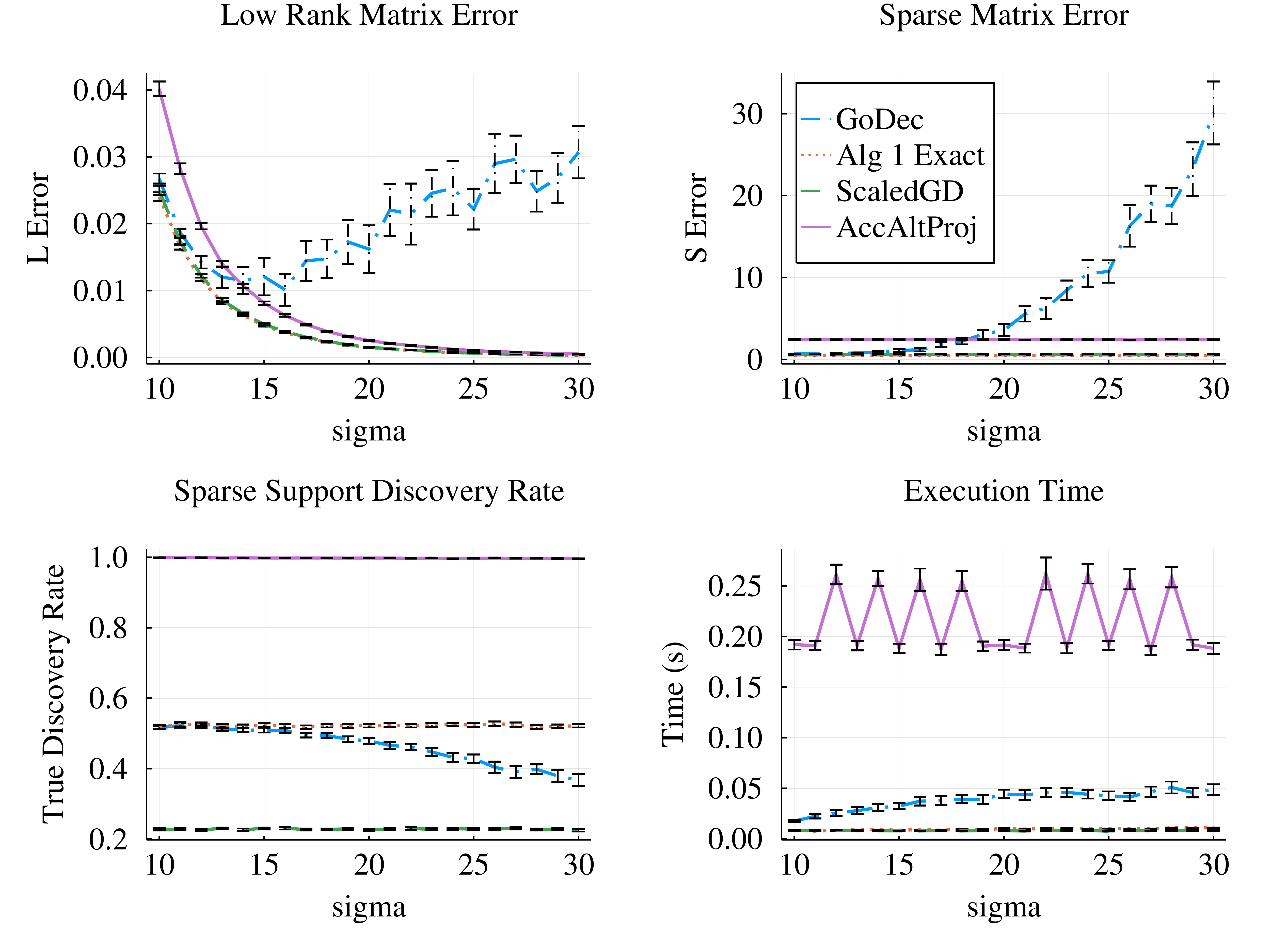}
  \caption{ \color{black} Low-rank matrix reconstruction error (top left), sparse matrix reconstruction error (top right), sparse support discovery rate (bottom left) and execution time (bottom right) versus $\sigma$ with $n=100$, $k_0 = 5$ and $k_1 = 500$. Averaged over $50$ trials for each parameter configuration.}
  \label{fig:noise}
\end{figure*}

We report the low-rank matrix reconstruction error, the sparse matrix reconstruction error, the sparse support discovery rate, and the execution time for each method in Figure \ref{fig:noise}. {\color{black} Figure \ref{fig:noise} includes only results for values of $\sigma \in [10, 30]$ to aid visualization due to significant differences in scale between these results and those for $\sigma \in [1, 10]$. We report the results for the full range $\sigma \in [1, 30]$ in Figure \ref{fig:noise_full} of Appendix \ref{sec:appendix_tables}.}

Our main findings from this set of experiments are:
\begin{enumerate}[topsep=0.5ex,itemsep=-0.25ex]
    \item Consistent with previous experiments, Algorithm \ref{alg:AM} outperforms GoDec{\color{black}, AccAltProj } {\color{black}and ScaledGD} across {\color{black} most} trials by obtaining a lower sparse and low-rank matrix reconstruction error while {\color{black}maintaining} a {\color{black} comparable} execution time and exhibiting superior sparse support discovery rates {\color{black} (compared to GoDec and ScaledGD)}. The superior performance of Algorithm \ref{alg:AM} {\color{black}relative to GoDec} becomes more extreme as the signal-to-noise ratio increases.
    \item The low-rank reconstruction error of Algorithm \ref{alg:AM} decreases as $\sigma$ increases. This is consistent with the intuition that larger values of $\sigma$ correspond to easier problem instances, so it should be easier to recover the low-rank matrix. Further, the plotted trend suggests that should $\sigma$ be further increased, Algorithm \ref{alg:AM} would exactly recover $\bm{L}$. Somewhat surprisingly, the performance of GoDec appears to break down at higher levels of $\sigma$. The sparse matrix reconstruction error of Algorithm \ref{alg:AM} also declines as $\sigma$ increases, whereas that of GoDec again breaks down. {\color{black} ScaledGD exhibits a poor sparse recovery rate in these experiments.}
    \item The sparse support discovery rate of Algorithm \ref{alg:AM} slightly declines as $\sigma$ increases, whereas that of GoDec drops sharply. Though one might expect the sparse support discovery rate to increase with the signal-to-noise level, recall that $\sigma$ controls the signal-to-noise level of the low-rank matrix compared to the noise matrix and not that of the sparse matrix. Consequently, as $\sigma$ increases, it should become easier to recover the low-rank matrix but more difficult to recover the sparse matrix.
\end{enumerate}

\subsection{Sensitivity to Rank}

We present a comparison of Algorithm \ref{alg:AM} with GoDec{\color{black}, AccAltProj} {\color{black}and ScaledGD} as we vary the rank $k_0$ of the underlying low-rank matrix $\bm{L}$. We report results for the exact implementations of Algorithm \ref{alg:AM} and GoDec that exactly compute the singular value decomposition step. We fixed $n=100$, $k_1 = 500$, $\sigma=10$ across all trials and considered values of $k_0 \in \{2, 4, 6, ..., 50\}$. For each value of $k_0$, we performed $50$ trials.

\begin{figure*}[h]\centering
  \includegraphics[width=0.9\textwidth]{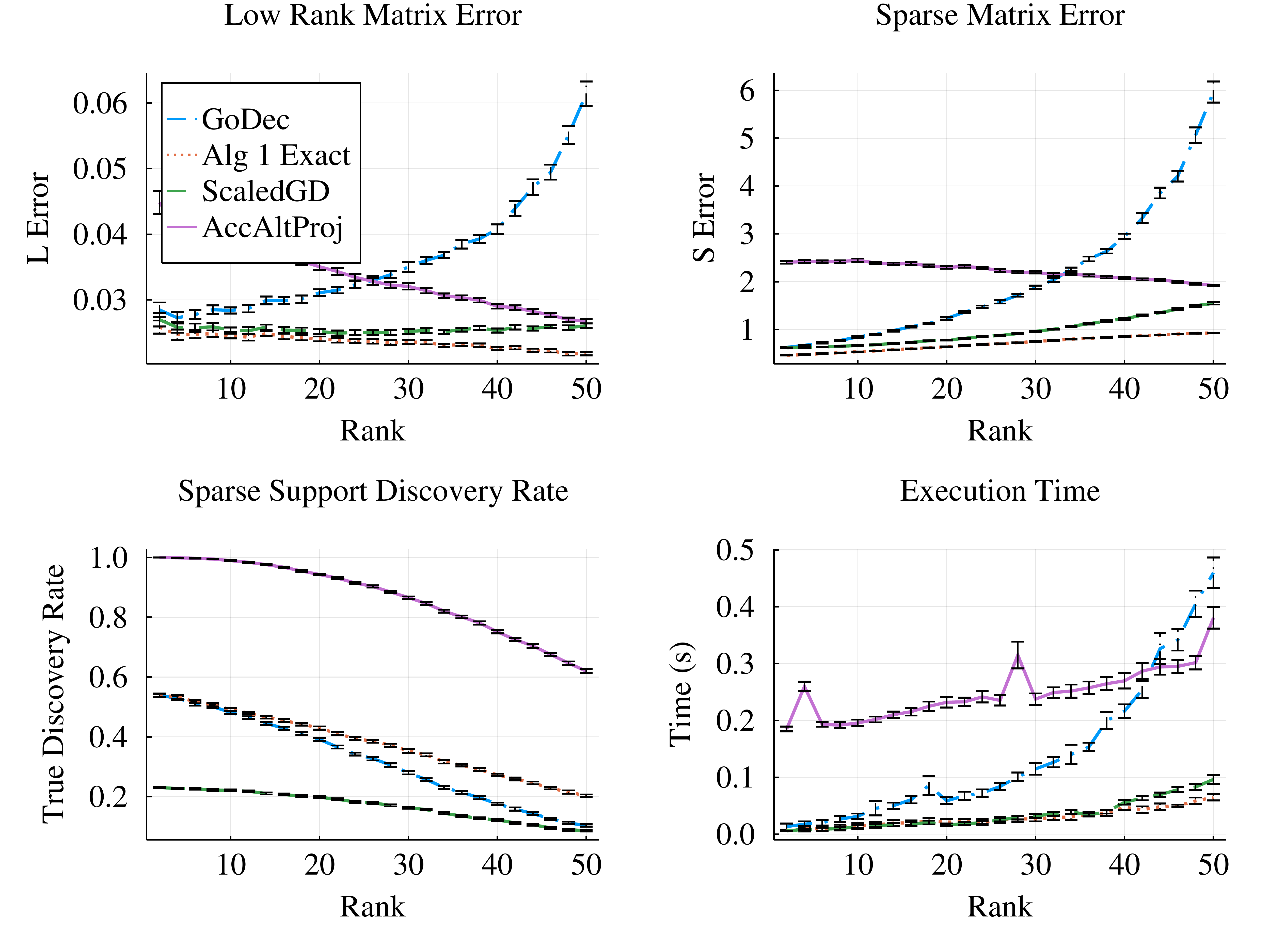}
  \caption{ \color{black}Low-rank matrix reconstruction error (top left), sparse matrix reconstruction error (top right), sparse support discovery rate (bottom left) and execution time (bottom right) versus $k_0$ with $n=100$, $k_1 = 500$ and $\sigma=10$. Averaged over $50$ trials for each parameter configuration.}
  \label{fig:rank}
\end{figure*}

We report the low-rank matrix reconstruction error, the sparse matrix reconstruction error, the sparse support discovery rate, and the runtime for each method in Figure \ref{fig:rank}.

Our main findings from this set of experiments are:
\begin{enumerate}[topsep=0.5ex,itemsep=-0.25ex]
    \item Consistent with previous experiments, Algorithm \ref{alg:AM} outperforms GoDec{\color{black}, AccAltProj} {\color{black} and ScaledGD} across all trials by obtaining a lower low-rank matrix reconstruction error and sparse matrix reconstruction error while having a lesser {\color{black}(in the case of GoDec {\color{black} and AccAltProj}) or comparable (in the case of ScaledGD)} execution time and exhibiting superior sparse support discovery rates {\color{black} than GoDec and ScaledGD}. The superior performance of Algorithm \ref{alg:AM} becomes more extreme as the rank increases.
    \item The low-rank reconstruction error of Algorithm \ref{alg:AM} {\color{black}and that of AccAltProj} decrease as $k_0$ increases whereas the low-rank reconstruction error of GoDec increases with increasing $k_0$ and {\color{black}that of ScaledGD remains roughly constant}.
    \item {\color{black} Algorithm \ref{alg:AM}'s and ScaledGD's sparse matrix reconstruction error increases slightly, while GoDec's error increases significantly and AccAltProj's decreases slightly}.
\end{enumerate}

\subsection{Sensitivity to Sparsity}

We present a comparison of Algorithm \ref{alg:AM} with GoDec{\color{black}, AccAltProj} {\color{black}and ScaledGD} as we vary the sparsity level $k_1$ of the underlying sparse matrix $\bm{S}$. We report results for the exact implementations of Algorithm \ref{alg:AM} and GoDec that exactly compute the singular value decomposition step. We fixed $n=100$, $k_0 = 5$, $\sigma = 10$ across all trials and considered values of $k_1 \in \{50, 100, 150, ..., 1000\}$. For each value of $k_1$, we performed $50$ trials.

\begin{figure*}[h]\centering
  \includegraphics[width=0.9\textwidth]{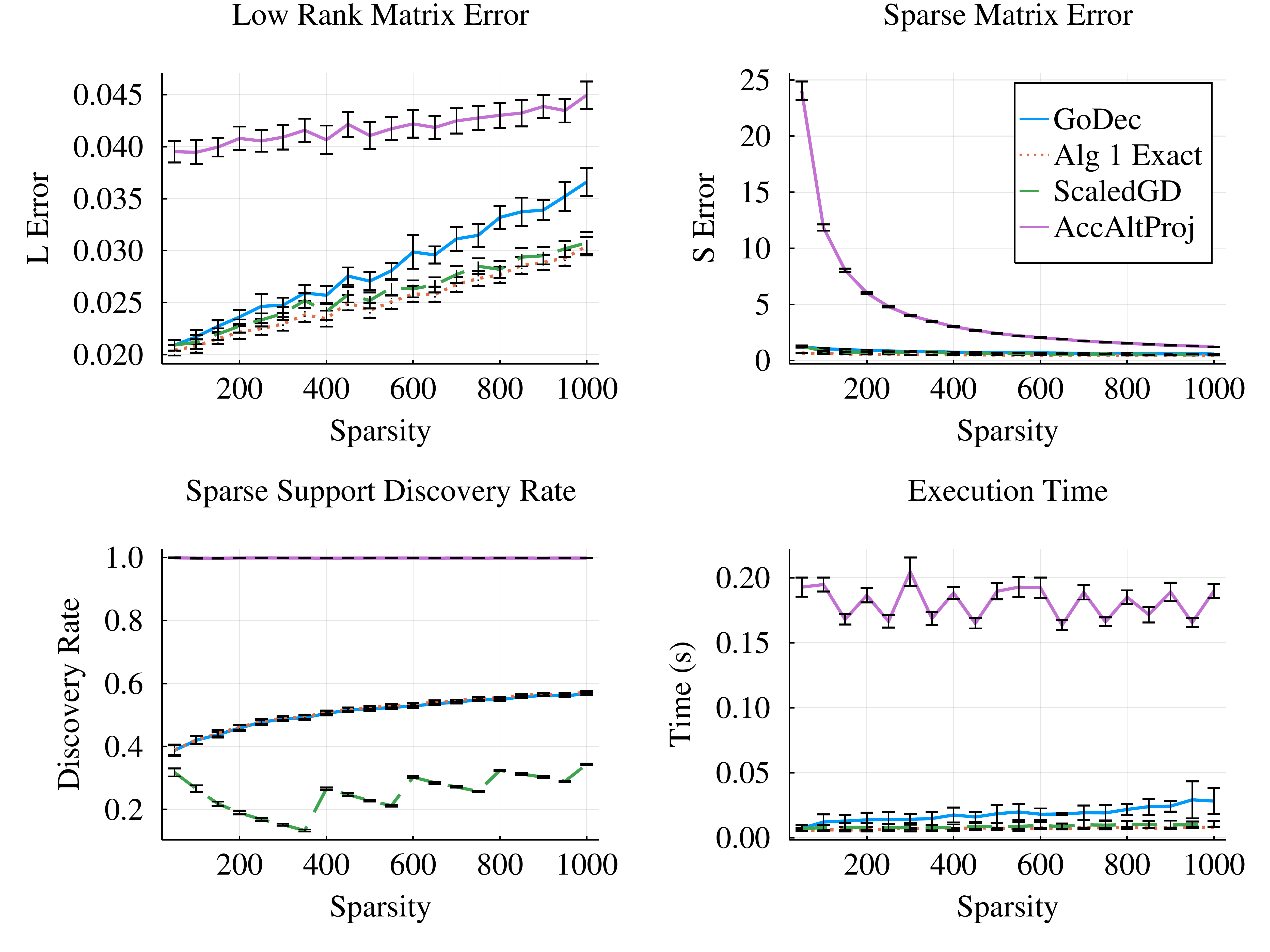}
  \caption{ \color{black} Low-rank matrix reconstruction error (top left), sparse matrix reconstruction error (top right), sparse support discovery rate (bottom left) and execution time (bottom right) versus $k_1$ with $n=100$, $k_0 = 5$ and $\sigma = 10$. Averaged over $50$ trials for each parameter configuration.}
  \label{fig:sparse}
\end{figure*}

We report the low-rank matrix reconstruction error, the sparse matrix reconstruction error, the sparse support discovery rate, and the runtime for each method in Figure \ref{fig:sparse}.

Our main findings from this set of experiments are:
\begin{enumerate}[topsep=0pt,itemsep=-0.25ex]
    \item Consistent with previous experiments, Algorithm \ref{alg:AM} outperforms GoDec{\color{black}, AccAltProj}  {\color{black}and ScaledGD} across all trials by obtaining a lower low-rank matrix reconstruction error and sparse matrix reconstruction error while having a lesser execution time{\color{black}. Algorithm \ref{alg:AM} also exhibits a superior accuracy rate than GoDec and ScaledGD.}
    \item The low-rank reconstruction error of Algorithm \ref{alg:AM}{\color{black}, GoDec{\color{black}, AccAltProj} and ScaledGD} increase as $k_1$ increases. This is consistent with the intuition that as the sparsity of the underlying spare matrix increases, it becomes more difficult to identify the true low-rank matrix.
    \item The sparse matrix reconstruction error of Algorithm \ref{alg:AM}{\color{black}, ScaledGD}{\color{black}, AccAltProj} and GoDec decline as $k_1$ increases.
\end{enumerate}

{\color{black}\subsection{Performance of Algorithm \ref{alg:BNB}}}

We report the performance of Algorithm \ref{alg:BNB} on several problem instances. In these experiments, calls that Algorithm \ref{alg:BNB} make to Algorithm \ref{alg:AM} employ the exact implementation of Algorithm \ref{alg:AM}. We fix $\sigma = 10$ and set $\epsilon = 0.05$, meaning that Algorithm \ref{alg:AM} terminates when it has computed a solution to \eqref{opt:main_problem} that is certifiably within $5\%$ of the globally optimal solution. We report the optimality gap between the root node upper bound and the root node lower bound, the total number of nodes explored, and the execution time of Algorithm \ref{alg:BNB} for $14$ problem instances in Table \ref{tbl:BNB}.

\begin{table}
  \centering
  \caption{Performance of Algorithm \ref{alg:BNB} for $\epsilon = 0.05$. Reported root node gap is a percentage.}\label{tbl:BNB}
  \begin{tabular}{ccc || ccc}
\toprule
 N &  $k_0$ &  $k_1$ &  Root Node Gap &  Nodes Explored &  Time (s) \\
\midrule
10 &      1 &     10 &           5.66 &               3 &        41 \\
10 &      1 &     15 &           2.94 &               1 &        43 \\
10 &      2 &     20 &           2.37 &               1 &        43 \\
15 &      1 &     22 &           7.34 &              33 &        58 \\
15 &      2 &     33 &           5.08 &               3 &        47 \\
15 &      3 &     45 &           3.26 &               1 &        40 \\
20 &      1 &     20 &           5.48 &               5 &        44 \\
20 &      2 &     40 &           6.44 &             123 &       126 \\
20 &      3 &     60 &           4.33 &               1 &        40 \\
20 &      4 &     80 &           4.15 &               1 &        41 \\
25 &      1 &     31 &           7.43 &             205 &       479 \\
25 &      2 &     62 &           8.30 &           14709 &     28977 \\
25 &      3 &     93 &           6.60 &            1053 &      2485 \\
25 &      5 &    125 &           7.50 &             653 &      1631 \\
\bottomrule
\end{tabular}

\end{table}

As expected, when the root node optimality gap is less than $\epsilon$, no additional nodes are explored. The total number of possible terminal nodes in any branch-and-bound instance is equal to the number of distinct sparsity patterns, given by ${n^2 \choose k_1}$. This implies that the total number of possible nodes in any branch-and-bound instance is given by $2 \cdot {n^2 \choose k_1} - 1$. In the case of the last instance given in Table \ref{tbl:BNB}, this quantity is roughly equal to $5.3 \times 10^{134}$. Thus, the results of Table \ref{tbl:BNB} indicate that Algorithm \ref{alg:BNB} is able to prune the vast majority of possible nodes in the branch-and-bound tree. We note that the execution time explodes as the number of nodes explored increases. One of the main limitations of the current implementation of Algorithm \ref{alg:BNB} is that it requires solving \eqref{opt:bnb_problem_relax}, a semidefinite optimization problem, at every node that is explored. This becomes a computational bottleneck as the most efficient interior point solvers for SDPs exhibit poor scaling.

\begin{figure*}[h]\centering
  \includegraphics[width=0.9\textwidth]{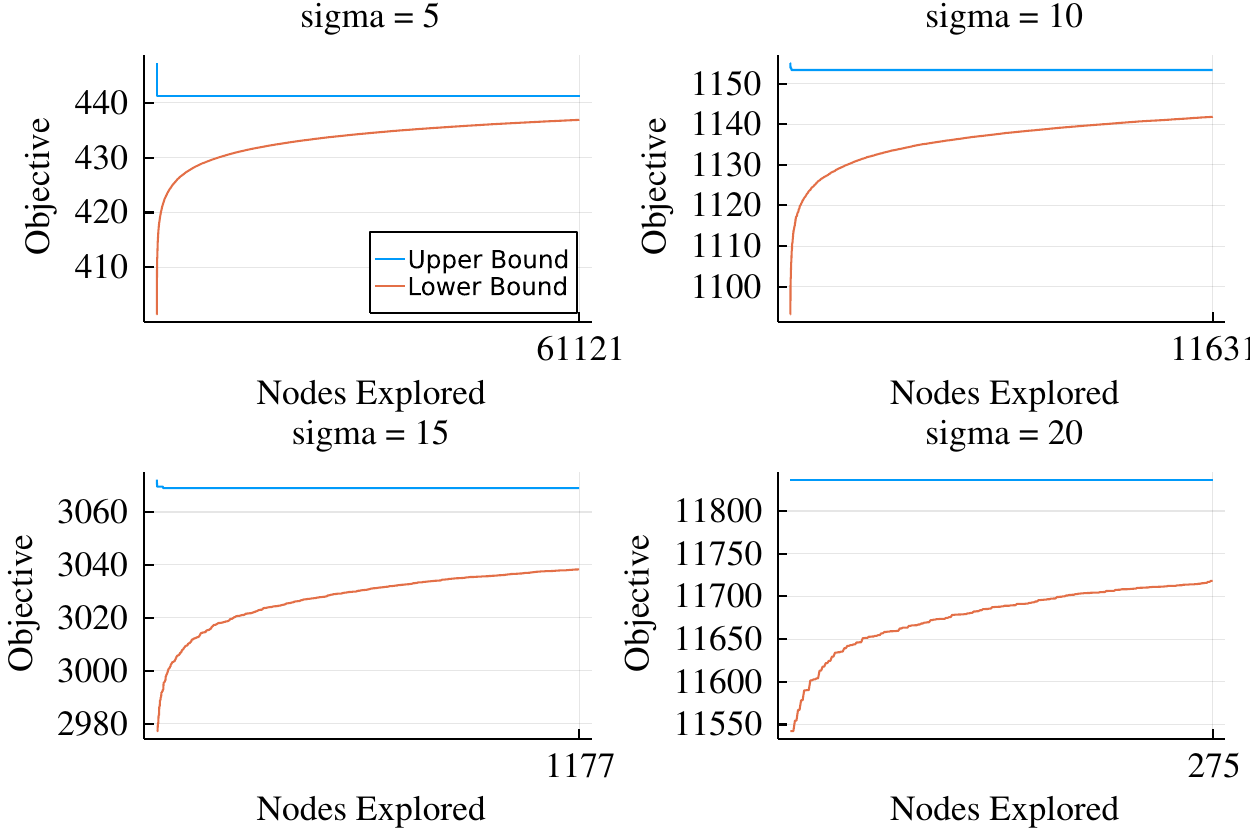}
  \caption{{\color{black} Algorithm \ref{alg:BNB} upper and lower bound evolution (for a single instance) for $\sigma=5$ (top left), $\sigma=10$ (top right), $\sigma=15$ (bottom left) and $\sigma=20$ (bottom right) with $n=15$, $k_0 = 1$, $k_1 = 22$ and $\epsilon = 0.01$.}}
  \label{fig:bound}
\end{figure*}

\begin{figure*}[h]
    \centering
  \includegraphics[width=0.9\textwidth]{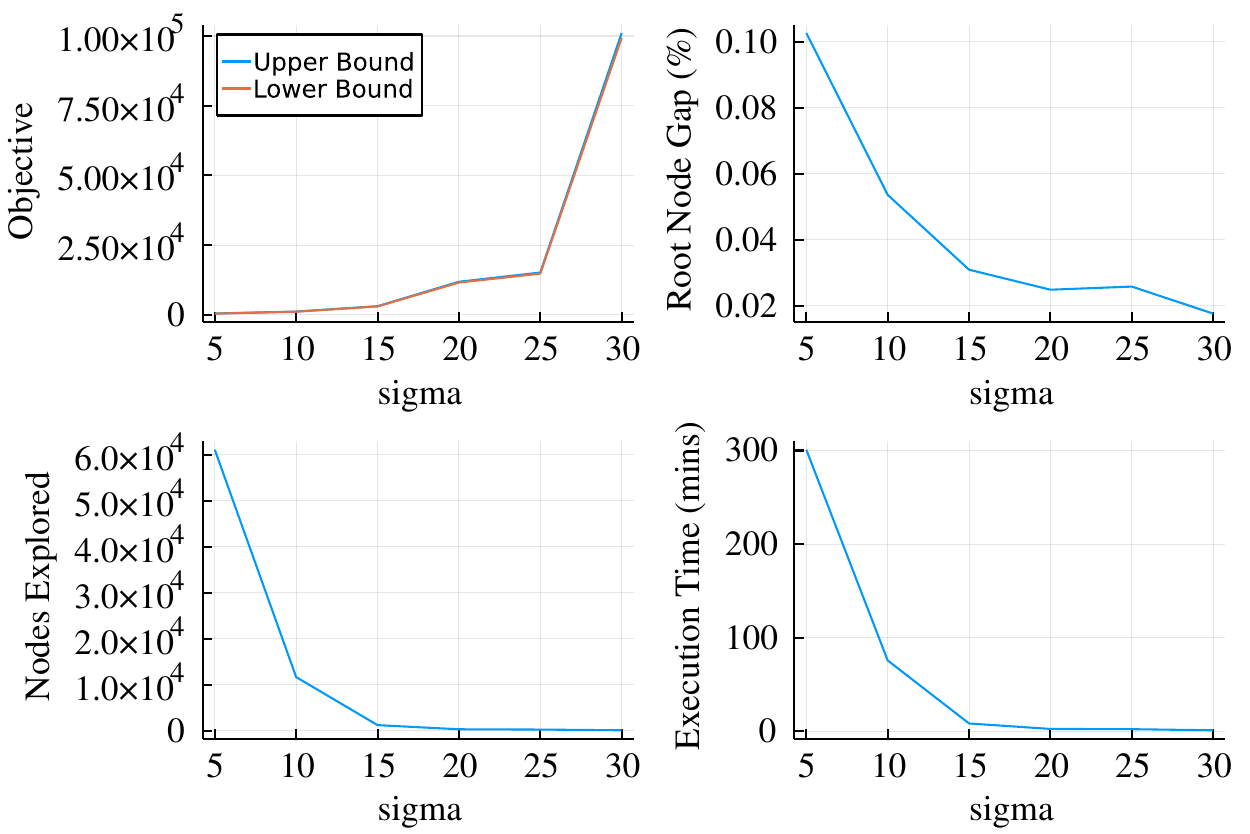}
  \caption{{\color{black} Algorithm \ref{alg:BNB} root node upper and lower bound (top left), root node optimality gap (top right), number of nodes explored (bottom left) and execution time (bottom right) versus $\sigma$ with $n=15$, $k_0 = 1$, $k_1 = 50$ and $\epsilon = 0.01$.}}
  \label{fig:bound_revised}
\end{figure*}

Figure \ref{fig:bound} illustrates that Algorithm \ref{alg:BNB} only occasionally updates the global upper bound and that the vast majority of computational time is spent certifying optimality. This behavior is consistent across all problem instances in which the root node upper bound is not already $\epsilon$ optimal. Moreover, Figure \ref{fig:bound_revised} illustrates that Algorithm \ref{alg:BNB} successfully solves instances where $n=15$ for all values of $\sigma$, and is fastest when there is the least amount of noise.

\subsection{Summary of Findings From Numerical Experiments}

We are now in a position to answer the {\color{black}four} questions introduced at the start of this section. Our findings are as follows:

\begin{enumerate}[topsep=0.5ex,itemsep=-0.25ex]
    \item Algorithm \ref{alg:AM} outperforms GoDec across all trials by obtaining a lower low-rank matrix reconstruction error and sparse matrix reconstruction error while having a lesser execution time and exhibiting superior sparse support discovery rates. The superior performance of Algorithm \ref{alg:AM} is most extreme in regimes where the signal-to-noise level $\sigma$ is high and separately when the rank $k_0$ of the underlying low-rank matrix is high. Further, Algorithm \ref{alg:AM} outperforms S-PCP{\color{black}, AccAltProj} {\color{black}and fRPCA} across all trials by obtaining lower low-rank {\color{black}and sparse matrix reconstruction errors.} {\color{black}With cross-validation, Algorithm \ref{alg:AM} obtains low-rank matrices with a lower rank and a comparable reconstruction error than ScaledGD, and with a rank constraint on both methods it obtains a lower low-rank error that ScaledGD on all but 3 trials. Moreover, it always achieves a lesser sparse matrix reconstruction error than ScaledGD.}
    \item The exact implementation of Algorithm \ref{alg:AM} outperforms the accelerated implementation by achieving a lower reconstruction error across all trials. However, across all trials, the accelerated implementation of Algorithm \ref{alg:AM} has a faster average execution time than the exact implementation.
    \item \begin{enumerate}[topsep=0.5ex,itemsep=-0.25ex]
        \item Increasing the matrix dimension $n$ results in linear increases in the low-rank matrix reconstruction error and the sparse matrix reconstruction error for Algorithm \ref{alg:AM}, GoDec {\color{black} and ScaledGD}. {\color{black}Increasing the matrix dimension $n$ results in a linear increase in the low-rank matrix reconstruction error and a superlinear increase in the sparse matrix reconstruction error for AccAltProj.} The sparse support discovery rate decreases with $n$ {\color{black}for Algorithm \ref{alg:AM} and GoDec} while the execution time of each method scales {\color{black}superlinearly} with $n$.
         \item The low-rank matrix and sparse matrix reconstruction errors of Algorithm \ref{alg:AM}{\color{black}, AccAltProj} {\color{black}and ScaledGD} decrease with increasing values of $\sigma$ and that of Algorithm \ref{alg:AM} appears to converge towards $0$. The sparse support discovery rate of Algorithm \ref{alg:AM} decreases slightly with $\sigma$ while its execution time remains roughly constant. Conversely, the low-rank matrix and sparse matrix reconstruction errors of GoDec explode for large values of $\sigma$. GoDec's sparse support discovery rate declines sharply in the high signal-to-noise level regime. {\color{black}ScaledGD generally has poor sparse support discovery.}{\color{black}AccAltProj tends to exhibit high sparse support discovery rate because the sparse matrix selected by AccAltProj is in general substantially more dense than the ground truth sparse matrix.}
        \item {\color{black} Increasing the rank of the low-rank matrix results in a slight decrease in the low-rank matrix reconstruction error and a slight increase in the sparse matrix reconstruction error for Algorithm \ref{alg:AM} and ScaledGD. In contrast, the low-rank matrix and sparse matrix reconstruction errors grow superlinearly for GoDec with increasing rank. The sparse support discovery rate {\color{black}, of Algorithm \ref{alg:AM}, GoDec and ScaledGD}, and the execution time of all methods grow with increasing rank.}
        \item Algorithm \ref{alg:AM}{\color{black}, ScaledGD} and GoDec exhibit similar behaviour as a function of sparsity $k_1$. As the sparsity level of the underlying sparse matrix increases, the low-rank matrix reconstruction error, sparse support discovery rate, and execution time of each {\color{black} of these }methods increase while the sparse matrix reconstruction error decreases.
    \end{enumerate} 
    \item Algorithm \ref{alg:BNB} solves \eqref{opt:main_problem} to certifiable {\color{black} optimality} for small problem instances (up to $n=25$) in reasonable wall clock time. The majority of Algorithm \ref{alg:BNB}'s execution time is spent certifying {\color{black} optimality}. This implies that the final solution returned by Algorithm \ref{alg:BNB} is, in general, only marginally better than the solution returned by Algorithm \ref{alg:AM}.
\end{enumerate}

\section{Conclusion} \label{sec:conclusion}

In this paper, we introduced a novel formulation \eqref{opt:main_problem} for SLR that exploits discreteness and leverages regularization. We presented Algorithm \ref{alg:AM}, an alternating minimization heuristic that can compute high-quality feasible solutions to \eqref{opt:main_problem} and can scale to $n=10000$ in {\color{black}minutes}. We developed a strong semidefinite relaxation \eqref{opt:convex_relax} that can certify the quality of the solutions returned by Algorithm \ref{alg:AM}. Finally, we presented Algorithm \ref{alg:BNB}, a branch-and-bound method that solves \eqref{opt:main_problem} to certifiable near-optimality and scales to $n=25$ in minutes. {\color{black} Moreover, we established sufficient conditions under which Algorithm \ref{alg:BNB} is optimal.} Further work could focus on increasing the scalability of our branch-and-bound method. When executing Algorithm \ref{alg:BNB}, a semidefinite optimization problem must be solved at every node in the branch-and-bound tree to compute a lower bound. This computation is quite costly. A possible extension would be to compute a second-order cone lower bound at each node which would be more scalable at the expense of being less tight. Algorithm \ref{alg:BNB} can also {\color{black} potentially} be further improved by adopting an alternate branch{\color{black}ing} rule.

{\color{black}
\subsection*{Acknowledgements}
We are very grateful to three anonymous referees for their insightful and helpful comments that improved the paper significantly.
}

{\footnotesize
\bibliography{biblio.bib}

\begin{thebibliography}{64}
\providecommand{\natexlab}[1]{#1}
\providecommand{\url}[1]{\texttt{#1}}
\expandafter\ifx\csname urlstyle\endcsname\relax
  \providecommand{\doi}[1]{doi: #1}\else
  \providecommand{\doi}{doi: \begingroup \urlstyle{rm}\Url}\fi

\bibitem[Arous et~al.(2020)Arous, Wein, and Zadik]{arous2020free}
G{\'e}rard~Ben Arous, Alexander~S Wein, and Ilias Zadik.
\newblock Free energy wells and overlap gap property in sparse {PCA}.
\newblock In \emph{Conference on Learning Theory}, pages 479--482. PMLR, 2020.

\bibitem[Askari et~al.(2022)Askari, d'Aspremont, and
  Ghaoui]{askari2022approximation}
Armin Askari, Alexandre d'Aspremont, and Laurent~El Ghaoui.
\newblock Approximation bounds for sparse programs.
\newblock \emph{SIAM Journal on Mathematics of Data Science}, 4\penalty0
  (2):\penalty0 514--530, 2022.

\bibitem[Basu et~al.(2019)Basu, Li, and Michailidis]{Basu_2019}
Sumanta Basu, Xianqi Li, and George Michailidis.
\newblock Low rank and structured modeling of high-dimensional vector
  autoregressions.
\newblock \emph{IEEE Transactions on Signal Processing}, 67\penalty0
  (5):\penalty0 1207–1222, 2019.

\bibitem[Ben-Tal and Den~Hertog(2014)]{ben2014hidden}
Aharon Ben-Tal and Dick Den~Hertog.
\newblock Hidden conic quadratic representation of some nonconvex quadratic
  optimization problems.
\newblock \emph{Mathematical Programming}, 143\penalty0 (1):\penalty0 1--29,
  2014.

\bibitem[Berk and Bertsimas(2019)]{berk2019certifiably}
Lauren Berk and Dimitris Bertsimas.
\newblock Certifiably optimal sparse principal component analysis.
\newblock \emph{Mathematical Programming Computation}, 11\penalty0
  (3):\penalty0 381--420, 2019.

\bibitem[Bertsekas(2016)]{bertsekas1999nonlinear}
Dimitri~P Bertsekas.
\newblock \emph{Nonlinear programming}.
\newblock Athena Scientific Belmont MA, 3rd edition, 2016.

\bibitem[Bertsimas and Copenhaver(2018)]{bertsimas2018characterization}
Dimitris Bertsimas and Martin~S Copenhaver.
\newblock Characterization of the equivalence of robustification and
  regularization in linear and matrix regression.
\newblock \emph{European Journal of Operational Research}, 270\penalty0
  (3):\penalty0 931--942, 2018.

\bibitem[Bertsimas and den Hertog(2020)]{bertsimas2020robust}
Dimitris Bertsimas and Dick den Hertog.
\newblock \emph{Robust and adaptive optimization}.
\newblock Dynamic Ideas LLC, 2020.

\bibitem[Bertsimas et~al.(2017)Bertsimas, Copenhaver, and
  Mazumder]{bertsimas2017certifiably}
Dimitris Bertsimas, Martin~S Copenhaver, and Rahul Mazumder.
\newblock Certifiably optimal low rank factor analysis.
\newblock \emph{Journal of Machine Learning Research}, 18\penalty0
  (1):\penalty0 907--959, 2017.

\bibitem[Bertsimas et~al.(2020)Bertsimas, Pauphilet, and
  Van~Parys]{bertsimas2020sparse}
Dimitris Bertsimas, Jean Pauphilet, and Bart Van~Parys.
\newblock Sparse regression: Scalable algorithms and empirical performance.
\newblock \emph{Statistical Science}, 35\penalty0 (4):\penalty0 555--578, 2020.

\bibitem[Bertsimas et~al.(2021)Bertsimas, Cory-Wright, and
  Pauphilet]{bertsimas2019unified}
Dimitris Bertsimas, Ryan Cory-Wright, and Jean Pauphilet.
\newblock A unified approach to mixed-integer optimization problems with
  logical constraints.
\newblock \emph{SIAM Journal on Optimization}, 31\penalty0 (3):\penalty0
  2340--2367, 2021.

\bibitem[Bertsimas et~al.(2022)Bertsimas, Cory-Wright, and
  Pauphilet]{bertsimas2020mixed}
Dimitris Bertsimas, Ryan Cory-Wright, and Jean Pauphilet.
\newblock Mixed-projection conic optimization: A new paradigm for modeling rank
  constraints.
\newblock \emph{Operations Research}, 70\penalty0 (6):\penalty0 3321--3344,
  2022.

\bibitem[Bertsimas et~al.(2023)Bertsimas, Cory-Wright, and
  Pauphilet]{bertsimas2021perspective}
Dimitris Bertsimas, Ryan Cory-Wright, and Jean Pauphilet.
\newblock A new perspective on low-rank optimization.
\newblock \emph{Mathematical Programming, articles in advance}, pages 1--46,
  2023.

\bibitem[Bezanson et~al.(2017)Bezanson, Edelman, Karpinski, and
  Shah]{bezanson2017julia}
Jeff Bezanson, Alan Edelman, Stefan Karpinski, and Viral~B Shah.
\newblock Julia: A fresh approach to numerical computing.
\newblock \emph{SIAM review}, 59\penalty0 (1):\penalty0 65--98, 2017.

\bibitem[Bienstock(2010)]{bienstock2010eigenvalue}
Daniel Bienstock.
\newblock Eigenvalue techniques for convex objective, nonconvex optimization
  problems.
\newblock In \emph{International Conference on Integer Programming and
  Combinatorial Optimization}, pages 29--42. Springer, 2010.

\bibitem[Bousquet and Elisseeff(2002)]{bousquet2002stability}
Olivier Bousquet and Andr{\'e} Elisseeff.
\newblock Stability and generalization.
\newblock \emph{Journal of Machine Learning Research}, 2:\penalty0 499--526,
  2002.

\bibitem[Boyd et~al.(1994)Boyd, El~Ghaoui, Feron, and
  Balakrishnan]{boyd1994linear}
Stephen Boyd, Laurent El~Ghaoui, Eric Feron, and Venkataramanan Balakrishnan.
\newblock \emph{Linear matrix inequalities in system and control theory}.
\newblock SIAM, 1994.

\bibitem[Burer and Monteiro(2003)]{burer2003nonlinear}
Samuel Burer and Renato~DC Monteiro.
\newblock A nonlinear programming algorithm for solving semidefinite programs
  via low-rank factorization.
\newblock \emph{Mathematical Programming}, 95\penalty0 (2):\penalty0 329--357,
  2003.

\bibitem[Burer and Monteiro(2005)]{burer2005local}
Samuel Burer and Renato~DC Monteiro.
\newblock Local minima and convergence in low-rank semidefinite programming.
\newblock \emph{Mathematical Programming}, 103\penalty0 (3):\penalty0 427--444,
  2005.

\bibitem[Cai et~al.(2019)Cai, Cai, and Wei]{cai2019accelerated}
HanQin Cai, Jian-Feng Cai, and Ke~Wei.
\newblock Accelerated alternating projections for robust principal component
  analysis.
\newblock \emph{Journal of Machine Learning Research}, 20\penalty0
  (1):\penalty0 685--717, 2019.

\bibitem[Candes and Plan(2010)]{candes2010matrix}
Emmanuel~J Candes and Yaniv Plan.
\newblock Matrix completion with noise.
\newblock \emph{Proceedings of the IEEE}, 98\penalty0 (6):\penalty0 925--936,
  2010.

\bibitem[Cand{\`e}s et~al.(2011)Cand{\`e}s, Li, Ma, and
  Wright]{candes2011robust}
Emmanuel~J Cand{\`e}s, Xiaodong Li, Yi~Ma, and John Wright.
\newblock Robust principal component analysis?
\newblock \emph{Journal of the ACM}, 58\penalty0 (3):\penalty0 1--37, 2011.

\bibitem[Chandrasekaran et~al.(2011)Chandrasekaran, Sanghavi, Parrilo, and
  Willsky]{chandrasekaran2011rank}
Venkat Chandrasekaran, Sujay Sanghavi, Pablo~A Parrilo, and Alan~S Willsky.
\newblock Rank-sparsity incoherence for matrix decomposition.
\newblock \emph{SIAM Journal on Optimization}, 21\penalty0 (2):\penalty0
  572--596, 2011.

\bibitem[Chen et~al.(2017)Chen, Liu, and Huang]{chen2017low}
Junbo Chen, Shouyin Liu, and Min Huang.
\newblock Low-rank and sparse decomposition model for accelerating dynamic
  {MRI} reconstruction.
\newblock \emph{Journal of Healthcare Engineering}, 2017, 2017.

\bibitem[Chen and Wainwright(2015)]{chen2015fast}
Yudong Chen and Martin~J Wainwright.
\newblock Fast low-rank estimation by projected gradient descent: General
  statistical and algorithmic guarantees.
\newblock \emph{arXiv preprint arXiv:1509.03025}, 2015.

\bibitem[Chi et~al.(2019)Chi, Lu, and Chen]{chi2019nonconvex}
Yuejie Chi, Yue~M Lu, and Yuxin Chen.
\newblock Nonconvex optimization meets low-rank matrix factorization: An
  overview.
\newblock \emph{IEEE Transactions on Signal Processing}, 67\penalty0
  (20):\penalty0 5239--5269, 2019.

\bibitem[Dong et~al.(2015)Dong, Chen, and Linderoth]{dong2015regularization}
Hongbo Dong, Kun Chen, and Jeff Linderoth.
\newblock Regularization vs. relaxation: A conic optimization perspective of
  statistical variable selection.
\newblock \emph{arXiv preprint arXiv:1510.06083}, 2015.

\bibitem[Fazel(2002)]{fazel2002matrix}
Maryam Fazel.
\newblock \emph{Matrix rank minimization with applications}.
\newblock PhD thesis, Stanford University, 2002.

\bibitem[Gamarnik(2021)]{gamarnik2021overlap}
David Gamarnik.
\newblock The overlap gap property: A topological barrier to optimizing over
  random structures.
\newblock \emph{Proceedings of the National Academy of Sciences}, 118\penalty0
  (41):\penalty0 e2108492118, 2021.

\bibitem[Gillis and Glineur(2011)]{gillis2011low}
Nicolas Gillis and Fran{\c{c}}ois Glineur.
\newblock Low-rank matrix approximation with weights or missing data is
  {NP}-hard.
\newblock \emph{SIAM Journal on Matrix Analysis and Applications}, 32\penalty0
  (4):\penalty0 1149--1165, 2011.

\bibitem[Glover(1975)]{glover1975improved}
Fred Glover.
\newblock Improved linear integer programming formulations of nonlinear integer
  problems.
\newblock \emph{Management Science}, 22\penalty0 (4):\penalty0 455--460, 1975.

\bibitem[Gu et~al.(2016)Gu, Wang, and Liu]{gu2016low}
Quanquan Gu, Zhaoran~Wang Wang, and Han Liu.
\newblock Low-rank and sparse structure pursuit via alternating minimization.
\newblock In \emph{Artificial Intelligence and Statistics}, pages 600--609.
  PMLR, 2016.

\bibitem[G{\"u}nl{\"u}k and Linderoth(2012)]{gunluk2012perspective}
Oktay G{\"u}nl{\"u}k and Jeff Linderoth.
\newblock Perspective reformulation and applications.
\newblock In \emph{Mixed Integer Nonlinear Programming}, pages 61--89.
  Springer, 2012.

\bibitem[Ha et~al.(2020)Ha, Liu, and Barber]{ha2020equivalence}
Wooseok Ha, Haoyang Liu, and Rina~Foygel Barber.
\newblock An equivalence between critical points for rank constraints versus
  low-rank factorizations.
\newblock \emph{SIAM Journal on Optimization}, 30\penalty0 (4):\penalty0
  2927--2955, 2020.

\bibitem[Halko et~al.(2011)Halko, Martinsson, and Tropp]{halko2011finding}
Nathan Halko, Per-Gunnar Martinsson, and Joel~A Tropp.
\newblock Finding structure with randomness: Probabilistic algorithms for
  constructing approximate matrix decompositions.
\newblock \emph{SIAM Review}, 53\penalty0 (2):\penalty0 217--288, 2011.

\bibitem[Jain et~al.(2013)Jain, Netrapalli, and Sanghavi]{jain2013low}
Prateek Jain, Praneeth Netrapalli, and Sujay Sanghavi.
\newblock Low-rank matrix completion using alternating minimization.
\newblock In \emph{Proceedings of the forty-fifth Annual ACM Symposium on
  Theory of Computing}, pages 665--674, 2013.

\bibitem[Ji et~al.(2010)Ji, Liu, Shen, and Xu]{ji2010robust}
Hui Ji, Chaoqiang Liu, Zuowei Shen, and Yuhong Xu.
\newblock Robust video denoising using low rank matrix completion.
\newblock In \emph{2010 IEEE Computer Society Conference on Computer Vision and
  Pattern Recognition}, pages 1791--1798. IEEE, 2010.

\bibitem[Kyrillidis and Cevher(2012)]{kyrillidis2012matrix}
Anastasios Kyrillidis and Volkan Cevher.
\newblock Matrix {ALPS}: Accelerated low rank and sparse matrix reconstruction.
\newblock In \emph{2012 IEEE Statistical Signal Processing Workshop (SSP)},
  pages 185--188. IEEE, 2012.

\bibitem[Land and Doig(2010)]{Land2010}
Ailsa~H. Land and Alison~G. Doig.
\newblock \emph{An Automatic Method for Solving Discrete Programming Problems},
  pages 105--132.
\newblock Springer Berlin Heidelberg, Berlin, Heidelberg, 2010.

\bibitem[Lee and Zou(2014)]{lee2014optimal}
Jon Lee and Bai Zou.
\newblock Optimal rank-sparsity decomposition.
\newblock \emph{Journal of Global Optimization}, 60\penalty0 (2):\penalty0
  307--315, 2014.

\bibitem[Little(1966)]{little1966branch}
John~DC Little.
\newblock \emph{Branch and bound methods for combinatorial problems}.
\newblock PhD thesis, MIT, 1966.

\bibitem[Majumdar et~al.(2020)Majumdar, Hall, and Ahmadi]{majumdar2020recent}
Anirudha Majumdar, Georgina Hall, and Amir~Ali Ahmadi.
\newblock Recent scalability improvements for semidefinite programming with
  applications in machine learning, control, and robotics.
\newblock \emph{Annual Review of Control, Robotics, and Autonomous Systems},
  3:\penalty0 331--360, 2020.

\bibitem[Morrison et~al.(2016)Morrison, Jacobson, Sauppe, and
  Sewell]{MORRISON201679}
David~R. Morrison, Sheldon~H. Jacobson, Jason~J. Sauppe, and Edward~C. Sewell.
\newblock Branch-and-bound algorithms: A survey of recent advances in
  searching, branching, and pruning.
\newblock \emph{Discrete Optimization}, 19:\penalty0 79--102, 2016.

\bibitem[Negahban and Wainwright(2011)]{negahban2011estimation}
Sahand Negahban and Martin~J Wainwright.
\newblock Estimation of (near) low-rank matrices with noise and
  high-dimensional scaling.
\newblock \emph{The Annals of Statistics}, pages 1069--1097, 2011.

\bibitem[Netrapalli et~al.(2014)Netrapalli, Niranjan, Sanghavi, Anandkumar, and
  Jain]{netrapalli2014non}
Praneeth Netrapalli, UN~Niranjan, Sujay Sanghavi, Animashree Anandkumar, and
  Prateek Jain.
\newblock Non-convex robust {PCA}.
\newblock \emph{arXiv preprint arXiv:1410.7660}, 2014.

\bibitem[Overton and Womersley(1992)]{overton1992sum}
Michael~L Overton and Robert~S Womersley.
\newblock On the sum of the largest eigenvalues of a symmetric matrix.
\newblock \emph{SIAM Journal on Matrix Analysis and Applications}, 13\penalty0
  (1):\penalty0 41--45, 1992.

\bibitem[Owen and Perry(2009)]{validation}
Art~B. Owen and Patrick~O. Perry.
\newblock {Bi-cross-validation of the SVD and the nonnegative matrix
  factorization}.
\newblock \emph{The Annals of Applied Statistics}, 3\penalty0 (2):\penalty0 564
  -- 594, 2009.

\bibitem[Pearson(1901)]{pearson1901liii}
Karl Pearson.
\newblock On lines and planes of closest fit to systems of points in space.
\newblock \emph{The London, Edinburgh, and Dublin Philosophical Magazine and
  Journal of Science}, 2\penalty0 (11):\penalty0 559--572, 1901.

\bibitem[Pilanci et~al.(2015)Pilanci, Wainwright, and
  El~Ghaoui]{pilanci2015sparse}
Mert Pilanci, Martin~J Wainwright, and Laurent El~Ghaoui.
\newblock Sparse learning via {B}oolean relaxations.
\newblock \emph{Mathematical Programming}, 151\penalty0 (1):\penalty0 63--87,
  2015.

\bibitem[Recht(2012)]{recht2012projected}
Benjamin Recht.
\newblock Projected gradient methods.
\newblock \emph{Course Notes}, 2012.

\bibitem[Recht et~al.(2010)Recht, Fazel, and Parrilo]{recht2010guaranteed}
Benjamin Recht, Maryam Fazel, and Pablo~A Parrilo.
\newblock Guaranteed minimum-rank solutions of linear matrix equations via
  nuclear norm minimization.
\newblock \emph{SIAM Review}, 52\penalty0 (3):\penalty0 471--501, 2010.

\bibitem[Reuther et~al.(2018)Reuther, Kepner, Byun, Samsi, Arcand, Bestor,
  Bergeron, Gadepally, Houle, Hubbell, Jones, Klein, Milechin, Mullen, Prout,
  Rosa, Yee, and Michaleas]{reuther2018interactive}
Albert Reuther, Jeremy Kepner, Chansup Byun, Siddharth Samsi, William Arcand,
  David Bestor, Bill Bergeron, Vijay Gadepally, Michael Houle, Matthew Hubbell,
  Michael Jones, Anna Klein, Lauren Milechin, Julia Mullen, Andrew Prout,
  Antonio Rosa, Charles Yee, and Peter Michaleas.
\newblock Interactive supercomputing on 40,000 cores for machine learning and
  data analysis.
\newblock In \emph{2018 IEEE High Performance extreme Computing Conference
  (HPEC)}, pages 1--6. IEEE, 2018.

\bibitem[Roos et~al.(2020)Roos, Balvert, Gorissen, and den
  Hertog]{roos2020universal}
Kees Roos, Marleen Balvert, Bram~L Gorissen, and Dick den Hertog.
\newblock A universal and structured way to derive dual optimization problem
  formulations.
\newblock \emph{{INFORMS} Journal on Optimization}, 2\penalty0 (4):\penalty0
  229--255, 2020.

\bibitem[Skajaa and Ye(2015)]{skajaa2015homogeneous}
Anders Skajaa and Yinyu Ye.
\newblock A homogeneous interior-point algorithm for nonsymmetric convex conic
  optimization.
\newblock \emph{Mathematical Programming}, 150\penalty0 (2):\penalty0 391--422,
  2015.

\bibitem[Tillmann and Pfetsch(2013)]{tillmann2013computational}
Andreas~M Tillmann and Marc~E Pfetsch.
\newblock The computational complexity of the restricted isometry property, the
  nullspace property, and related concepts in compressed sensing.
\newblock \emph{IEEE Transactions on Information Theory}, 60\penalty0
  (2):\penalty0 1248--1259, 2013.

\bibitem[Tong et~al.(2021)Tong, Ma, and Chi]{tong2021accelerating}
Tian Tong, Cong Ma, and Yuejie Chi.
\newblock Accelerating ill-conditioned low-rank matrix estimation via scaled
  gradient descent.
\newblock \emph{Journal of Machine Learning Research}, 22\penalty0
  (1):\penalty0 6639--6701, 2021.

\bibitem[Wold et~al.(1987)Wold, Esbensen, and Geladi]{PCA}
Svante Wold, Kim Esbensen, and Paul Geladi.
\newblock Principal component analysis.
\newblock \emph{Chemometrics and Intelligent Laboratory Systems}, 2\penalty0
  (1):\penalty0 37--52, 1987.
\newblock Proceedings of the Multivariate Statistical Workshop for Geologists
  and Geochemists.

\bibitem[Xu et~al.(2009)Xu, Caramanis, and Mannor]{xu2009robustness}
Huan Xu, Constantine Caramanis, and Shie Mannor.
\newblock Robustness and regularization of support vector machines.
\newblock \emph{Journal of Machine Learning Research}, 10\penalty0 (7), 2009.

\bibitem[Yan et~al.(2015)Yan, Ye, and Shen]{yan2015simultaneous}
Qi~Yan, Jieping Ye, and Xiaotong Shen.
\newblock Simultaneous pursuit of sparseness and rank structures for matrix
  decomposition.
\newblock \emph{Journal of Machine Learning Research}, 16\penalty0
  (1):\penalty0 47--75, 2015.

\bibitem[Yi et~al.(2016)Yi, Park, Chen, and Caramanis]{yi2016fast}
Xinyang Yi, Dohyung Park, Yudong Chen, and Constantine Caramanis.
\newblock Fast algorithms for robust {PCA} via gradient descent.
\newblock \emph{Advances in Neural Information Processing Systems}, 29, 2016.

\bibitem[Yuan and Yang(2013)]{yuan2009sparse}
Xiaoming Yuan and Junfeng Yang.
\newblock Sparse and low-rank matrix decomposition via alternating direction
  methods.
\newblock \emph{Pacific Journal of Optimization}, 9\penalty0 (1):\penalty0
  167--180, 2013.

\bibitem[Zhang and Yang(2018)]{zhang2017robust}
Teng Zhang and Yi~Yang.
\newblock Robust {PCA} by manifold optimization.
\newblock \emph{The Journal of Machine Learning Research}, 19\penalty0
  (1):\penalty0 3101--3139, 2018.

\bibitem[Zhou and Tao(2011)]{goDec}
Tianyi Zhou and Dacheng Tao.
\newblock Godec: Randomized low-rank \& sparse matrix decomposition in noisy
  case.
\newblock \emph{Proceedings of the 28th International Conference on Machine
  Learning}, 35:\penalty0 33--40, 2011.

\bibitem[Zhou et~al.(2010)Zhou, Li, Wright, Candès, and Ma]{stable}
Zihan Zhou, Xiaodong Li, John Wright, Emmanuel Candès, and Yi~Ma.
\newblock Stable principal component pursuit.
\newblock In \emph{2010 IEEE International Symposium on Information Theory},
  pages 1518--1522, 2010.

\end{thebibliography}
}

\FloatBarrier
\appendix

{\color{black}
\section{SLR Formulation Properties Omitted Proofs} \label{sec:formulation_properties_proofs} } 

Recall that Proposition \ref{prop:strong_convex} states that $f(\bm{X}, \bm{Y}) = \Vert\bm{D} - \bm{X} - \bm{Y}\Vert_F^2 + \lambda \Vert\bm{X}\Vert_F^2 + \mu \Vert\bm{Y}\Vert_F^2$ is jointly $m$-strongly convex in $(\bm{X}, \bm{Y})$. We prove this fact below:
\begin{proof}
Consider any two points $(\bm{X}_1, \bm{Y}_1), (\bm{X}_2, \bm{Y}_2) \in \mathbb{R}^{n \times n} \times \mathbb{R}^{n \times n}$ and any $t \in [0, 1]$. We have
\begin{equation*}\belowdisplayskip=-12pt
\begin{aligned}
    g(&t\bm{X}_1 + (1-t)\bm{X}_2, \, t\bm{Y}_1 + (1-t)\bm{Y}_2) = \Vert\bm{D} - t\bm{X}_1 + (1-t)\bm{X}_2 - t\bm{Y}_1 + (1-t)\bm{Y}_2\Vert_F^2 + \\
    & \quad \quad (\lambda - \min(\lambda, \mu)) \Vert t\bm{X}_1 + (1-t)\bm{X}_2\Vert_F^2 + (\mu - \min(\lambda, \mu)) \Vert t\bm{Y}_1 + (1-t)\bm{Y}_2\Vert_F^2 \\
    & \quad \quad (\lambda - \min(\lambda, \mu)) \Vert t\bm{X}_1 + (1-t)\bm{X}_2\Vert_F^2 + (\mu - \min(\lambda, \mu)) \Vert t\bm{Y}_1 + (1-t)\bm{Y}_2\Vert_F^2 \\
    &\leq t \cdot \bigg{[}\Vert\bm{D} - \bm{X}_1 - \bm{Y}_1\Vert_F^2 + (\lambda - \min(\lambda, \mu)) \Vert\bm{X}_1\Vert_F^2 + (\mu - \min(\lambda, \mu)) \Vert\bm{Y}_1\Vert_F^2 \bigg{]}+\\
    & \quad \quad (1-t) \bigg{[} \Vert\bm{D} - \bm{X}_2 - \bm{Y}_2\Vert_F^2 + (\lambda - \min(\lambda, \mu)) \Vert\bm{X}_2\Vert_F^2 + (\mu - \min(\lambda, \mu)) \Vert\bm{Y}_2\Vert_F^2 \bigg{]} \\
    &= t \cdot g(\bm{X}_1, \, \bm{Y}_1) + (1-t) \cdot g(\bm{X}_2, \, \bm{Y}_2). 
\end{aligned} \end{equation*} \end{proof}

Recall that Proposition \ref{prop:lipschitz} states that $f(\bm{X}, \bm{Y}) = \Vert\bm{D} - \bm{X} - \bm{Y}\Vert_F^2 + \lambda \Vert\bm{X}\Vert_F^2 + \mu \Vert\bm{Y}\Vert_F^2$ is $L$-Lipschitz continuous in $(\bm{X}, \bm{Y})$. We prove this fact below:
\begin{proof}
To establish Proposition \ref{prop:lipschitz}, it suffices to show that $h(\bm{X}, \bm{Y}) = \frac{L}{2}(\Vert\bm{X}\Vert_F^2+\Vert\bm{Y}\Vert_F^2) - f(\bm{X}, \bm{Y})$ is convex for $L = 2 \cdot \max (\lambda, \mu) + 6$. We have
\begin{equation*}
\begin{aligned}
    h(&\bm{X}, \, \bm{Y}) = \frac{L}{2} (\Vert\bm{X}\Vert_F^2+\Vert\bm{Y}\Vert_F^2) - \lambda \Vert\bm{X}\Vert_F^2 - \mu \Vert\bm{Y}\Vert_F^2 - \Vert\bm{D} - \bm{X} - \bm{Y}\Vert_F^2\\
    &= \bigg{(}\frac{L}{2}-\lambda-1\bigg{)}\Vert\bm{X}\Vert_F^2 + \bigg{(}\frac{L}{2}-\mu-1\bigg{)}\Vert\bm{Y}\Vert_F^2 + 2 \Big{(} \langle \bm{D}, \bm{X} \rangle + \langle \bm{D}, \bm{Y} \rangle - \langle \bm{X}, \bm{Y} \rangle \Big{)} - \Vert\bm{D}\Vert_F^2\\
    &= \bigg{(}\frac{L}{2}-\lambda-2\bigg{)}\Vert\bm{X}\Vert_F^2 + \bigg{(}\frac{L}{2}-\mu-2\bigg{)}\Vert\bm{Y}\Vert_F^2 + \Vert \bm{X} - \bm{Y} \Vert_F^2+\\
    & \quad \quad \quad 2 \Big{(} \langle \bm{D}, \bm{X} \rangle + \langle \bm{D}, \bm{Y} \rangle +\Vert\bm{D}\Vert_F^2 \Big{)} - 3\Vert\bm{D}\Vert_F^2\\
    &= \bigg{(}\frac{L}{2}-\lambda-3\bigg{)}\Vert\bm{X}\Vert_F^2 + \bigg{(}\frac{L}{2}-\mu-3\bigg{)}\Vert\bm{Y}\Vert_F^2 + \Vert \bm{X} - \bm{Y} \Vert_F^2+ \\
    & \quad \quad \quad \Vert \bm{X} - \bm{D} \Vert_F^2 + \Vert \bm{Y} - \bm{D} \Vert_F^2 - 3\Vert\bm{D}\Vert_F^2.
\end{aligned} \end{equation*} Taking $L =  2 \cdot \max (\lambda, \mu) + 6$, we have $\frac{L}{2}-\lambda-3=\max (\lambda, \mu) - \lambda \geq 0$ and $\frac{L}{2}-\mu-3=\max (\lambda, \mu) - \mu \geq 0$. Thus, we have written $h(\bm{X}, \bm{Y})$ as the sum of convex quadratic functions of $(\bm{X}, \bm{Y})$ which immediately implies $h(\bm{X}, \bm{Y})$'s joint convexity. \end{proof}

Recall that Proposition \ref{prop:1} states if we let $\mathcal{U}_\lambda(\bm{X}) = \{\bm{\Delta} \in \mathbb{R}^{n \times n} : \Vert\bm{\Delta}\Vert_F \leq \lambda \Vert \bm{X} \Vert_F\}$ for $\bm{X} \in \mathbb{R}^{n \times n}, \lambda > 0$, then \eqref{thm:robust_prob} is equivalent to \eqref{thm:reg_prob}. We prove this result below:

\begin{proof}
Consider the inner maximization problem in \eqref{thm:robust_prob} and first note that by applying the triangle inequality for the Frobenius norm, we have

\begin{equation*}
\begin{aligned}
    \max_{\substack{\bm{\Delta}_1 \in \mathcal{U}_\lambda(\bm{X}) \\ \bm{\Delta}_2 \in \mathcal{U}_\mu(\bm{Y})}} \Vert\bm{D} + \bm{\Delta_1} + \bm{\Delta_2} - \bm{X} - \bm{Y}\Vert_F \leq \Vert\bm{D} - \bm{X} - \bm{Y}\Vert_F + \lambda \Vert\bm{X}\Vert_F + \mu \Vert\bm{Y}\Vert_F.
\end{aligned}
\end{equation*} Next, note that by taking \[\bm{\Delta_1}^* = \frac{\bm{D}-\bm{X}-\bm{Y}}{\Vert\bm{D}-\bm{X}-\bm{Y}\Vert_F} \cdot \lambda \Vert\bm{X}\Vert_F, \text{ and}\] \[\bm{\Delta_2}^* = \frac{\bm{D}-\bm{X}-\bm{Y}}{\Vert\bm{D}-\bm{X}-\bm{Y}\Vert_F} \cdot \mu \Vert\bm{Y}\Vert_F,\] the upper bound on the maximization problem is attained:

\begin{equation*}
\begin{aligned}
    \Vert\bm{D} + \bm{\Delta_1}^* + \bm{\Delta_2}^* - \bm{X} - \bm{Y}\Vert_F &= \bigg{|}\bigg{|}(\bm{D} - \bm{X} - \bm{Y}) \cdot \bigg{(}1 + \frac{\lambda \Vert\bm{X}\Vert_F+\mu \Vert\bm{Y}\Vert_F}{\Vert\bm{D}-\bm{X}-\bm{Y}\Vert_F}\bigg{)}\bigg{|}\bigg{|}_F \\
    &= \Vert\bm{D}-\bm{X}-\bm{Y}\Vert_F+\lambda \Vert\bm{X}\Vert_F+\mu \Vert\bm{Y}\Vert_F.
\end{aligned}
\end{equation*} The proof is concluded by noting that we have $\bm{\Delta_1}^* \in \mathcal{U}_\lambda(\bm{X})$ and $\bm{\Delta_2}^* \in \mathcal{U}_\mu(\bm{Y})$.
\end{proof}

We now provide a formal proof of Proposition \ref{prop:regrobust_full}:
\begin{proof}
Let us rewrite Problem \eqref{opt:main_problem} as
\begin{align*}
\min_{\bm{X}, \bm{Y}, \bm{U}, \bm{V}} \quad & \Vert \bm{D}-\bm{X}-\bm{Y}\Vert_F^2+\lambda \Vert \bm{U}\Vert_F^2+\mu \Vert \bm{V}\Vert_F^2 \\
\text{s.t.} \quad & \mathrm{Rank}(\bm{X}) \leq k_0, \ \Vert \bm{Y}\Vert_0 \leq k_1, \bm{X}=\bm{U}, \bm{Y}=\bm{V},
\end{align*}
and associate matrices of dual multipliers $\bm{\alpha}, \bm{\beta}$ with the linear constraints $\bm{X}=\bm{U}$ and $\bm{Y}=\bm{V}$ respectively. Then, this problem can be rewritten as 
\begin{align*}
\min_{\bm{X}, \bm{Y}} \min_{\bm{U}, \bm{V}} \max_{\bm{\alpha}, \bm{\beta}} \quad & \Vert \bm{D}-\bm{X}-\bm{Y}\Vert_F^2+\lambda \Vert \bm{U}\Vert_F^2+\mu \Vert \bm{V}\Vert_F^2 +\langle \bm{\alpha}, \bm{X}-\bm{U}\rangle+\langle \bm{\beta}, \bm{Y}-\bm{V}\rangle \\
\text{s.t.} \quad & \mathrm{Rank}(\bm{X}) \leq k_0, \ \Vert \bm{Y}\Vert_0 \leq k_1.
\end{align*}
Therefore, let us fix $\bm{X}, \bm{Y}$ and use a standard minimax theorem \citep[see, e.g.,][Chap. 6]{bertsekas1999nonlinear} to exchange the order of minimizing $\bm{U}, \bm{V}$ and maximizing $\bm{\alpha}, \bm{\beta}$. This gives the following subproblem in $\bm{U}, \bm{V}$ for a fixed $\bm{\alpha}, \bm{\beta}$: 
\begin{align*}
\min_{\bm{U}, \bm{V}} \quad & \lambda \Vert \bm{U}\Vert_F^2+\mu \Vert \bm{V}\Vert_F^2 +\langle \bm{\alpha}, -\bm{U}\rangle+\langle \bm{\beta}, -\bm{V}\rangle.
\end{align*}
By differentiating and setting the gradient to zero, it is not too hard to see that this subproblem takes the value $\frac{-1}{4\lambda} \Vert \bm{\alpha}\Vert_F^2-\frac{1}{4\mu}\Vert \bm{\beta}\Vert_F^2$. This implies the result.
\end{proof}

Recall that Proposition \ref{prop:2} establishes that \eqref{opt:main_problem} reduces to regularized matrix completion with $\Omega = \{(i, j): Z_{ij} = 0\}$ where $\bm{Z}$ denotes a valid sparsity pattern and we take $\mu=0$. We prove this result below:

\begin{proof}
Given a valid sparsity pattern $\bm{Z}$ and letting $\Omega = \{(i, j): Z_{ij} = 0\}$, Problem \eqref{opt:main_problem} can be expressed as
\begin{equation*}
\begin{aligned}
    \min_{\bm{X}, \bm{Y} \in \mathbb{R}^{n \times n}} \quad & \lambda \Vert \bm{X} \Vert _F^2 + \sum_{(i, j) \in \Omega} (D_{ij}-X_{ij}-Y_{ij})^2+\mu Y_{ij}^2 +\sum_{(i, j) \notin \Omega} (D_{ij}-X_{ij}-Y_{ij})^2+\mu Y_{ij}^2\\
    \text{s.t.}\quad & \mathrm{Rank}(\bm{X}) \leq k_0, \,\, Y_{ij} = 0 \,\, \forall \,\, (i,j) \in \Omega.
\end{aligned}
\end{equation*} Simple unconstrained minimization gives $Y_{ij} = \frac{D_{ij}-X_{ij}}{1+\mu}$ for $(i,j) \notin \Omega$. Using this relationship, Problem \eqref{opt:main_problem} can be further simplified to
\begin{equation}
\begin{aligned}
    \min_{\bm{X}\in \mathbb{R}^{n \times n}} \quad & \lambda \cdot \Vert \bm{X} \Vert _F^2 + \sum_{(i, j) \in \Omega}(D_{ij}-X_{ij})^2 + \frac{\mu}{1+\mu} \cdot \sum_{(i, j) \notin \Omega} (D_{ij}-X_{ij})^2 \cdot \\
    \text{s.t.} \quad & \mathrm{Rank}(\bm{X}) \leq k_0.
\end{aligned} \label{opt:mat_comp_reg}
\end{equation} The result then follows by observing that the last term in the objective function of \eqref{opt:mat_comp_reg} disappears when $\mu=0$. Moreover, if we take $\lambda = 0$, then \eqref{opt:mat_comp_reg} exactly becomes \eqref{opt:mat_comp}.
\end{proof}

We now provide a formal proof of Proposition \ref{prop:PGD}:

\begin{proof}
First, note that given the full sparsity pattern, the iterates $(\bm{X}_t^{AM}, \bm{Y}_t^{AM})$ produced by Algorithm \ref{alg:AM} satisfy $\bm{Y}_{t+1}^{AM} = \bm{S}^* \circ \Big{(}\frac{\bm{D}-\bm{X}_t^{AM}}{1 + \mu}\Big{)}$ and $\bm{X}_{t+1}^{AM} = \frac{1}{1+\lambda} \mathcal{P}_\Omega(\bm{D}-\bm{Y}_{t+1}^{AM})$. This implies that
\begin{equation}
\begin{aligned}
\bm{X}_{t+1}^{AM} = \mathcal{P}_\Omega\bigg{(}\frac{1}{1+\lambda} \bigg{[}\bm{D}-\bm{S}^* \circ \bigg{(}\frac{\bm{D}-\bm{X}_t^{AM}}{1 + \mu}\bigg{)}\bigg{]}\bigg{)}.
\end{aligned} \label{eq:AM_update}
\end{equation} Next, note that the gradient of $g(\bm{X}_t)$ is given by
\[
\nabla g(\bm{X}_t) = 2\bigg{(} (1+\lambda)\bm{X}_t-\bm{D}+ \bm{S}^* \circ \bigg{(}\frac{\bm{D}-\bm{X}_t}{1 + \mu}\bigg{)} \bigg{)}.
\] The result follows by noting that the Projected Gradient Descent update $\bm{X}_{t+1} = \mathcal{P}_\Omega(\bm{X}_t - \eta \nabla g(\bm{X}_t))$ is the same as the update given by \eqref{eq:AM_update} when $\eta = \frac{1}{2(1+\lambda)}$.
\end{proof}

We now provide a formal proof of Proposition \ref{prop:main_reform}:
\begin{proof}
We show that given a feasible solution to \eqref{opt:main_problem_reform}, we can construct a feasible solution to \eqref{opt:main_problem} that achieves the same objective value and vice versa.

Consider an arbitrary feasible solution $(\bm{\Bar{X}}, \bm{\Bar{Y}}, \bm{\Bar{Z}}, \bm{\Bar{P}})$ to \eqref{opt:main_problem_reform}. Since $\bm{\Bar{Z}} \in \mathcal{Z}_{k_1}$ and $\bm{\Bar{Y}} = \bm{\Bar{Z}} \circ \bm{\Bar{Y}}$, we have $\Vert \bm{\Bar{Y}} \Vert_0 \leq k_1$. Moreover, since $\bm{\Bar{P}} \in \mathcal{P}_{k_0}$ and $\bm{\Bar{X}} = \bm{\Bar{P}}\bm{\Bar{X}}$, we have $\mathrm{Rank}(\bm{\Bar{X}}) \leq k_0$. Thus, $(\bm{\Bar{X}}, \bm{\Bar{Y}})$ is feasible to \eqref{opt:main_problem}. Since both \eqref{opt:main_problem_reform} and \eqref{opt:main_problem} have the same objective function, $(\bm{\Bar{X}}, \bm{\Bar{Y}})$ achieves the same objective in \eqref{opt:main_problem} as $(\bm{\Bar{X}}, \bm{\Bar{Y}}, \bm{\Bar{Z}}, \bm{\Bar{P}})$ does in \eqref{opt:main_problem_reform}.

Consider an arbitrary feasible solution $(\bm{\Bar{X}}, \bm{\Bar{Y}})$ to \eqref{opt:main_problem}. Let $\bm{\Bar{Z}} \in \{0, 1\}^{n \times n}$ be the binary matrix such that $\Bar{Z}_{ij} = 1$ if $\Bar{Y}_{ij} \neq 0$ and $\Bar{Z}_{ij} = 0$ otherwise. Further, let $\bm{\Bar{P}} = \bm{U} \bm{U}^T$ where $\bm{\Bar{X}} = \bm{U} \bm{\Sigma} \bm{V}^T$ is a singular value decomposition of $\bm{\Bar{X}}$. By construction, we have $\bm{\Bar{Z}} \in \mathcal{Z}_{k_1}$ and $\bm{\Bar{P}} \in \mathcal{P}_{k_0}$ since $\Vert \bm{\Bar{Y}} \Vert _0 \leq k_1$ and $\mathrm{Rank}(\bm{\Bar{X}}) \leq k_0$. Thus, $(\bm{\Bar{X}}, \bm{\Bar{Y}}, \bm{\Bar{Z}}, \bm{\Bar{P}})$ is feasible to \eqref{opt:main_problem_reform} and achieves the same objective as $(\bm{\Bar{X}}, \bm{\Bar{Y}})$ does in \eqref{opt:main_problem}. This completes the proof. 
\end{proof}

We now provide a formal proof of Theorem \ref{thm:convex_relax}:
\begin{proof}
Clearly Problem \eqref{opt:convex_relax} is a convex optimization problem. We will show that given any feasible solution to Problem \eqref{opt:main_problem}, we can construct a feasible solution to \eqref{opt:convex_relax} that achieves the same objective value.

Consider an arbitrary feasible solution $(\bm{\Bar{X}}, \bm{\Bar{Y}})$ to \eqref{opt:main_problem}. Let $\bm{\Bar{Z}} \in \{0, 1\}^{n \times n}$ be the binary matrix such that $\Bar{Z}_{ij} = 1$ if $\Bar{Y}_{ij} \neq 0$ and $\Bar{Z}_{ij} = 0$ otherwise and let $\bm{\Bar{\alpha}} \in \mathbb{R}^{n \times n}$ be the matrix such that $\Bar{\alpha}_{ij} = \Bar{Y}_{ij}^2$. Further, let $\bm{\Bar{P}} = \bm{U} \bm{U}^T$ where $\bm{\Bar{X}} = \bm{U} \bm{\Sigma} \bm{V}^T$ is a singular value decomposition of $\bm{\Bar{X}}$ and let $\bm{\Bar{\Theta}} = \bm{\Bar{X}}^T \bm{\Bar{X}}$. By construction, we have $\bm{\Bar{Z}} \in \mathcal{Z}_{k_1}$ and $\bm{\Bar{P}} \in \mathcal{P}_{k_0}$ since $\Vert \bm{\Bar{Y}} \Vert _0 \leq k_1$ and $\mathrm{Rank}(\bm{\Bar{X}}) \leq k_0$ which implies that $\text{tr}(\bm{E}\bm{\Bar{Z}}) \leq k_1, 0 \leq \bm{\Bar{Z}} \leq 1, \bm{\Bar{P}} \succeq 0, \mathbb{I} - \bm{\Bar{P}} \succeq 0$ and $\text{tr}(\bm{\Bar{P}}) \leq k_0$. It is straightforward to see that we have $\Bar{Y}_{ij}^2 \leq \Bar{\alpha}_{ij} \Bar{Z}_{ij} \,\, \forall \, (i, j)$. Finally, we have $\bm{\Bar{\Theta}} = \bm{\Bar{X}}^T \bm{\Bar{X}} = \bm{\Bar{X}}^T \bm{\Bar{P}} \bm{\Bar{X}} = \bm{\Bar{X}}^T \bm{\Bar{P}}^\dag \bm{\Bar{X}}$ so we have $\begin{pmatrix}\bm{\Bar{\Theta}} & \bm{\Bar{X}}\\ \bm{\Bar{X}}^T & \bm{\Bar{P}}\end{pmatrix} \succeq 0$. Thus, we have shown that $(\bm{\Bar{X}}, \bm{\Bar{Y}}, \bm{\Bar{Z}}, \bm{\Bar{P}}, \bm{\Bar{\Theta}}, \bm{\Bar{\alpha}})$ is feasible to \eqref{opt:convex_relax}. This achieves an objective of

\begin{equation*}
\begin{aligned}
   \Vert\bm{D} - \bm{\Bar{X}} - \bm{\Bar{Y}}\Vert_F^2 + \lambda \text{tr}(\bm{\Bar{\Theta}}) + \mu \text{tr}(\bm{E}\bm{\Bar{\alpha}})&= \Vert\bm{D} - \bm{\Bar{X}} - \bm{\Bar{Y}}\Vert_F^2 + \lambda \text{tr}(\bm{\Bar{X}}^T \bm{\Bar{X}}) + \mu \sum_{ij} \Bar{Y}_{ij}^2\\
    &= \Vert\bm{D} - \bm{\Bar{X}} - \bm{\Bar{Y}}\Vert_F^2 + \lambda \Vert \bm{\Bar{X}} \Vert_F^2 + \mu \Vert \bm{\Bar{Y}}\Vert_F^2.
\end{aligned}
\end{equation*} which is the same objective achieved by $(\bm{\Bar{X}}, \bm{\Bar{Y}})$ in \eqref{opt:main_problem}. This completes the proof. \end{proof}

{\color{black}
\section{Alternative Proof of Proposition \ref{prop:rank_subproblem}} \label{sec:appendix_rank_proof} }

\begin{proof}
Clearly, $\bm{X^*}$ is feasible for \eqref{opt:fixed_Y}. Let $\bm{P^*} = \bm{U}_{k_0}\bm{U}_{k_0}^T$ and $\bm{\Theta^*} = \bm{X}^{*T}\bm{X}^*$. As established in the proof of Theorem \ref{thm:pca}, $(\bm{X^*}, \bm{P^*}, \bm{\Theta^*})$ is feasible to \eqref{opt:pca_sdp} and achieves the same objective as $\bm{X^*}$ does in \eqref{opt:fixed_Y}. We prove Proposition \ref{prop:rank_subproblem} by deriving the dual of \eqref{opt:pca_sdp} and constructing a dual feasible solution that achieves the same objective value as $(\bm{X^*}, \bm{P^*}, \bm{\Theta^*})$ achieves in \eqref{opt:pca_sdp}. By duality, this then implies that $(\bm{X^*}, \bm{P^*}, \bm{\Theta^*})$ is optimal for \eqref{opt:pca_sdp} which in turn implies that $\bm{X^*}$ is optimal for \eqref{opt:fixed_Y}.

The dual of \eqref{opt:pca_sdp} is given by
\begin{equation}
\begin{aligned}
    \max_{\bm{A}, \bm{B} \in \mathcal{S}^n_+, \sigma \geq 0} \quad & \Vert\bm{\Bar{D}}\Vert_F^2 + \sigma(n-k_0)-\text{tr}(\bm{B})\\
    \text{s.t.} \quad & (1+\lambda)\mathbb{I} \succeq \bm{A}, \ \bm{B} \succeq \sigma \mathbb{I}, \ \begin{pmatrix}\bm{A} & \bm{\bar{D}}\\ \bm{\bar{D}}^T & \bm{B}\end{pmatrix} \succeq 0.
\end{aligned} \label{opt:pca_sdp_dual}
\end{equation} Let $\{\phi_i\}_{i=1}^n$ denote the collection of singular values of $\bm{\bar{D}}$ in non-increasing order (so that $\phi_i \geq \phi_{i+1} \,\, \forall \, i$). Let $\sigma^* = \frac{1}{1+\lambda}\phi_{k_0}^2$. Let $\nu_i^* = \frac{1}{1+\lambda}\phi_i^2 \,\, \forall \, i < k_0$ and let $\nu_i^* = \sigma^* \,\, \forall \, k_0 \leq i \leq n$. Let $\bm{A^*} = (1+\lambda) \mathbb{I}$ and $\bm{B^*} = \bm{U} Diag(\bm{\nu}) \bm{U}^T$ where $\bm{\Bar{D}} = \bm{U}\bm{\Phi}\bm{U}^T$ is a spectral decomposition of $\bm{\Bar{D}}$ and $Diag(\bm{\nu})$ denotes the $n \times n$ diagonal matrix with diagonal entries given by the entries of $\bm{\nu}$. Note that the solution $(\bm{A^*}, \bm{B^*}, \sigma^*)$ is feasible to \eqref{opt:pca_sdp_dual}. To see this, observe that by construction, we have $\bm{A^*}, \bm{B^*} \in \mathcal{S}^n_+, \sigma^* \geq 0,$ and $(1+\lambda)\mathbb{I} \succeq \bm{A^*}$. Moreover, since $\{\phi_i\}_{i=1}^n$ are in non-increasing order, we have $\min_i \nu_i \geq \sigma^*$ which implies $\bm{B^*} \succeq \sigma \mathbb{I}$. Finally, we have $\nu_i \geq \frac{1}{1+\lambda}\phi_i^2 \,\, \forall \, i$ which implies that $\bm{B^*} \succeq \bm{\Bar{D}}^T \bm{A}^{*-1} \bm{\Bar{D}}$ and $\begin{pmatrix}\bm{A^*} & \bm{\bar{D}}\\ \bm{\bar{D}}^T & \bm{B^*}\end{pmatrix} \succeq 0$. The feasible solution $(\bm{A^*}, \bm{B^*}, \sigma^*)$ achieves an objective of:

\begin{equation*}
\begin{aligned}
    \Vert\bm{\Bar{D}}\Vert_F^2 + \sigma^*(n-k_0)-\text{tr}(\bm{B^*})&= \sum_{i=1}^n \phi_i^2 + \frac{n-k_0}{1+\lambda}\phi_{k_0}^2-\frac{1}{1+\lambda}\sum_{i=1}^{k_0-1}\phi_i^2-\frac{1}{1+\lambda}\sum_{i=k_0}^{n}\phi_{k_0}^2\\
    &= \frac{\lambda}{1+\lambda}\sum_{i=1}^{k_0} \phi_i^2+\sum_{i=k_0+1}^n\phi_i^2
\end{aligned}
\end{equation*} in \eqref{opt:pca_sdp_dual}. Moreover, the solution $(\bm{X^*}, \bm{P^*}, \bm{\Theta^*})$ achieves the same objective in \eqref{opt:pca_sdp}:

\begin{equation*}
\begin{aligned}
    \Vert\bm{\Bar{D}}\Vert_F^2 + (1+\lambda)\text{tr}(\bm{\Bar{\Theta^*}}) - 2 \cdot \text{tr}(\bm{\Bar{X^*}}\bm{\Bar{D}})&= \sum_{i=1}^n \phi_i^2 + \frac{1}{1+\lambda}\sum_{i+1}^{k_0} \phi_i^2 - \frac{2}{1+\lambda}\sum_{i=1}^{k_0} \phi_i^2\\
    &= \frac{\lambda}{1+\lambda}\sum_{i=1}^{k_0} \phi_i^2+\sum_{i=k_0+1}^n\phi_i^2.
\end{aligned}
\end{equation*} By duality, the objective value of any feasible solution to \eqref{opt:pca_sdp_dual} provides a lower bound on the objective of \eqref{opt:pca_sdp}. Since $(\bm{X^*}, \bm{P^*}, \bm{\Theta^*})$ is primal feasible and achieves the same objective as a feasible dual solution, it must be optimal for \eqref{opt:pca_sdp}. This in turn implies that $\bm{X^*}$ is optimal to \eqref{opt:fixed_Y} by Theorem \ref{thm:pca}. This completes the proof.
\end{proof}

\section{Proof of Convexity in the Low-Rank Subproblem}\label{sec:proofofconv}

\begin{proof}
We prove the equivalence in two steps. First, we show that given a feasible solution to \eqref{opt:fixed_Y}, we can construct a feasible solution to \eqref{opt:pca_sdp} that achieves the same objective value. 
Second, we show that given a feasible solution to \eqref{opt:pca_sdp}, we can construct a feasible solution to \eqref{opt:fixed_Y} that achieves the same or lower objective. Given an arbitrary feasible solution to \eqref{opt:pca_sdp}, we construct a linear optimization problem in which feasible solutions correspond to feasible solutions to \eqref{opt:pca_sdp} and extreme points of the feasible set of the linear optimization problem correspond to feasible solutions to \eqref{opt:fixed_Y}. The initial feasible solution to \eqref{opt:pca_sdp} is feasible to this linear optimization problem, so there is an extreme point corresponding to a feasible solution to \eqref{opt:fixed_Y} that achieves an equal or lower objective value.

Consider an arbitrary feasible solution $\bm{\Bar{X}}$ to Problem \eqref{opt:fixed_Y}. Since $\bm{D}$ is symmetric, we can restrict ourselves to considering symmetric feasible solutions. Since we have $\mathrm{Rank}(\bm{\Bar{X}}) \leq k$ and $\bm{\Bar{X}}$ is symmetric, we can factor $\bm{\Bar{X}}$ as $\bm{\Bar{X}} = \bm{U}\bm{\Sigma} \bm{U}^T$ where $\bm{U} \in \mathbb{R}^{n \times k_0}$, $\bm{U}^T\bm{U} = \mathbb{I}_{k_0}$, $\bm{\Sigma} \in \mathbb{R}^{k_0 \times k_0}$ and $\bm{\Sigma}$ is diagonal. Let $\bm{\Bar{P}} = \bm{U}\bm{U}^T$. $\bm{\Bar{P}}$ is the orthogonal projection matrix onto the $k_0$ dimensional column space of $\bm{U}$. This implies that $\bm{\Bar{P}} \succeq 0$, $\mathbb{I} - \bm{\Bar{P}} \succeq 0$ and $\text{tr}(\bm{\Bar{P}}) \leq k_0$. Let $\bm{\Bar{\Theta}} = \bm{\Bar{X}}^T\bm{\Bar{X}} \succeq 0$. Note that $\bm{\Bar{P}}\bm{\Bar{X}} = \bm{\Bar{X}}$ and $\bm{\Bar{P}}=\bm{\Bar{P}}^\dag$, where $\bm{\Bar{P}}^\dag$ denotes the pseudo-inverse of $\bm{\Bar{P}}$, since $\bm{\Bar{P}}$ is an orthogonal projection matrix. Thus, we have $\bm{\Bar{\Theta}}-\bm{\Bar{X}}^T\bm{\Bar{P}}^\dag\bm{\Bar{X}} = 0 \implies \begin{pmatrix}\bm{\Bar{\Theta}} & \bm{\Bar{X}}\\ \bm{\Bar{X}}^T & \bm{\Bar{P}}\end{pmatrix} \succeq 0$. We have shown that $(\bm{\Bar{X}}, \bm{\Bar{P}}, \bm{\Bar{\Theta}})$ is feasible to \eqref{opt:pca_sdp}. To see that this solution achieves the same objective as $\bm{\Bar{X}}$ achieves in \eqref{opt:fixed_Y}, note that

\begin{equation*}
\begin{aligned}
    \Vert\bm{\Bar{D}} - \bm{\Bar{X}}\Vert_F^2 + \lambda \Vert\bm\Bar{X} \Vert_F^2&= \Vert\bm{\Bar{D}}\Vert_F^2 + (1+\lambda)\Vert\bm{\Bar{X}}\Vert_F^2 - 2 \cdot \text{tr}(\bm{\Bar{X}}\bm{\Bar{D}})\\
    &= \Vert\bm{\Bar{D}}\Vert_F^2 + (1+\lambda)\text{tr}(\bm{\Bar{\Theta}}) - 2 \cdot \text{tr}(\bm{\Bar{X}}\bm{\Bar{D}}).
\end{aligned}
\end{equation*} 

Now, consider an arbitrary feasible solution $(\bm{\Bar{X}}, \bm{\Bar{P}}, \bm{\Bar{\Theta}})$ to \eqref{opt:pca_sdp}. Since the objective function of \eqref{opt:pca_sdp} includes the term $\text{tr}(\bm{\Theta})$ and feasibility requires $\bm{\Theta} \succeq \bm{X}^T \bm{P}^\dag \bm{X}$, we can take $\bm{\Theta}' = \bm{\Bar{X}}^T \bm{\Bar{P}}^\dag \bm{\Bar{X}}$ and the solution $(\bm{\Bar{X}}, \bm{\Bar{P}} , \bm{\Theta}')$ will be feasible to \eqref{opt:pca_sdp} with an objective value no greater than that of the original feasible solution. Since $\bm{\Bar{P}}$ is PSD, it can be written as $\bm{\Bar{P}} = \sum_{i=1}^n \phi_i u_i u_i^T$ where $u_i^Tu_i=1$ for all $i$, $u_i^Tu_j=0$ for all $i \neq j$ and the feasibility of $\bm{\Bar{P}}$ implies $0 \leq \phi_i \leq 1$ for all $i$. Moreover, we have $\bm{\Bar{P}}^\dag= \sum_{i:\phi_i \neq 0} \frac{1}{\phi_i}u_i u_i^T$. Further, since the feasibility condition $\begin{pmatrix}\bm{\Theta}' & \bm{\Bar{X}}\\ \bm{\Bar{X}}^T & \bm{\Bar{P}}\end{pmatrix} \succeq 0$ implies that $\bm{\Bar{X}} = \bm{\Bar{P}}^\dag\bm{\Bar{P}}\bm{\Bar{X}}$ by the generalized Schur complement lemma (see Boyd et al. 1994, Equation 2.41) and $\bm{\Bar{X}}$ is symmetric, without loss of generality it can be written as $\bm{\Bar{X}} = \sum_{i=1}^n \sigma_i u_i u_i^T$. The solution $(\bm{\Bar{X}}, \bm{\Bar{P}} , \bm{\Theta}')$ achieves an objective of 

\begin{equation*}
\begin{aligned}
    h(\bm{\Bar{X}}, \bm{\Bar{P}} , \bm{\Theta}') &= \Vert\bm{\Bar{D}}\Vert_F^2 + (1+\lambda)\text{tr}(\bm{\Theta}') - 2 \cdot \text{tr}(\bm{\Bar{X}}\bm{\Bar{D}}) \\
    &= \Vert\bm{\Bar{D}}\Vert_F^2 + \sum_{i:\phi_i \neq 0} \bigg{[}\frac{1+\lambda}{\phi_i}\sigma_i^2 - 2 \cdot \sigma_i\text{tr}( u_i u_i^T\bm{\Bar{D}}) \bigg{]}.
\end{aligned}
\end{equation*} Note that if we view the above as a function of $\sigma_i$ and $\phi_i$ (denoted $f(\phi, \sigma)$), then this expression corresponds to the objective value achieved by some feasible solution to \eqref{opt:pca_sdp} provided we constrain $0 \leq \phi_i \leq 1$ and $\sum_i \phi_i \leq k_0$. $h(\phi, \sigma)$ is a convex quadratic in $\sigma_i$. It is minimized when $\nabla_{\sigma_i}h(\phi, \sigma) = \frac{2(1+\lambda)}{\phi_i}\sigma_i - 2\text{tr}(u_iu_i^T\bm{\Bar{D}}) = 0 \implies \sigma_i = \frac{\phi_i}{1+\lambda}\text{tr}(u_iu_i^T\bm{\Bar{D}})$. Substituting the optimal value of $\sigma_i$ into $h(\phi, \sigma)$, we obtain
\begin{equation*}
\begin{aligned}
    h(\phi) = \min_\sigma f(\phi, \sigma) = \Vert\bm{\Bar{D}}\Vert_F^2 - \sum_{i:\phi_i \neq 0} \frac{\phi_i}{1+\lambda}[\text{tr}(u_iu_i^T\bm{\Bar{D}})]^2 = \Vert\bm{\Bar{D}}\Vert_F^2 - \sum_{i=1}^n \frac{\phi_i}{1+\lambda}[\text{tr}(u_iu_i^T\bm{D}^*)]^2.
\end{aligned}
\end{equation*} $h(\phi)$ is a linear function of $\phi$. Therefore, the minimum of $h(\phi)$ over the set $0 \leq \phi_i \leq 1$ for all $i$, $\sum_i \phi_i \leq k_0$ is achieved at some $\phi^* \in \{0, 1\}^{n \times n}$. Let $\bm{P}^* = \sum_{i=1}^n \phi_i^* u_i u_i^T$, $\bm{X}^* = \sum_{i=1}^n \phi_i^*\text{tr}(u_iu_i^T\bm{\Bar{D}}) u_i u_i^T$ and $\bm{\Theta}^* = \bm{X}^{*T}\bm{P}^*\bm{X}$. Then $(\bm{X}^*, \bm{P}^*, \bm{\Theta}^*)$ is feasible to \eqref{opt:pca_sdp} and achieves objective $h(\phi^*)$. By construction, we have \[h(\phi^*) \leq h(\bm{\Bar{X}}, \bm{\Bar{P}} , \bm{\Theta}') \leq h(\bm{\Bar{X}}, \bm{\Bar{P}} , \bm{\Bar{\Theta}}).\] Further, since $\phi^* \in \{0, 1\}^{n \times n}$ and $\sum_i \phi_i^* \leq k_0$, we have $\mathrm{Rank}(\bm{X}^*) \leq k_0$ which means that $\bm{X}^*$ is feasible to \eqref{opt:fixed_Y} and achieves objective $h(\phi^*)$. This completes the proof.
\end{proof}

\section{Alternative Proof of Proposition \ref{prop:sparsesubproblem}} \label{sec:appendix_sparse_proof}

\begin{proof}
Let $f(\bm{Y}) = \Vert\bm{\Tilde{D}} - \bm{Y}\Vert_F^2 + \mu \Vert\bm{Y}\Vert_F^2$, the objective function of Problem \eqref{opt:fixed_X}. We can rewrite $f(\bm{Y})$ as:
\begin{equation*}
\begin{aligned}
    f(\bm{Y}) &= \Vert\bm{\Tilde{D}} - \bm{Y}\Vert_F^2 + \mu \Vert\bm{Y}\Vert_F^2 = \sum_{ij}(\Tilde{d}_{ij}-y_{ij})^2 + \mu \sum_{ij}y_{ij}^2 \\
    &= \sum_{ij}\Big{[} (\Tilde{d}_{ij}-y_{ij})^2 + y_{ij}^2 \Big{]} = \sum_{ij} f_{ij}(y),
\end{aligned}
\end{equation*} where we define $f_{ij}(y) = (\Tilde{d}_{ij}-y)^2 + y^2$. We have shown that the objective function is separable, so Problem \eqref{opt:fixed_X} can be solved by minimizing each function $f_{ij}(y)$. $f_{ij}(y)$ is a convex quadratic function, and simple univariate calculus allows us to conclude that it achieves its minimum when $y^*=\frac{\Tilde{d}_{ij}}{1+\mu}$. The minimum value of $f_{ij}$ is therefore $f_{ij}(y^*)=\frac{\mu}{1+\mu}\Tilde{d}_{ij}^2$. However, due to the sparsity constraint on $\bm{Y}$, at most $k_1$ entries of $\bm{Y}$ can be non-zero. By introducing binary variables $s_{ij}$ and noting that $f_{ij}(0) = \Tilde{d}_{ij}^2$, we can rewrite the objective of problem 2 as a function of the binary matrix $\bm{S}$:
\begin{equation*}
\begin{aligned}
    f(\bm{S}) &= \mathlarger{\sum}_{ij} \bigg{[}s_{ij} \cdot \frac{\mu}{1+\mu}\Tilde{d}_{ij}^2 + (1-s_{ij}) \cdot \Tilde{d}_{ij}^2 \bigg{]}.
\end{aligned}
\end{equation*} Due to the sparsity constraint, at most $k_1$ of the variables $s_{ij}$ can be $1$ while all others must be $0$. If $s_{ij}=0$, the objective increases by $\Tilde{d}_{ij}^2$ whereas if $s_{ij} = 1$, the objective only increases by $\frac{\mu}{1+\mu}\Tilde{d}_{ij}^2$. It follows immediately that the objective will be minimized when $s_{ij}=1$ if and only if $\Tilde{d}_{ij}$ is one of the $k_1$ largest entries in absolute value of the matrix $\bm{\Tilde{D}}$. Note that in the case that the $k_1^{th}$ largest entry in absolute value and the $(k_1 + 1)^{th}$ largest entry in absolute value are not distinct, the tie can be broken arbitrarily. Letting $\bm{S}^*$ represent the binary matrix formed by an optimal choice of the binary variables $s_{ij}$, the solution to Problem \eqref{opt:fixed_X} is given by $\bm{Y}^* = \bm{S}^* \circ \Big{(}\frac{\bm{\Tilde{D}}}{1 + \mu}\Big{)}$.
\end{proof}

{\color{black}\section{Supplemental Computational Results}} \label{sec:appendix_tables}

\begin{sidewaystable}
  \centering
  \caption{ \color{black}Comparison of average low-rank matrix reconstruction error generated by S-PCP, GoDec, ScaledGD, AccAltProj, fRPCA, and Algorithm \ref{alg:AM}. Results are reported for the exact SVD implementation of GoDec. Averaged over $10$ trials for each parameter configuration.}\label{tbl:spcp_godec_alg1}
  \begin{tabular}{ccc || cc || cccccc}
\toprule
    \multicolumn{3}{c}{} & \multicolumn{2}{c}{} & \multicolumn{6}{c}{L Error} \\
    \cmidrule(l){1-3} \cmidrule(l){4-5} \cmidrule(l){6-11}
    N & $k_0$ & $k_1$ & S-PCP Rank & S-PCP Sparsity &  S-PCP &  GoDec & ScaledGD & AccAltProj & fRPCA &     Alg 1 Exact \\
\midrule
 20 &     1 &    20 &        5.7 &           95.6 & 0.0176 & 0.0101 &   0.0082 &             0.0111 & 0.0088 & \textbf{0.0072} \\
 20 &     2 &    40 &       12.0 &          197.4 & 0.0178 & 0.0430 &   0.0062 &             0.0074 & 0.0068 & \textbf{0.0057} \\
 20 &     3 &    60 &       15.3 &          275.3 & 0.1123 & 0.1136 &   0.0084 &             0.0083 & 0.0077 & \textbf{0.0075} \\
 20 &     4 &    80 &       17.5 &          341.1 & 0.1510 & 0.3247 &   0.0087 &             0.0092 & 0.0088 & \textbf{0.0079} \\
 40 &     2 &    80 &        5.4 &          286.6 & 0.0233 & 0.0121 &   0.0147 &             0.0168 & 0.0174 & \textbf{0.0110} \\
 40 &     4 &   160 &       16.4 &          417.2 & 0.0272 & 0.0189 &   0.0122 &             0.0143 & 0.0136 & \textbf{0.0113} \\
 40 &     6 &   240 &       27.3 &          731.3 & 0.0334 & 0.0996 &   0.0159 &             0.0171 & 0.0165 & \textbf{0.0145} \\
 40 &     8 &   320 &       36.7 &         1365.1 & 0.0453 & 0.3225 &   0.0170 &             0.0178 & 0.0157 & \textbf{0.0149} \\
 60 &     3 &   180 &        7.8 &          631.6 & 0.0311 & 0.0158 &   0.0182 &             0.0231 & 0.0197 & \textbf{0.0149} \\
 60 &     6 &   360 &       13.0 &          777.6 & 0.0328 & 0.0247 &   0.0171 &             0.0222 & 0.0177 & \textbf{0.0150} \\
 60 &     9 &   540 &       36.3 &         1181.1 & 0.0439 & 0.0520 &   0.0236 &             0.0251 & 0.0226 & \textbf{0.0202} \\
 60 &    12 &   720 &       55.9 &         2930.5 & 0.0577 & 0.2696 &   0.0236 &             0.0316 & 0.0242 & \textbf{0.0209} \\
 80 &     4 &   320 &       10.9 &         1128.5 & 0.0345 & 0.0176 &   0.0230 &             0.0272 & 0.0238 & \textbf{0.0166} \\
 80 &     8 &   640 &       15.4 &         1380.1 & 0.0448 & 0.0293 &   0.0240 &             0.0314 & 0.0248 & \textbf{0.0223} \\
 80 &    12 &   960 &       34.0 &         1634.6 & 0.0569 & 0.0537 &   0.0271 &             0.0307 & 0.0269 & \textbf{0.0246} \\
 80 &    16 &  1280 &       62.7 &         3316.8 & 0.0737 & 0.2989 &   0.0339 &             0.0378 & 0.0339 & \textbf{0.0300} \\
100 &     5 &   500 &       13.8 &         1771.6 & 0.0443 & 0.0255 &   0.0288 &             0.0383 & 0.0267 & \textbf{0.0239} \\
100 &    10 &  1000 &       19.2 &         2139.9 & 0.0531 & 0.0357 &   0.0318 &             0.0385 & 0.0345 & \textbf{0.0271} \\
100 &    15 &  1500 &       36.4 &         2525.9 & 0.0640 & 0.0679 &   0.0356 &             0.0392 & 0.0330 & \textbf{0.0304} \\
100 &    20 &  2000 &       63.4 &         3145.1 & 0.0840 & 0.3675 &   0.0399 &             0.0471 & 0.0395 & \textbf{0.0381} \\
120 &    12 &  1440 &       21.3 &         3067.7 & 0.0644 & 0.0423 &   0.0368 &             0.0474 & 0.0400 & \textbf{0.0333} \\
120 &    18 &  2160 &       38.8 &         3628.4 & 0.0789 & 0.0858 &   0.0440 &             0.0497 & 0.0424 & \textbf{0.0388} \\
120 &    24 &  2880 &       72.0 &         4288.3 & 0.0968 & 0.3838 &   0.0512 &             0.0570 & 0.0498 & \textbf{0.0464} \\
140 &     7 &   980 &       19.3 &         3436.0 & 0.0613 & 0.0365 &   0.0386 &             0.0553 & 0.0375 & \textbf{0.0331} \\
140 &    21 &  2940 &       37.9 &         4911.8 & 0.0868 & 0.0910 &   0.0506 &             0.0573 & 0.0479 & \textbf{0.0442} \\
140 &    28 &  3920 &       76.7 &         5790.9 & 0.1085 & 0.4156 &   0.0607 &             0.0695 & 0.0598 & \textbf{0.0566} \\
\bottomrule
\end{tabular}

\end{sidewaystable}

\begin{sidewaystable}
  \centering
  \caption{ {\color{black} Bound gap of Algorithm \ref{alg:AM} derived using \eqref{opt:convex_relax}. Averaged over $10$ trials for each parameter configuration.}}\label{tbl:alg1_exact_accelrated_gap}
  \begin{tabular}{ccc || cccccc || cc}
\toprule
    \multicolumn{3}{c}{} & \multicolumn{6}{c}{L Error} & \multicolumn{2}{c}{} \\
    \cmidrule(l){1-3} \cmidrule(l){4-9} \cmidrule(l){10-11}
    N & $k_0$ & $k_1$ & S-PCP & GoDec & ScaledGD & AccAltProj & fRPCA & Alg 1 Exact & Alg 1 Bound Gap & Bound Time (s) \\
\midrule
 20 &     1 &    20 & .0176 & .0101 &   0.0082 &     0.0111 & 0.0088 & \textbf{0.0072} &          0.7052 &         3.7200 \\
 60 &     6 &   360 & .0328 & .0247 &   0.0171 &     0.0222 & 0.0177 & \textbf{0.0150} &          0.8543 &       189.1900 \\
 60 &     9 &   540 & .0439 &  .052 &   0.0236 &     0.0251 & 0.0226 & \textbf{0.0202} &          0.8601 &       184.9500 \\
 60 &    12 &   720 & .0577 & .2696 &   0.0236 &     0.0316 & 0.0242 & \textbf{0.0209} &          0.7709 &       155.2800 \\
 80 &     4 &   320 & .0345 & .0176 &   0.0230 &     0.0272 & 0.0238 & \textbf{0.0166} &          0.9180 &       577.8400 \\
 80 &     8 &   640 & .0448 & .0293 &   0.0240 &     0.0314 & 0.0248 & \textbf{0.0223} &          0.9267 &       765.9100 \\
 80 &    12 &   960 & .0569 & .0537 &   0.0271 &     0.0307 & 0.0269 & \textbf{0.0246} &          0.7944 &       691.5500 \\
 80 &    16 &  1280 & .0737 & .2989 &   0.0339 &     0.0378 & 0.0339 & \textbf{0.0300} &          0.7803 &       611.4700 \\
100 &     5 &   500 & .0443 & .0255 &   0.0288 &     0.0383 & 0.0267 & \textbf{0.0239} &          0.9592 &      1936.2600 \\
100 &    10 &  1000 & .0531 & .0357 &   0.0318 &     0.0385 & 0.0345 & \textbf{0.0271} &          0.9382 &      2987.0800 \\
100 &    15 &  1500 &  .064 & .0679 &   0.0356 &     0.0392 & 0.0330 & \textbf{0.0304} &          0.9062 &      2224.6100 \\
 20 &     2 &    40 & .0178 &  .043 &   0.0062 &     0.0074 & 0.0068 & \textbf{0.0057} &          0.5935 &         3.8200 \\
100 &    20 &  2000 &  .084 & .3675 &   0.0399 &     0.0471 & 0.0395 & \textbf{0.0381} &          0.8145 &      2188.6600 \\
120 &    12 &  1440 & .0644 & .0423 &   0.0368 &     0.0474 & 0.0400 & \textbf{0.0333} &          0.8951 &      6759.9200 \\
120 &    18 &  2160 & .0789 & .0858 &   0.0440 &     0.0497 & 0.0424 & \textbf{0.0388} &          0.8968 &      6878.3600 \\
120 &    24 &  2880 & .0968 & .3838 &   0.0512 &     0.0570 & 0.0498 & \textbf{0.0464} &          0.7877 &      5310.5800 \\
140 &     7 &   980 & .0613 & .0365 &   0.0386 &     0.0553 & 0.0375 & \textbf{0.0331} &          0.9014 &     14731.2500 \\
140 &    21 &  2940 & .0868 &  .091 &   0.0506 &     0.0573 & 0.0479 & \textbf{0.0442} &          0.8854 &     11260.5200 \\
140 &    28 &  3920 & .1085 & .4156 &   0.0607 &     0.0695 & 0.0598 & \textbf{0.0566} &          0.8116 &     11840.3000 \\
 20 &     3 &    60 & .1123 & .1136 &   0.0084 &     0.0083 & 0.0077 & \textbf{0.0075} &          0.5443 &         3.9600 \\
 20 &     4 &    80 &  .151 & .3247 &   0.0087 &     0.0092 & 0.0088 & \textbf{0.0079} &          0.7146 &         4.0500 \\
 40 &     2 &    80 & .0233 & .0121 &   0.0147 &     0.0168 & 0.0174 & \textbf{0.0110} &          0.8214 &        30.6200 \\
 40 &     4 &   160 & .0272 & .0189 &   0.0122 &     0.0143 & 0.0136 & \textbf{0.0113} &          0.8804 &        27.9200 \\
 40 &     6 &   240 & .0334 & .0996 &   0.0159 &     0.0171 & 0.0165 & \textbf{0.0145} &          0.7937 &        28.4700 \\
 40 &     8 &   320 & .0453 & .3225 &   0.0170 &     0.0178 & 0.0157 & \textbf{0.0149} &          0.7051 &        23.8700 \\
 60 &     3 &   180 & .0311 & .0158 &   0.0182 &     0.0231 & 0.0197 & \textbf{0.0149} &          0.8075 &       154.9000 \\
\bottomrule
\end{tabular}

\end{sidewaystable}

\begin{sidewaystable}
  \centering
  \caption{Running time of the exact implementation of Algorithm \ref{alg:AM} and the accelerated implementation of Algorithm \ref{alg:AM}. In the exact implementation, the SVD step is computed exactly, whereas in the accelerated implementation, a randomized SVD is employed in all but the final SVD step. Averaged over $10$ trials for each parameter configuration.}\label{tbl:alg1_exact_accelerated_time}
  \begin{tabular}{ccc || cc || cc ||c}
\toprule
    \multicolumn{3}{c}{} & \multicolumn{2}{c}{L Error} & \multicolumn{2}{c}{Time (s)} & \multicolumn{1}{c}{}\\
    \cmidrule(l){1-3} \cmidrule(l){4-5} \cmidrule(l){6-7} \cmidrule(l){8-8}
    N &  $k_0$ &  $k_1$ & Alg 1 Exact & Alg 1 Acc & Alg 1 Exact & Alg 1 Acc &  Time Decrease (\%) \\
\midrule
 20 &      1 &     20 &     \textbf{0.0072} &            0.0094 &               0.1351 &    \textbf{0.0986} &              27.06 \\
 20 &      2 &     40 &     \textbf{0.0057} &            0.0084 &               0.2342 &    \textbf{0.1071} &              54.27 \\
 20 &      3 &     60 &     \textbf{0.0075} &            0.0084 &               0.5713 &    \textbf{0.1394} &              75.59 \\
 20 &      4 &     80 &     \textbf{0.0079} &            0.0085 &               0.8126 &    \textbf{0.1519} &              81.31 \\
 40 &      2 &     80 &     \textbf{0.0110} &            0.0123 &               0.4157 &    \textbf{0.1982} &              52.31 \\
 40 &      4 &    160 &     \textbf{0.0113} &            0.0139 &               0.9250 &    \textbf{0.2536} &              72.59 \\
 40 &      6 &    240 &     \textbf{0.0145} &            0.0183 &               2.0046 &    \textbf{0.3574} &              82.17 \\
 40 &      8 &    320 &     \textbf{0.0149} &            0.0192 &               2.8281 &    \textbf{0.4309} &              84.76 \\
 60 &      3 &    180 &     \textbf{0.0149} &            0.0178 &               0.7407 &    \textbf{0.3964} &              46.47 \\
 60 &      6 &    360 &     \textbf{0.0150} &            0.0198 &               2.2547 &    \textbf{0.5103} &              77.37 \\
 60 &      9 &    540 &     \textbf{0.0202} &            0.0286 &               4.4260 &    \textbf{0.6930} &              84.34 \\
 60 &     12 &    720 &     \textbf{0.0209} &            0.0300 &               7.2143 &    \textbf{0.8724} &              87.91 \\
 80 &      4 &    320 &     \textbf{0.0166} &            0.0199 &               1.2156 &    \textbf{0.6214} &              48.88 \\
 80 &      8 &    640 &     \textbf{0.0223} &            0.0331 &               4.1513 &    \textbf{0.8543} &              79.42 \\
 80 &     12 &    960 &     \textbf{0.0246} &            0.0399 &               8.0393 &    \textbf{1.1153} &              86.13 \\
 80 &     16 &   1280 &     \textbf{0.0300} &            0.0488 &              13.5348 &    \textbf{1.2970} &              90.42 \\
100 &      5 &    500 &     \textbf{0.0239} &            0.0289 &               1.5669 &    \textbf{0.9722} &              37.95 \\
100 &     10 &   1000 &     \textbf{0.0271} &            0.0439 &               6.4084 &    \textbf{1.2111} &              81.10 \\
100 &     15 &   1500 &     \textbf{0.0304} &            0.0540 &              12.8520 &    \textbf{1.5614} &              87.85 \\
100 &     20 &   2000 &     \textbf{0.0381} &            0.0671 &              13.5619 &    \textbf{1.4767} &              89.11 \\
120 &     12 &   1440 &     \textbf{0.0333} &            0.0564 &               9.2897 &    \textbf{1.6930} &              81.78 \\
120 &     18 &   2160 &     \textbf{0.0388} &            0.0752 &              18.0824 &    \textbf{2.1187} &              88.28 \\
120 &     24 &   2880 &     \textbf{0.0464} &            0.0932 &              19.8079 &    \textbf{1.9967} &              89.92 \\
140 &      7 &    980 &     \textbf{0.0331} &            0.0428 &               2.6152 &    \textbf{1.6039} &              38.67 \\
140 &     21 &   2940 &     \textbf{0.0442} &            0.0922 &              18.1729 &    \textbf{2.1653} &              88.08 \\
140 &     28 &   3920 &     \textbf{0.0566} &            0.1296 &              29.6370 &    \textbf{2.6352} &              91.11 \\
\bottomrule
\end{tabular}

\end{sidewaystable}

\begin{sidewaystable}
  \centering
  \caption{{\color{black} Low-rank matrix reconstruction error, sparse matrix reconstruction error and execution time of Algorithm 1, GoDec and ScaledGD.}}\label{tbl:scaling_data}
  \begin{tabular}{c c c || ccc || ccc || ccc}
\toprule
    \multicolumn{3}{c}{} & \multicolumn{3}{c}{L Error} & \multicolumn{3}{c}{S Error} & \multicolumn{3}{c}{Time (s)}\\
    \cmidrule(l){1-3} \cmidrule(l){4-6} \cmidrule(l){7-9} \cmidrule(l){10-12}
    N & $k_0$ & $k_1$ & Alg 1 Exact & GoDec & ScaledGD & Alg 1 Exact & GoDec & ScaledGD & Alg 1 Exact & GoDec & ScaledGD \\
\midrule
  200 &      5 &    500 &     \textbf{0.0442} &        0.0458 &           0.0449 &     \textbf{0.5677} &        0.9246 &           0.7379 &               0.0185 &          0.0187 &   \textbf{0.0134} \\
  250 &      5 &    500 &     \textbf{0.0538} &        0.0553 &           0.0544 &     \textbf{0.6176} &        1.0208 &           0.7417 &      \textbf{0.0191} &          0.0250 &            0.0225 \\
  300 &      5 &    500 &     \textbf{0.0641} &        0.0654 &           0.0644 &     \textbf{0.6725} &        1.1036 &           0.7741 &               0.0314 & \textbf{0.0290} &            0.0321 \\
  350 &      5 &    500 &     \textbf{0.0755} &        0.0766 &           0.0757 &     \textbf{0.7307} &        1.1955 &           0.8259 &               0.0436 & \textbf{0.0411} &            0.0454 \\
  400 &      5 &    500 &     \textbf{0.0852} &        0.0863 &           0.0854 &     \textbf{0.7716} &        1.2483 &           0.8578 &               0.0574 & \textbf{0.0517} &            0.0562 \\
  450 &      5 &    500 &     \textbf{0.0970} &        0.0980 &           0.0972 &     \textbf{0.8038} &        1.2918 &           0.9134 &               0.0792 & \textbf{0.0712} &            0.0751 \\
  500 &      5 &    500 &     \textbf{0.1083} &        0.1093 &           0.1085 &     \textbf{0.8530} &        1.3585 &           0.9746 &               0.0906 &          0.0918 &   \textbf{0.0895} \\
  550 &      5 &    500 &     \textbf{0.1213} &        0.1222 &           0.1215 &     \textbf{0.8918} &        1.4021 &           1.0518 &               0.1138 & \textbf{0.1049} &            0.1083 \\
  600 &      5 &    500 &     \textbf{0.1322} &        0.1331 &           0.1324 &     \textbf{0.9377} &        1.4593 &           1.1210 &               0.1357 &          0.1384 &   \textbf{0.1228} \\
  650 &      5 &    500 &     \textbf{0.1430} &        0.1438 &           0.1433 &     \textbf{0.9624} &        1.4842 &           1.1881 &      \textbf{0.1538} &          0.1693 &            0.1590 \\
  700 &      5 &    500 &     \textbf{0.1554} &        0.1562 &           0.1556 &     \textbf{1.0126} &        1.5524 &           1.2712 &               0.1810 &          0.2022 &   \textbf{0.1587} \\
  750 &      5 &    500 &     \textbf{0.1681} &        0.1689 &           0.1682 &     \textbf{1.0244} &        1.5587 &           1.3332 &      \textbf{0.3668} &          0.5669 &            0.5734 \\
  800 &      5 &    500 &     \textbf{0.1812} &        0.1820 &  \textbf{0.1812} &     \textbf{1.0676} &        1.6062 &           1.4105 &      \textbf{0.3395} &          0.5000 &            1.1244 \\
  850 &      5 &    500 &              0.1918 &        0.1925 &  \textbf{0.1917} &     \textbf{1.0967} &        1.6372 &           1.4958 &      \textbf{0.9337} &          1.0395 &            1.3067 \\
  900 &      5 &    500 &              0.2057 &        0.2064 &  \textbf{0.2056} &     \textbf{1.1348} &        1.6852 &           1.5847 &               1.7587 &          1.5853 &   \textbf{1.1520} \\
  950 &      5 &    500 &     \textbf{0.2174} &        0.2181 &           0.2175 &     \textbf{1.1543} &        1.6942 &           1.6608 &               0.7749 & \textbf{0.7494} &            2.0345 \\
 1000 &      5 &    500 &              0.2306 &        0.2313 &  \textbf{0.2305} &     \textbf{1.1783} &        1.7207 &           1.7417 &               3.2104 & \textbf{3.1600} &            3.3916 \\
 2000 &      2 &    500 &     \textbf{0.5171} &        0.5177 &           0.5173 &     \textbf{1.5707} &        2.1098 &           3.6181 &               1.3195 &          1.3155 &   \textbf{1.0648} \\
 4000 &      2 &    500 &     \textbf{1.3013} &        1.3019 &           1.3018 &     \textbf{2.1207} &        2.6438 &           7.9775 &              35.1148 &         36.9397 &  \textbf{19.1202} \\
 6000 &      2 &    500 &     \textbf{2.3694} &        2.3700 &           2.3704 &     \textbf{2.3058} &        2.7742 &          11.9133 &              84.7058 &         87.7330 &  \textbf{64.5782} \\
 8000 &      2 &    500 &     \textbf{3.5365} &        3.5373 &  \textbf{3.5365} &     \textbf{2.5880} &        3.0463 &          16.8837 &             158.5785 &        160.0202 & \textbf{132.8005} \\
10000 &      2 &    500 &     \textbf{4.8465} &        4.8472 &           4.8486 &     \textbf{2.7586} &        3.1967 &          21.5332 &    \textbf{133.3238} &        145.8102 &          249.2882 \\
\bottomrule
\end{tabular}

\end{sidewaystable}

\begin{figure*}[h]\centering
  \includegraphics[width=0.9\textwidth]{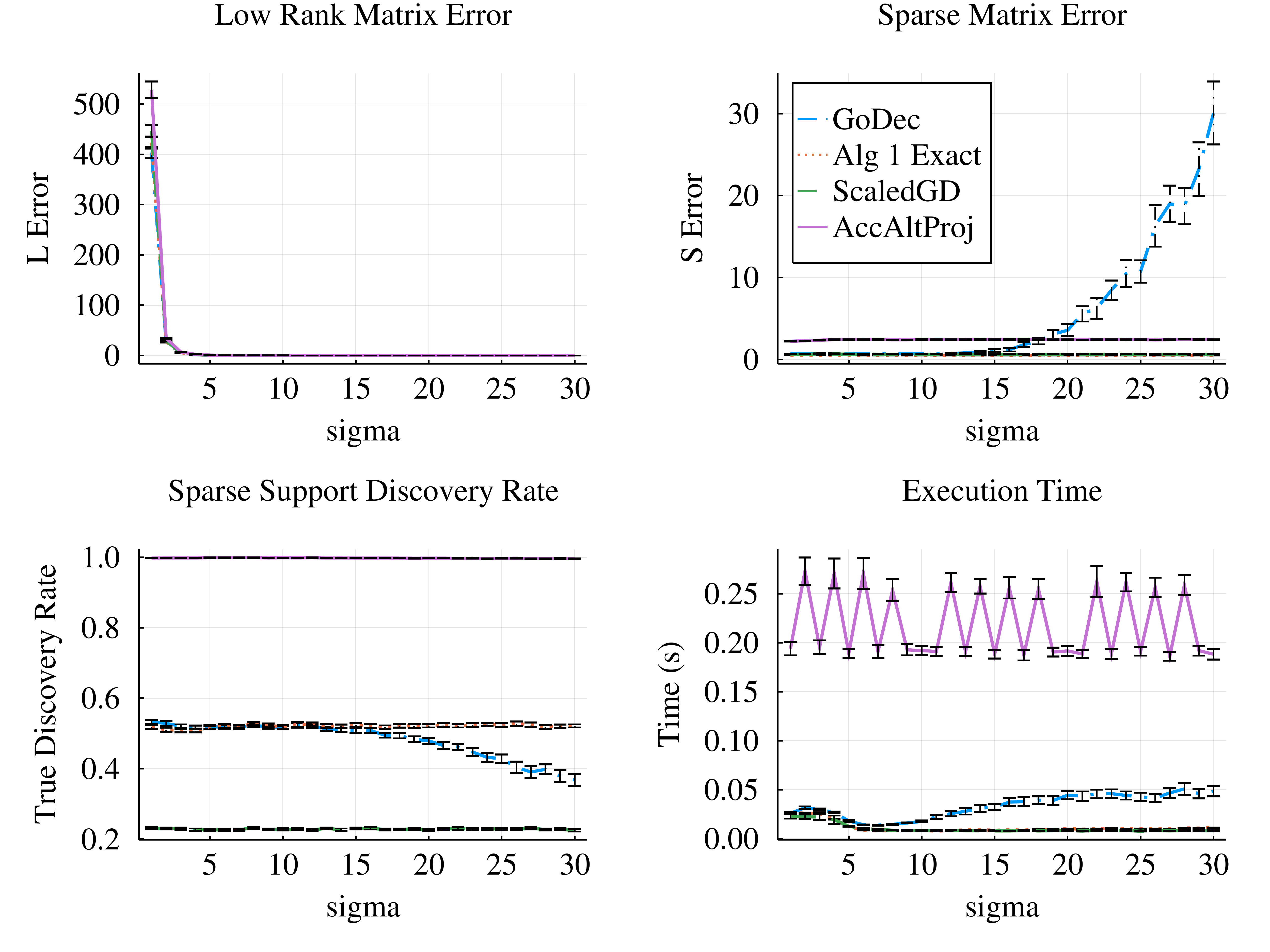}
  \caption{ \color{black} Low-rank matrix reconstruction error (top left), sparse matrix reconstruction error (top right), sparse support discovery rate (bottom left) and execution time (bottom right) versus $\sigma$ with $n=100$, $k_0 = 5$ and $k_1 = 500$. Averaged over $50$ trials for each parameter configuration.}
  \label{fig:noise_full}
\end{figure*}

\FloatBarrier

\end{document}